\documentclass[twoide,11pt]{article}
\usepackage{jair,theapa,rawfonts}
\usepackage[T1]{fontenc}
\usepackage[latin9]{inputenc}
\usepackage{array}
\usepackage{lipsum,calc}
\usepackage{verbatim}
\usepackage{float}
\usepackage{multirow}
\usepackage{amsmath, amssymb}
\usepackage{amsthm}
\usepackage{tablefootnote}
\usepackage{pgfplots}
\usepackage{booktabs}
\usepackage{threeparttable, tablefootnote}
\usepackage{colortbl}
\usepackage{xcolor}
\usepackage{hhline}
\usepackage{graphics}
\usepackage{bigstrut} 
\usepackage{soul}
\usepackage{url}
\usepackage{enumitem} 

\makeatletter
\let\MYcaption\@makecaption
\makeatother
\usepackage[font=footnotesize]{subcaption}
\makeatletter
\let\@makecaption\MYcaption
\makeatother

\makeatletter
\newcommand*{\rom}[1]{\expandafter\@slowromancap\romannumeral #1@}
\makeatother

\floatstyle{ruled}
\newfloat{algorithm}{tbp}{loa}
\providecommand{\algorithmname}{Algorithm}
\floatname{algorithm}{\protect\algorithmname}

\usepackage{algorithm, algpseudocode}

\algnewcommand{\Initialize}[1]{\State \textbf{Initialize: \newline}}
\algnewcommand{\LineComment}[1]{\State  \(\triangleright\) #1 \hfill~}
\algnewcommand{\IIf}[1]{\State\algorithmicif\ #1\ \algorithmicthen\ }
\algnewcommand{\IElse}{\unskip\ \algorithmicelse\ }
\algnewcommand{\EndIIf}{\unskip}
\algnewcommand{\pElse}[1]{\State \algorithmicelse\ #1}

\algnewcommand{\FFor}[1]{\State\algorithmicfor\ #1 \algorithmicdo }
\algnewcommand{\EndFFor}{\unskip}

\theoremstyle{definition}
\newtheorem{definition}{Definition}
\theoremstyle{proposition}
\newtheorem{proposition}{Proposition}
\theoremstyle{theorem}
\newtheorem{theorem}{Theorem}

\newcounter{step}[section]
\newcommand\step{\refstepcounter{step}\par \textit{\underline{ Step \thestep}:~}}

\usepackage{tikz}
\usepackage{pgfplots}
\usetikzlibrary{automata, positioning, shapes.geometric, arrows, decorations, fit, backgrounds}
\usetikzlibrary{trees,positioning,shapes,shadows,arrows}
\usetikzlibrary{arrows.meta}
\usepgfplotslibrary{statistics}
\usetikzlibrary{backgrounds}

\tikzset{orgClique/.style={thick, draw=gray, rectangle, align=center, minimum height=16pt}}
\tikzset{pathClique/.style={thick, draw=gray, rectangle, align=center, fill=red!20, minimum height=16pt}}
\tikzset{newSubtreeClique/.style={thick, draw=gray, rectangle, align=center, fill=teal!20, minimum height=16pt}}
\tikzset{newClique/.style={thick, draw=gray, rectangle, align=center, fill=teal!20, minimum height=16pt}}
\tikzstyle{loop1} = [draw,line width=5pt,-,red!20]
\tikzstyle{loop2} = [draw,line width=5pt,-,teal!20]	
\tikzstyle{loop3} = [draw,line width=5pt,-,magenta!40]
\tikzset{graphNode/.style={draw=gray, circle, align=center, inner sep=1pt, minimum size=10pt}}
\tikzset{reverseclip/.style={insert path={(-100cm,100cm) rectangle (100cm,-100cm)}}}
\definecolor{mygray}{rgb}{.906,  .902,  .902}
\definecolor{mygreen}{rgb}{.886,  .937,  .855}
\definecolor{myblue}{rgb}{.867,  .922,  .969}
\definecolor{myyellow}{rgb}{1,  .949,  .8}
\definecolor{myred}{rgb}{.969,  .835,  .859}
\newcommand{\drawoutside}[2][]{
	\begin{scope}[#1, even odd rule]
		\begin{pgfinterruptboundingbox}
			\clip[reverseclip] #2;
		\end{pgfinterruptboundingbox}
		\path[draw, fill=none] #2;
	\end{scope}
	\path [#1, draw=none] #2;
}

\usepackage{float}
\usepackage{caption}
\usepackage{pifont}

\usepackage{pgfplots}
\usepackage{tikz}

\tikzset{roundnode/.style={circle,draw, minimum size=3mm,inner sep=2pt},
  squarenode/.style={rectangle, draw=blue!60, very thick, minimum size=3mm}}

\tikzset{fatarrow/.style={single arrow,shape border rotate=270,
                          thick,draw=blue!70,fill=blue!30,
                          minimum height=10mm}}

\tikzstyle{vertex}=[circle,fill=black!20,minimum size=15pt,inner sep=2pt]
\tikzstyle{vertexpt}=[circle,fill=black!25,minimum size=10pt,inner sep=0pt]
\tikzstyle{vertexr}=[rectangle,fill=black!15,minimum size=20pt,inner sep=2pt]
\tikzstyle{vertexrnofill}=[rectangle,draw=black,minimum size=20pt,inner sep=2pt]
\tikzstyle{str}=[rectangle,fill=black!15,draw=black,minimum size=20pt,inner sep=2pt]
\tikzstyle{selected vertex}=[rectangle,fill=red!15,minimum size=20pt,inner sep=2pt]
\tikzstyle{selected vertext}=[rectangle,fill=teal!20,minimum size=20pt,inner sep=2pt]
\tikzstyle{selected vertexy}=[rectangle,fill=orange!20,minimum size=20pt,inner sep=2pt]
\tikzstyle{selected vertexr}=[rectangle,fill=cyan!15,minimum size=20pt,inner sep=2pt]
\tikzstyle{selected vertexb}=[rectangle,fill=blue!15,minimum size=20pt,inner sep=2pt]
\tikzstyle{edge} = [draw,thick,->]
\tikzstyle{dashededge} = [draw,dashed,thick,-]
\tikzstyle{udashededge} = [draw,ultra thick,dashed,magenta]
\tikzstyle{tealedge} = [draw,ultra thick,->,teal]
\tikzstyle{blackedge} = [draw,ultra thick,->]
\tikzstyle{edge1} = [draw,thick,<-]
\tikzstyle{edge2} = [draw,thick,-]
\tikzstyle{edge3} = [draw,thick,blue,->]
\tikzstyle{edge4} = [draw,blue,ultra thick]
\tikzstyle{edge5} = [->, >=latex, line width=1pt]
\tikzstyle{weight} = [font=\small,blue]
\tikzstyle{selected edge} = [draw,line width=5pt,->,red!50]
\tikzstyle{ignored edge} = [draw,line width=5pt,-,black!20]
\tikzstyle{loop1} = [draw,line width=5pt,-,red!20]
\tikzstyle{loop2} = [draw,line width=5pt,-,teal!20]	
\tikzstyle{loop3} = [draw,line width=5pt,-,magenta!40]
\usetikzlibrary{arrows}
\usetikzlibrary{shapes.geometric,backgrounds}

\newcommand{\cmark}{\ding{51}}%

\begin{document}

\title{IBIA: Bayesian Inference via Incremental Build-Infer-Approximate operations on Clique Trees.}

\author{\name Shivani~Bathla \email ee13s064@ee.iitm.ac.in\\
  \name Vinita~Vasudevan \email vinita@ee.iitm.ac.in \\
	\addr Department of Electrical Engineering,
		IIT Madras, Chennai 600036
	}
\maketitle
\begin{abstract}
Exact inference in Bayesian networks is intractable and has an exponential dependence on the size of the largest clique in the corresponding clique tree (CT), necessitating approximations.
   Factor based methods to bound clique sizes are more accurate than structure based methods, but expensive since they involve inference of beliefs in a large number of candidate structure or region graphs.
We propose an alternative approach for approximate inference based on an \textit{incremental build-infer-approximate} (IBIA) paradigm, which converts the Bayesian network into a data structure containing a sequence of linked clique tree forests (SLCTF), with clique sizes bounded by a user-specified value. In the \textit{incremental build} stage of this approach, CTFs are constructed incrementally by adding variables to the CTFs as long as clique sizes are within the specified bound. Once the clique size constraint is reached, the CTs in the CTF are calibrated in the \textit{infer} stage of IBIA. The resulting clique beliefs are used in the \textit{approximate} phase to get an approximate CTF with reduced clique sizes. The approximate CTF forms the starting point for the next CTF in the sequence. These steps are repeated until all variables are added to a CTF in the sequence. We prove that our algorithm for incremental construction of clique trees always generates a valid CT and our approximation technique preserves the joint beliefs of the variables within a clique.
Based on this, we show that the SLCTF data structure can be used for efficient approximate inference of partition function and prior and posterior marginals.
More than 500 benchmarks were used to test the method and the results show a significant reduction in error when compared to other approximate methods, with competitive runtimes.

\end{abstract}

\section{Introduction}\label{sec:introduction}
Bayesian Network (BN) based algorithms have been used for probabilistic inference in a wide variety of applications. 
Methods for exact inference include variable elimination \cite{Dechter96}, belief propagation (BP) on join trees \cite{Lauritzen1988,Shenoy1986} and weighted model counting \cite{Darwiche01,Chavira2006,Chavira08,Sang2005,Dudek2020,Dilkas2021}. Exact inference is known to be \#P complete \cite{Roth1996}, being time and space exponential in the treewidth of the graph. Even some relatively small BNs have large treewidth, necessitating approximations. Although approximate inference with error bounds is also NP-hard \cite{Roth1996}, many of the approximation techniques work well in practice.

A widely used method of approximation is the ``loopy belief propagation'' (LBP) algorithm, which is based on using Pearl's message passing (MP) algorithm for trees iteratively on networks with cycles \cite{Frey1998,Murphy1999}. The converged solution of LBP  was shown to be a fixed point of the Bethe free energy  \cite{Yedidia2000,Heskes03}. 
Alternatives to the basic LBP include fractional BP (FBP) \cite{Wiegerinck2003}, the tree-reweighted BP (TRWBP) \cite{Wainwright2003}, mean field (MF) \cite{Winn2005} and expectation propagation (EP) \cite{Minka2001,Minka2004a}, all of which can also be viewed as minimization of a particular information divergence measure \cite{Minka2005}. 
LBP and its variants have been empirically found to give good solutions in many cases, especially if the influence of the loops is not very large.
  
  The accuracy can be improved by performing iterative belief propagation between clusters of nodes or regions \cite{Yedidia2000,Dechter02,Yedidia2005,Mooij07}. These algorithms include generalized belief propagation (GBP), iterative join graph propagation (IJGP), cluster variation (CVM) and region graph based methods. The converged solutions of GBP are shown to be the fixed points of the Kikuchi free energy \cite{Kikuchi51}.
    The two issues in these methods are convergence and accuracy of the converged solution. 
The iterative message passing algorithm for inference of beliefs is not guaranteed to converge in general, but there are alternative methods of solution that guarantee convergence \cite{Yuille2002,Heskes2006}. 
    Convergence of these methods has been studied in  \citeA{Elidan06}.
    More recently, there are methods~\shortcite{Heess2013,Lin2015,Yoon2019,Kuck2020} that use neural networks to learn messages and accelerate convergence.

     The accuracy of the solution obtained using GBP and related methods depends on the choice of regions that give bounded cluster sizes.
 There are several approaches for region identification. Methods proposed in  \citeA{Gelfand2012,Welling2005}, \shortciteA{Dechter02,Mateescu2010} use structural information of the graph to bound the size of the clusters. 
Methods like region pursuit \cite{Welling2004,Hazan2012}, cluster pursuit \shortcite{Sontag2008} and triplet region construction (TRC) \cite{Lin2020} use factor based information. Although factor based partitioning is preferable, it typically involves multiple rounds of region reconstruction/addition and belief estimation, which can become expensive \cite{Welling2004}. To reduce this computational cost, localized computations are used to add clusters to the outer region \cite{Sontag2008,Lin2020}. In \citeA{Sontag2008}, several potential clusters are scored based on the potential improvement of a lower bound using pairwise and joint factors. A problem is huge number of candidate clusters. TRC uses differences in conditional entropy computed using locally computed pseudo-marginals to identify interaction triplets.

Another technique for approximate inference based on BP  is the use of approximate factorized messages. Algorithms in this class include mini-bucket elimination (MBE) \cite{Dechter97,Dechter03} and weighted MBE (WMB) \shortcite{Liu2011,Forouzan2015,Lee2020} and mini-clustering (MC) \shortcite{Mateescu2010}.
These algorithms trade off accuracy and complexity by limiting the scope of the factors in a mini-bucket or a mini-cluster to a preset bound ($ibound$).
 The key idea in these algorithms is the following - 
 instead of using the sum/max operator after finding the product of the factors in a cluster, they migrate the sum/max operator to each mini-cluster to give an upper bound on the beliefs and the partition function.   WMB uses weighted factors and the Holder inequality to get the approximate messages. MC uses mini-bucket heuristics to partition clusters in a join-tree into mini clusters. Most of the algorithms use scope based partitioning of buckets or clusters. As is the case with region selection, scope based partitioning can be carried out in many ways and each partition could give a different result.
Like the cluster and region pursuit methods to identify optimum regions, factor based information to find clusters has been explored in \citeA{Rollon2010} and \citeA{Forouzan2015}. In \citeA{Rollon2010}, a greedy heuristic based on a local relative error is used to find a partition that best approximates the bucket function. In \citeA{Forouzan2015}, after using WMB on a join graph to get initial beliefs, mini-buckets within a bucket are considered for merging based on a score that is computed using localized changes in messages. When compared to the method in \citeA{Sontag2008}, the search for candidate region is more structured and well defined. 

Another approach is to bound clique sizes by simplifying the network. In the thin junction tree algorithm~\shortcite{Bach2001,Elidan2008,Scanagatta2018}, the objective is to select the set of "dominant" features that approximate the distribution well and result in a graph that has bounded treewidth. 
The remaining features (nodes and edges of the graph) are ignored.
Feature selection is based on the gain in KL divergence which in turn requires inference of beliefs.

In a series of papers, Choi and Darwiche \cite{Choi05,Choi2006,Choi07,Choi08} propose the relax-compensate-recover scheme which starts with a simpler network obtained by deleting edges, duplicating nodes and adding auxiliary evidence nodes along with some consistency conditions. In each iteration of the compensate and recover phase, new edges are added to obtain better approximations. Beliefs for each configuration are obtained using iterative belief propagation.
More generally, the relax-compensate-recover method computes beliefs by relaxing equivalence constraints and recovers by adding some of the constraints \cite{Choi2010}.

The other technique to reduce clique sizes is to partition the BN into overlapping subgraphs \cite{Xiang1993,Lin1997,Murphy2002,Xiang2003,Bhanja2004} and build a clique tree (CT) for each subgraph. However, since the partitioning is done at the network level, these methods do not guarantee a bound on the maximum clique size.  
These methods have not been used extensively and it is not clear how queries such as partition function can be handled.

    In this paper, we propose a novel framework for approximate inference that we call the \textit{incremental build-infer-approximate} (IBIA) paradigm. In this framework, each directed acyclic graph (DAG) in the Bayesian network (BN) is converted into a derived data structure that contains a \textit{Sequence of Linked Clique Tree Forests} (SLCTF).
The maximum clique size in the clique tree forests (CTFs) is bounded by a user-defined parameter. We show that the resultant SLCTFs can be used for approximate inference of the partition function ($PR$) and the prior and posterior marginals ($MAR_p$ and $MAR_e$).

    Instead of doing several iterations to identify regions or structures with bounded clique sizes, our approach is to build the CTs incrementally, adding as many variables as possible to the CTs until the maximum clique size reaches a user-defined bound $mcs_p$. 
We propose a novel algorithm to add variables to an existing CT. We prove that our algorithm for incremental construction will always result in valid CTs. 
This is the \textit{incremental build} phase of IBIA. 

Since our aim is to do belief based approximations, the build phase is followed by an \textit{infer} phase. In this step,
the standard join tree based BP algorithm with two rounds of message passing is used to calibrate the CTs and infer the beliefs of all cliques exactly.
This is efficient because the maximum clique size in each CT of the CTF is bounded.

In the \textit{approximate} phase, the inferred beliefs are used to approximate the CTF and reduce the maximum clique size to another user-defined bound $mcs_{im}$. This approximate CTF is the starting point to construct the next CTF in the sequence. Since $mcs_{im}$ is less than $mcs_p$, it allows for addition of more variables to form the next CTF.

 The SLCTF for a DAG is constructed by repeatedly using the 
 incremental build, infer and approximate steps until all variables in the DAG are present in atleast one of the CTFs in the SLCTF. A property of our approximation algorithm is that it preserves both the probability of evidence and the joint beliefs of variables within a clique. This property is used to link cliques in adjacent CTFs. The resultant linked CTFs are used for approximate inference of queries.

 Additionally, our algorithm has the following features.
  \begin{itemize}
  \item IBIA gives CTFs with a user-specified maximum clique size without partitioning at the network level. Finding partitions of a network that meet a specified bound on the clique size is difficult.
  \item The trade-off between accuracy and runtime can be achieved using two user-defined parameters,  $mcs_p$ and $mcs_{im}$.
\item Since CTFs in the SLCTF are built incrementally, IBIA allows for the flexibility of incorporating incremental addition/deletions to portions of the network.  
      \item  IBIA is non-iterative in the sense that it does not require inference of beliefs in a large number of candidate structure or region graphs for choice of optimal regions.
         Nor does it have loopy join graphs, which require iterative BP or its convergent alternative. 
  \end{itemize}

We have compared the results of our algorithm with three other deterministic approximate inference methods -  FBP,  WMB and IJGP for 500+ benchmarks. 
The focus of this paper is ``deterministic'' approximate inference as opposed to sampling based techniques. However, we have also compared our results with Gibbs sampling implemented in libDAI \cite{LibdaiPaper} and wherever possible with the results reported in \citeA{Gogate2011} for IJGP with sample search.

The rest of this paper is organized as follows.
Section 2 provides background and notation.
We present an overview of the IBIA framework and the main algorithm in Section 3, the methodology for constructing the SLCTF in Section 4, approximate inference of queries in Section 5 and solution guarantees and complexity analysis in Section 6. 
Section 7 has a comparison of IBIA with related work. Section 8 has the results. 
Finally, we present our conclusions in Section 9. We show proofs for all propositions and theorems in Appendix~\ref{app:proofs}, evaluation of inference algorithms used for comparison in Appendix~\ref{app:eval} and a glossary of terms used in multiple algorithms in Appendix~\ref{app:notations}.

\section{Background}\label{sec:background}

This section has the notation and the definitions related to inference on  Bayesian networks. Throughout the paper, we use the terms clique tree, join tree and junction tree interchangeably. Also, as is common in the literature, we use the term $C_i$ as a label for the clique as well as to denote the set of variables in the clique.

\begin{definition}
    Bayesian Network (BN):
        A Bayesian network consists of one or more directed acyclic graphs (DAG), $G$.
    Nodes in a BN represent random variables $\mathcal{X} = \{ X_1, X_2, \cdots X_n\}$ with associated domains $D = \{D_{X_1},D_{X_2}, \cdots D_{X_n}\}$. It has directed edges from a subset of nodes $\textrm{Pa}_{X_i} = \{X_k, k \neq i\}$ to $X_i$, representing a conditional probability distribution (CPD) $P_i = P\{X_i|\textrm{Pa}_{X_i}\}$. The BN represents the joint probability distribution (JPD) of $\mathcal{X}$, given by $P(\mathcal{X}) = \prod_{i=1}^n P_i$.
\end{definition}
Throughout the paper, we use the terms variables and nodes in the BN interchangeably.
\begin{definition}\label{def:parent}
    Parent variables ($\textrm{Pa}_{X_i}$): Parents of a variable $X_i$ are the variables connected via an incoming edge to $X_i$ in the DAG.
\end{definition}
\begin{definition}
    Child variables: Children of a variable $X_i$ are variables connected to it via outgoing edges in the DAG.
\end{definition}
\begin{definition}\label{def:pi}
  Primary inputs (PI): The BN allows for a natural topological ordering of variables. We use PI to denote the set of variables that do not have any parents.
\end{definition}

\begin{definition}
    Moralized graph ($\mathcal{M}(G)$): It is an undirected graph over $\mathcal{X}$ that contains an edge between two variables $X_i$ and $X_j$ if (a) there is a directed edge in either direction between $X_i$ and $X_j$ in $G$ or (b) $X_i$ and $X_j$ are parents of a variable in $G$.
\end{definition}
\begin{definition}
  Chordal graph ($\mathcal{H}$): It is an undirected graph that contains no cycle of length greater than three.

  It is obtained from $\mathcal{M}(G)$ by triangulating all cycles of length greater than three.
\end{definition}
\begin{definition}
    Clique: A clique is a subset of nodes in an undirected graph such that all pairs of nodes are adjacent.
\end{definition}
\begin{definition}
    Maximal clique: A clique in an undirected graph is called a maximal clique if it is not contained within any other clique in the graph.
\end{definition}
\begin{definition}
  \label{def:vct}
    Junction tree or Join tree or Clique tree (CT) \cite{Koller2009}: The CT is a hypertree with nodes $\{C_1,C_2,\cdots, C_n\}$ that are the set of cliques in $\mathcal{H}$.  An edge between $C_i$ and $C_j$ is associated with a sepset $S_{i,j} = C_i \cap C_j$. The CT satisfies the following
  \begin{enumerate}
    \item[(a)] All cliques are maximal cliques i.e., there is no $C_j$ such that $C_j \subset C_i$.
  \item[ (b)] Each factor $P_i$ is associated with a single node $C_i$ such that $\textrm{Scope}(P_i) \subseteq \mathcal{C}_i$.
    The product of all factors gives the joint probability distribution over all variables in the BN.
   \item[ (c)] It satisfies the \textit{running intersection property}(RIP), which states that for all variables $X$, if  $X \in C_i$ and $X \in C_j$, then $X$ is present in every node in the unique path between $C_i$ and $C_j$.
  \end{enumerate}
\end{definition}
Exact inference in a CT is done using the belief propagation (BP) algorithm~\cite{Lauritzen1988,Koller2009} that is equivalent to two rounds of message passing along the edges of the CT, an upward pass (from the leaf nodes to the root node) and a downward pass (from the root node to the leaves). Following this each node has an associated belief $\beta(C_i) = P(C_i) = \sum\limits_{\mathcal{X}\setminus C_i}P(\mathcal{X})$. 

A calibrated CT is defined as follows.
\begin{definition}\label{def:calCT}
Calibrated CT~\cite{Koller2009}:  Let $\beta(C_i)$ and $\beta(C_j)$ denote the beliefs associated with adjacent cliques $C_i$ and $C_j$. The cliques are said to be calibrated if 
   \begin{equation*}
   \sum\limits_{C_i\setminus S_{i,j}}\beta(C_i) =   \sum\limits_{C_j\setminus S_{i,j}}\beta(C_j) = \mu(S_{i,j})
  \end{equation*}
  Here, $S_{i,j}$ is the sepset corresponding to $C_i$ and $C_j$ and $\mu_{i,j}$ is the associated belief. The CT is said to be calibrated if all pairs of adjacent cliques are calibrated.
\end{definition}
In a calibrated CT, adjacent cliques agree on the marginals. Hence, the marginal of a variable or a set of variables can be computed from any clique containing the variable. The joint probability distribution, $P(\mathcal{X})$,  can be reparameterized in terms of the sepset and clique beliefs as follows:
\begin{equation*}
  P(\mathcal{X}) = \frac{\prod_{i \in \mathcal{V}_T} \beta(C_i)}{\prod_{(i-j)\in \mathcal{E}_T}\mu(S_{i,j})}
\end{equation*}
where $\mathcal{V}_T$ and $\mathcal{E}_T$ are the set of nodes and edges in the CT.

The complexity of exact inference is space and time exponential in the treewidth, defined as follows.
\begin{definition}
  Treewidth: The treewidth is one less than the minimum possible value of the maximum clique size over all possible triangulations of the moralized graph.
\end{definition}

The queries we are interested in are the prior and posterior marginals ($MAR_p,~MAR_e$) and the partition function ($PR$). Definitions related to these are the following.
\begin{definition}
  Evidence Variables ($E_v$): Evidence variables are set of instantiated variables with corresponding observed state $e$.
\end{definition}
\begin{definition}\label{def:pr}
    Partition function (PR): It is defined as the probability of evidence~$P(E_v=\{e\})$.
\end{definition}
In the presence of evidence, the beliefs obtained after calibration are un-normalized. The \textit{normalization constant} obtained by summing over any clique belief in a calibrated CT is the same. It is equal to the partition function.
\begin{definition}
    Posterior marginals ($MAR_e$): It is the conditional probability of a variable $X$, given a fixed evidence state $e$, $P(X|e)$.
\end{definition}
\begin{definition}
    Prior marginals ($MAR_p$): It is the marginal probability of a variable $X$ when there is no evidence.
\end{definition}

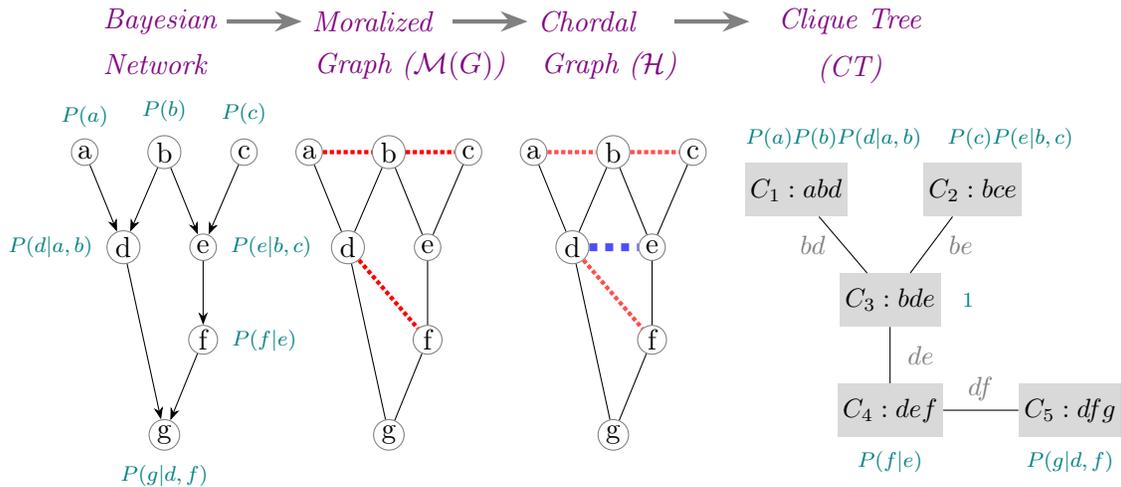
\begin{figure}
	\centering  
	\begin{tikzpicture}
	\node (d) {};
	\node (bnLabel) [below=0 of d, xshift=-1cm] { \bf \it \color{violet} Bayesian};
	\node (b2) [align=left, below=0 of bnLabel] { \bf \it \color{violet} Network };
	\node (mnLabel) [align=left, right=1 of bnLabel] { \bf \color{violet} \it Moralized };
	\node (m2) [align=left, below= 0 of mnLabel, xshift=0.5cm] { \bf \color{violet} \it Graph ($\mathcal{M}(G)$) };
	\node (cnLabel) [align=left, right= 1 of mnLabel] { \bf \it \color{violet} Chordal };
	\node (c2) [align=left, below= 0 of cnLabel, xshift=0.3cm] { \bf \color{violet} \it Graph ($\mathcal{H}$) };
	\node (ctLabel) [align=left, right= 1.5 of cnLabel] { \bf \it \color{violet} Clique Tree};
	\node (ct2) [align=left, below= 0 of ctLabel] { \bf \color{violet} \it (CT)};
	
	\draw [-{Stealth[scale=1]}, ultra thick, color=gray](bnLabel)--(mnLabel);
	\draw [-{Stealth[scale=1]}, ultra thick, color=gray](mnLabel)--(cnLabel);
	\draw [-{Stealth[scale=1]}, ultra thick, color=gray]([xshift=0.25cm]cnLabel.east) -- ([xshift=-0.25cm]ctLabel.west);
	
	\node[graphNode](abn) [align=center, below=1.25 of bnLabel, xshift=-1cm] {a};
	\node (al) [align=center, above=0.02 of abn] {\scriptsize \color{teal}$P(a)$};
	\node[graphNode](bbn) [align=center, right=0.65 of abn] {b};
	\node (bl) [align=center, above=0.05 of bbn] {\scriptsize \color{teal}$P(b)$};
	\node[graphNode](cbn) [align=center, right=0.65 of bbn] {c};
	\node (cl) [align=center, above=0.05 of cbn] {\scriptsize \color{teal}$P(c)$};
	\node[graphNode] (dbn) [align=center, below right=0.32 of abn, yshift=-0.75cm] {d};
	\node (dl) [align=center, left=0.05 of dbn] {\scriptsize \color{teal}$P(d|a,b)$};
	\node[graphNode] (ebn) [align=center, below right=0.32 of bbn, yshift=-0.75cm] {e};
	\node (el) [align=center, right=0.05 of ebn] {\scriptsize \color{teal}$P(e|b,c)$};
	\node[graphNode] (fbn) [align=center, below=0.1 of ebn, yshift=-0.75cm] {f};
	\node (fl) [align=center, right=0.05 of fbn] {\scriptsize \color{teal}$P(f|e)$};
	\node[graphNode] (gbn) [align=center, below left=0.32 of fbn, yshift=-0.75cm] {g};
	\node (gl) [align=center, below=0.05 of gbn] {\scriptsize \color{teal}$P(g|d,f)$};
	\draw [-{Stealth}] (abn)--(dbn);
	\draw [-{Stealth}] (bbn)--(dbn);
	\draw [-{Stealth}] (bbn)--(ebn);
	\draw [-{Stealth}] (cbn)--(ebn);
	\draw [-{Stealth}] (ebn)--(fbn);
	\draw [-{Stealth}] (dbn)--(gbn);
	\draw [-{Stealth}] (fbn)--(gbn);

	\node[graphNode](amn) [align=center, right=0.5 of cbn] {a};
	\node[graphNode](bmn) [align=center, right=0.65 of amn] {b};
	\node[graphNode](cmn) [align=center, right=0.65 of bmn] {c};
	\node[graphNode] (dmn) [align=center, below right=0.32 of amn, yshift=-0.75cm] {d};
	\node[graphNode] (emn) [align=center, below right=0.32 of bmn, yshift=-0.75cm] {e};
	\node[graphNode] (fmn) [align=center, below=0.1 of emn, yshift=-0.75cm] {f};
	\node[graphNode] (gmn) [align=center, below left=0.32 of fmn, yshift=-0.75cm] {g};
	\draw (amn)--(dmn);
	\draw (bmn)--(dmn);
	\draw (bmn)--(emn);
	\draw (cmn)--(emn);
	\draw (emn)--(fmn);
	\draw (dmn)--(gmn);
	\draw (fmn)--(gmn);	
	\draw [densely dotted, color=red, ultra thick] (amn)--(bmn);
	\draw [densely dotted, color=red, ultra thick] (bmn)--(cmn);
	\draw [densely dotted, color=red, ultra thick] (dmn)--(fmn);

	\node[graphNode](acn) [align=center, right=0.5 of cmn] {a};
	\node[graphNode](bcn) [align=center, right=0.65 of acn] {b};
	\node[graphNode](ccn) [align=center, right=0.65 of bcn] {c};
	\node[graphNode] (dcn) [align=center, below right=0.32 of acn, yshift=-0.75cm] {d};
	\node[graphNode] (ecn) [align=center, below right=0.32 of bcn, yshift=-0.75cm] {e};
	\node[graphNode] (fcn) [align=center, below=0.1 of ecn, yshift=-0.75cm] {f};
	\node[graphNode] (gcn) [align=center, below left=0.32 of fcn, yshift=-0.75cm] {g};
	\draw (acn)--(dcn);
	\draw (bcn)--(dcn);
	\draw (bcn)--(ecn);
	\draw (ccn)--(ecn);
	\draw (ecn)--(fcn);
	\draw (dcn)--(gcn);
	\draw (fcn)--(gcn);	
	\draw [densely dotted, color=red!70, ultra thick] (acn)--(bcn);
	\draw [densely dotted, color=red!70, ultra thick] (bcn)--(ccn);
	\draw [densely dotted, color=red!70, ultra thick] (dcn)--(fcn);
	\draw [dashed,color=blue!70, line width=3pt] (dcn)--(ecn);
	
	\node [vertexr](c1) [align=center, right=0.5 of ccn, yshift=-0.5cm] {\small $C_1: abd$};
	\node (c1l) [align=center, above=0.05 of c1, xshift=0.5cm] {\scriptsize \color{teal}$P(a)P(b)P(d|a,b)$};
	\node [vertexr](c2) [align=center, right=1 of c1] {\small $C_2: bce$};
	\node (c5l) [align=center, above=0.05 of c2, xshift=0.5cm] {\scriptsize \color{teal}$P(c) P(e|b,c)$};
	\node [vertexr](c3) [align=center, below=0.75 of c1, xshift=1.25cm] {\small $C_3: bde$};
	\node (c3r) [align=center, right=0.05 of c3, xshift=0.1cm] {\scriptsize \color{teal}$1$};
	\node [vertexr](c4) [align=center, below=0.75 of c3] {\small $C_4: def$};
	\node (c4l) [align=center, below=0.05 of c4] {\scriptsize \color{teal}$P(f|e)$};
	\node [vertexr](c5) [align=center, right=1 of c4] {\small $C_5: dfg$};
	\node (c5l) [align=center, below=0.05 of c5] {\scriptsize \color{teal}$P(g|d,f)$};
	\draw (c1)--(c3) node [midway, left, xshift=-0.1cm] {\small \color{gray}$bd$};
	\draw (c2)--(c3) node [midway, right, xshift=0.1cm] {\small \color{gray}$be$};
	\draw (c3)--(c4) node [midway, right, xshift=0.1cm] {\small \color{gray}$de$};
	\draw (c4)--(c5) node [midway, above] {\small \color{gray}$df$};


	\end{tikzpicture}
    \caption{Compilation of a BN into a Clique Tree (CT). The dotted red lines are the moralizing edges between parents, and the dashed blue lines are the fill-in edges introduced during triangulation. Cliques in the CT are shown in gray boxes and the variables contained in each clique are indicated inside the box. The sepset variables corresponding to each edge in the CT are marked in gray and the factors assigned to each clique are shown in teal color.}
	\label{fig:ctModel}
\end{figure}

 Figure \ref{fig:ctModel} illustrates the steps involved in converting a BN into a CT model that can be used for inference. The BN is first converted to a moralized graph by removing edge orientations and adding pairwise edges between the parents of each variable, marked in the figure using dashed red lines. Next, the \textit{chordal} completion of the moralized graph is obtained by adding edges to break all loops of size greater than three. In the figure, loop $d\mbox{-}b\mbox{-}e\mbox{-}f\mbox{-}d$ is broken by adding a chord between variables $d$ and $e$. Finally, the CT is formed with the maximal cliques of the chordal graph as nodes. Each factor $P_i$ is assigned to a clique containing its scope. Following this, two rounds of message passing is performed to get the calibrated clique and sepset beliefs. 
 Since finding a triangulation in which the size of the largest clique is equal to the treewidth is NP-hard, several heuristics have been proposed. In this paper, we have used variable elimination along with the min-fill heuristic~\cite{Zhang1996,Koller2009}. 

\section{Overview of the IBIA paradigm and the main algorithm}\label{sec:overview}
This section gives an overview of the main data structure and definitions of terms used in various algorithms. This is followed by a description of the main algorithm. We also introduce a running example that will be used in various sections of this paper to illustrate the constituent algorithms.
\subsection{Overview and main data structure}


The IBIA framework uses a series of incremental build, infer and approximate steps to derive a data structure that we call Sequence of Linked Clique Tree Forests (SLCTF). The SLCTF  can be used for efficient approximate inference of the partition function and the prior and posterior marginals. The inputs to the algorithm are the BN, the set of evidence variables and two user-defined parameters, $mcs_p$ and $mcs_{im}$, which are used to specify clique size bounds.
The BN could consist of multiple DAGs. This can happen if the BN contains independent sets of variables or if the evidence variables split the BN into multiple DAGs. An SLCTF is built for each DAG separately. All the SLCTFs corresponding to a BN are used for inferring queries.

\begin{figure}[!htb]
\centering
\begin{tikzpicture}[thick,scale=0.7, every node/.style={transform shape}]
    \node (BN) at (1.5,1) {$\mathbf{G\in BN}$};
    \foreach \pos/\name in {(0,0)/v1, (1.5,0)/v2, (3,0)/v3, (-0.5,-1)/v4, (0.5,-1)/v5, (1.5,-1)/v6, (2.5,-1)/v7, (3.5,-1)/v8}
    \node[vertexpt] (\name) at \pos {};
    \foreach \pos/\name in {(-1,-2)/v9, (1.5,-2)/v10, (2.5,-2)/v11, (3.5,-2)/v12}
    \node[vertexpt] (\name) at \pos {};
    \foreach \pos/\name in {(-1.5,-3)/d1, (0.5,-3)/d2, (1.5,-3)/d3, (2.5,-3)/d4, (3.5,-3)/d5}
    \node (\name) at \pos {};
    \foreach \source/\dest  in {v1/v4, v1/v5, v2/v6, v3/v7, v3/v8, v4/v9, v4/v10, v6/v10, v6/v11, v7/v12, v8/v12, v7/v11}
    \path[edge] (\source) -- (\dest);
    \foreach \source/\dest  in {v9/d1, v9/d2, v10/d3, v11/d4, v12/d5}
    \path[dashededge] (\source) -- (\dest);

    \foreach \pos/\name/\label in {(9,1)/$\mathbf{CTF_1:}$/CTF1, (11,1)/$\mathbf{CT_{1,1}}$/CT1, (13,1)/$\mathbf{CT_{1,2}}$/CT2}
    \node (\label) at \pos {\color{blue}\name};
    \path[tealedge] (BN) -- node [above, midway, xshift=0.5cm]{\textbf{Build ($mcs_p$), Infer}} (CTF1);

    \foreach \pos/\name/\label in {(9,-0.75)/$\mathbf{CTF_{1,a}}$/CTF1a}
    \node (\label) at \pos {\color{cyan}\name};
    \path[blackedge] (CTF1) -- node(dx) [left, midway]{\textbf{Approximate~($mcs_{im}$)}} (CTF1a);

    \foreach \pos/\name/\label in {(9,-2.5)/$\mathbf{CTF_2:}$/CTF2, (12,-2.5)/$\mathbf{CT_{2,1}}$/CT21}
    \node (\label) at \pos {\color{blue}\name};
    \path[tealedge] (CTF1a) -- node (db1) [left, midway, yshift=0.25cm]{\textbf{Build ($mcs_p$),}}   (CTF2);
    \draw (db1) ++(-0.75,-0.5) node{\color{teal}\textbf{Infer}};

    \foreach \pos/\name/\label in {(9,-4.5)/$\mathbf{CTF_{n-1}:}$/CTFn1, (12,-4.5)/$\mathbf{CT_{n-1,1}}$/CTn1}
    \node (\label) at \pos {\color{blue}\name};
    \foreach \pos/\name/\label in { (9,-6)/$\mathbf{CTF_{n-1,a}}$/CTF2a}
    \node (\label) at \pos {\color{cyan}\name};
    \path[blackedge] (CTFn1) -- (CTF2a);
        \node [circle, fill, minimum size=3pt,inner sep=0pt, outer sep=0pt] (c1) at ([yshift=-0.4cm]CTF2.south) {};
        \node [circle, fill, minimum size=3pt,inner sep=0pt, outer sep=0pt] (c2) at ([yshift=-0.65cm]CTF2.south) {};
        \node [circle, fill, minimum size=3pt,inner sep=0pt, outer sep=0pt] (c3) at ([yshift=-0.9cm]CTF2.south) {};
        \node [circle, fill, minimum size=3pt,inner sep=0pt, outer sep=0pt] (c1) at ([yshift=-0.4cm]CT21.south) {};
        \node [circle, fill, minimum size=3pt,inner sep=0pt, outer sep=0pt] (c2) at ([yshift=-0.65cm]CT21.south) {};
        \node [circle, fill, minimum size=3pt,inner sep=0pt, outer sep=0pt] (c3) at ([yshift=-0.9cm]CT21.south) {};

    \foreach \pos/\name/\label in {(9,-7.5)/$\mathbf{CTF_n:}$/CTF3, (12,-7.5)/$\mathbf{CT_{n,1}}$/CT31}
    \node (\label) at \pos {\color{blue}\name};
    \path[tealedge] (CTF2a) --  (CTF3);

    \path[udashededge] (CT1) -- (CT21);
    \path[udashededge] (CT2) -- node[right,midway]{\color{magenta} $\mathbf{IM_1}$} (CT21);
    \path[udashededge] (CT31) -- node[right,midway]{\color{magenta} $\mathbf{IM_{n-1}}$} (CTn1);

    \node (slctf) [below=0.5 of CTF3,xshift=-0.5cm] {\textbf{SLCTF}};
    \node (lctf) [below=1.0 of CTF3,xshift=2cm] {\color{blue} $\mathbf{L_{CTF}=[CTF_1,CTF_2,\hdots,CTF_n]}$};
    \node (lim) [below=1.5 of CTF3,xshift=1.35cm] {\color{magenta} $\mathbf{L_{IM}=[IM_1,\hdots,IM_{n-1}]}$};
    \node (ei) [below=2 of CTF3,xshift=0.2cm] {$I_E=m$};
      mybackground/.style={execute at end picture={
        \begin{scope}[on background layer]
    \draw [fill=gray!20] (slctf.north west) rectangle ([xshift=4cm]ei.south east);
    \draw [fill=gray!20] (-1.7,1.5) rectangle (3.8,-3.8);
        \end{scope}
    }},
             
\end{tikzpicture}
    \caption{Steps involved in converting a DAG $G\in BN$ to a sequence of linked CTFs (SLCTF) using the IBIA framework. $mcs_p$ and $mcs_{im}$ are the maximum clique sizes in $CTF_k$ and $CTF_{k,a}$ respectively. $CT_{k,j}$ denotes a CT in $CTF_k$. The SLCTF is a triplet comprising of the list of CTFs ($L_{CTF}$), list of links between adjacent CTFs ($L_{IM}$) and index of the last CTF in the sequence to which new evidence variables are added ($I_E$).
    }
\label{fig:ibia}
\end{figure}
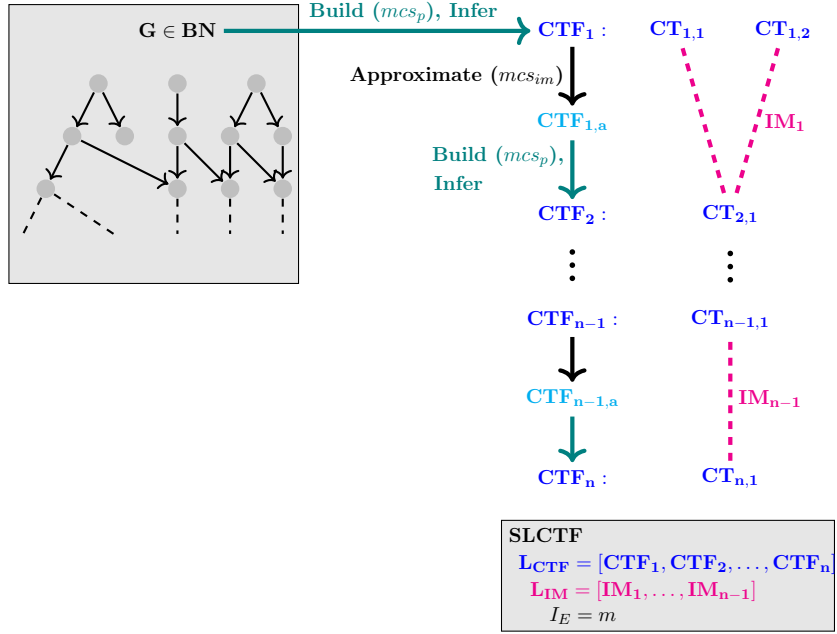
The process of constructing the SLCTF for a DAG $G \in BN$ is depicted in Figure \ref{fig:ibia}. The figure shows a DAG and the corresponding sequence of CTFs $\{CTF_k\}$, each of which consists of possibly disjoint clique trees, denoted $CT_{k,j}$. In the \textit{build} phase, a CTF is constructed incrementally by adding variables from the DAG to the CTF until the clique size reaches the specified bound $mcs_p$. This is followed by the \textit{infer} phase, where the CTF is calibrated using the standard BP algorithm~\cite{Lauritzen1988,Koller2009} for exact inference. The next step is the \textit{approximation} phase, in which the beliefs inferred from the calibrated CTs are used to compute metrics required to derive the approximate CTF, denoted as $CTF_{k,a}$ in the figure. The maximum clique size in $CTF_{k,a}$ is reduced to a lower value $mcs_{im}$. Since $CTF_{k,a}$ is the starting point for the construction of the next CTF, this allows for the addition of new variables to form the next CTF in the sequence, $CTF_{k+1}$. 
The incremental build, infer and approximate steps are used repeatedly until all variables in the DAG are present in atleast one of the CTFs in the sequence, stored as a list $L_{CTF}$. 
The index of the last CTF to which new evidence variables are added ($I_E$), is also obtained as a part of this process. For inference of posterior beliefs, we also need links between cliques in adjacent CTFs. These links are stored in \textit{interface maps} denoted $IM$. In the figure, the interface map $IM_1$ stores links between cliques in $CTF_1$ and $CTF_2$.

As shown in Figure~\ref{fig:ibia}, the SLCTF data structure is a triplet containing
\begin{enumerate}
\item $L_{CTF}$: List containing a sequence of CTFs with maximum clique size equal to $mcs_p$.
\item $L_{IM}$: A list containing interface maps ($IM$) that links cliques in adjacent CTFs.
\item $I_E$: Index of the last CTF to which new evidence variables are added.
\end{enumerate}

\subsection{Definitions}\label{sec:definitions}
We use the following definitions in the paper.
\begin{definition}\label{def:cs}
Clique size:
The clique size $cs_i$ of a clique $C_i$ is defined as follows.
  \begin{equation}\label{eqn:cs}
    cs_i = \log_2~(\prod\limits_{\forall~v~\in~ C_i} |D_v| ~)
  \end{equation}
    where $|D_v|$ is the cardinality or the number of states in the domain of the variable $v$.
\end{definition}
The clique size is defined as the logarithm (base 2) of the product of the domain sizes of the variables in the clique. It can be seen that it is the effective number of binary variables contained in the clique. It is assumed that the clique size constraint $mcs_p$ is specified so that it can accommodate the maximum CPD size in the input BN.

\begin{definition}\label{def:iv}
    Interface variables (IV): Variables in a $CTF$ whose children in the BN have not been added to it or to any preceding  CTF in the sequence. 
\end{definition}

Each CTF in the sequence has a different set of interface variables. The definition implies that the successors of these variables have not yet been added to any CTF built so far. IVs are needed to form the next CTF in the sequence. 

\begin{definition}\label{def:lv}
    Link variables (LV): All variables present in the approximate CTF are called link variables. 
\end{definition}
Link variables are defined with respect to a particular approximate CTF. They are needed to form links between adjacent cliques in $IM$.

\begin{definition}\label{def:msg}
    $MSG[V]$: Given a subset of variables ($V$) in a calibrated CTF, $MSG[V]$ is used to denote the minimal subgraph of the CTF that is needed to compute the joint beliefs of $V$. 
\end{definition}
    It is found by first identifying the subgraph of CTF that connects all the cliques that contain variables in the set $V$. 
Then, starting from the leaf nodes of the subgraph (nodes with degree equal to 1), cliques that contain the same set (or subset) of variables in $V$ as their neighbors are removed recursively.
For example, in Figure~\ref{fig:ctModel}, $MSG[\{b,g\}]$ is the subgraph of the CT containing cliques $C_3,~C_4$ and $C_5$.

\subsection{Functions and Main Algorithm}
\begin{figure}[!htb]
    \centering
    \begin{tikzpicture}[thick,scale=0.8, every node/.style={transform shape}]
    \foreach \pos/\name/\label in {(0,-2)/{IBIA}/ibia, (4,0)/{ConstructSLCTF}/cs, (4,-4)/{InferPRandMAR}/ainf, (8, 0)/{BuildCTF}/build, (8.4,-1)/{CalibrateCTF}/cal, (8.75,-2)/{ApproximateCTF}/approx, (9,-3)/{UpdateInterfaceMap}/im, (12.5,0)/{ModifyCTF}/mc}
        \node [rounded corners,vertexrnofill] (\label) at \pos {\bf \name};
        \foreach \source/\dest  in {ibia/ainf, cs/approx, cs/im, cs/im,cs/cal}
    \path [draw,edge5, rounded corners=5pt] (\source)|-(\dest);
        \foreach \source/\dest  in {cs/build,  build/mc}
    \path [draw,edge5] (\source.east)|-(\dest);
    \path [draw,edge5] (ibia.north)|-(cs);
    \end{tikzpicture}
    \caption{Flow chart of algorithms used within the IBIA framework.}
    \label{fig:algFlow}
\end{figure}
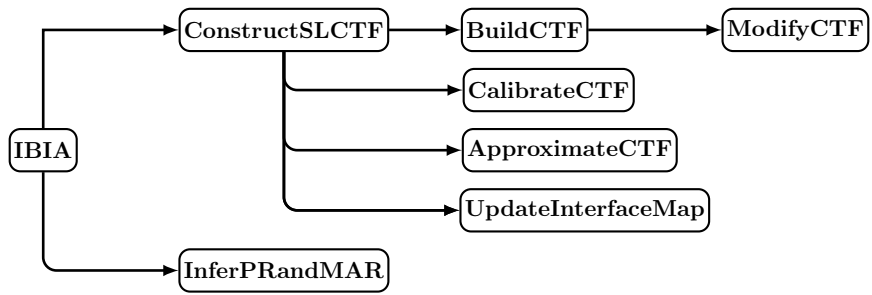

A flow chart of the algorithms used is shown in Figure~\ref{fig:algFlow}. The main algorithm (Algorithm~\ref{alg:inferenceAlgo}) calls two functions $ConstructSLCTF$ (Algorithm~\ref{alg:cSLCTF}) and $InferPRandMAR$ (Algorithm~\ref{alg:answerQueries}). $ConstructSLCTF$ constructs the SLCTF for a DAG using four functions.
      $BuildCTF$ (Algorithm~\ref{alg:BuildCTF}) uses $ModifyCTF$ (Algorithm~\ref{alg:modifyCTF}) to incrementally build the CTF and returns a CTF with  maximum clique size $mcs_p$. $CalibrateCTF$ runs the standard BP algorithm for exact inference to calibrate the CTF. $ApproximateCTF$ (Algorithm~\ref{alg:ApproximateCTF}) is used to approximate the CTF and set the links between the input and approximate CTF in the interface map $IM$. $UpdateInterfaceMap$ (Algorithm~\ref{alg:updateInterfaceMap}) completes $IM$ by adding  links between adjacent CTFs.
The algorithm $InferPRandMAR$ computes estimates of the partition function, prior and posterior singleton marginals.

Algorithm~\ref{alg:inferenceAlgo} shows the steps in the main algorithm of the IBIA framework. A Boolean variable $E$ is set if evidence variables are present and the query involves finding partition function or posterior marginals.   
This is followed by simplification of the BN and CPDs based on evidence and other fixed states using the methodology described in Section~\ref{sec:simplifyBN}. 
After simplification, we get a set of DAGs stored in the list $L_G$.
Each DAG $G$ in $L_G$ is processed separately. Depending on the indicator variable $E$, the subset of evidence variables present in $G$, $S_{eg}$ is identified. This subset is used by the function $ConstructSLCTF$ to build the SLCTF for $G$. 
    The SLCTFs for all DAGs in the BN are stored in the list $L_D$, which is used by the function $InferPRandMAR$ to get estimates of $PR,~MAR_e,~MAR_p$.

\begin{algorithm}[!h]
	\scriptsize
	\caption{IBIA~($BN,evidMap,mcs_p,mcs_{im},query$)}
	\label{alg:inferenceAlgo}
	\begin{algorithmic}[1]
		\Require ~$BN$: Bayesian Network \newline
        \indent$evidMap$: Evidence variables and corresponding states $<ev:ev_{state}>$\newline
		\indent$mcs_{p}$: Maximum clique size limit for CTFs in SLCTF\newline
		\indent$mcs_{im}$: Maximum clique size limit for the approximated $CTF$\newline
        \indent$query$: Prior marginals $MAR_p$, Partition function $PR$, Posterior marginals $MAR_e$
		\State Initialize:
        ~$E=False$ \Comment{{\color{teal!70} \scriptsize Boolean variable indicating presence of evidence}}
        \newline\indent\indent~~$L_D=[~]$ \Comment{{\color{teal!70} \scriptsize List of linked CTFs for all DAGs $\in BN$}}
        \IIf {$evidMap\neq \varnothing ~\&\&~ query\in \{MAR_e, PR\}$} $E=True$ \EndIIf
        \State $L_G\gets$ SimplifyBN($BN,~evidMap$) \Comment{{\color{teal!70}$L_G$: List of connected DAGs in the simplified BN }}
        \For{$G\in L_G$}
        \If{E}
        \State $S_{eg} \gets$ Find the evidence variables contained in $G$ 
        \Else
        \State $S_{eg}\gets\varnothing$
        \EndIf
        \State ($L_{CTF}, L_{IM}, I_E$) = ConstructSLCTF($G,S_{eg},E,mcs_p,mcs_{im}$)
        \State $L_D$.append($(L_{CTF}, L_{IM}, I_E)$) \Comment{{\color{teal!70}$L_D$: List of SLCTFs for the BN}}
       	\EndFor
        \State $PR,MAR_e,MAR_p\gets$ InferPRandMAR($L_D,query$)
        \end{algorithmic}
        \end{algorithm}

\subsection{Example}
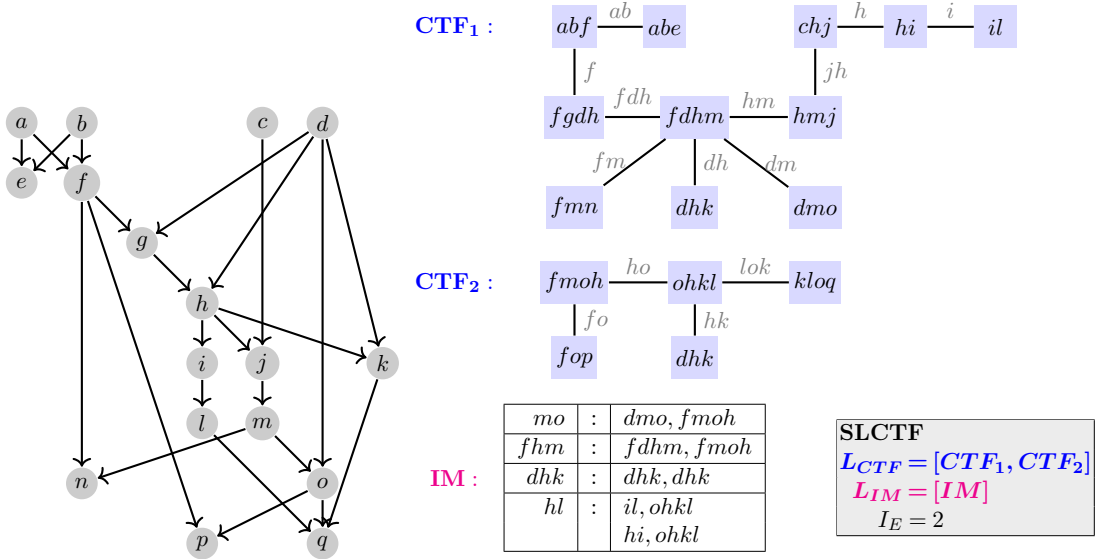
\begin{figure}[htb]
\centering
\captionsetup[subfigure]{position=b}
    \begin{subfigure}[t]{0.35\textwidth}
        \begin{tikzpicture}[thick,scale=0.8, every node/.style={transform shape}]
    \foreach \pos/\name in {(0,0)/a, (1,0)/b, (4,0)/c, (5,0)/d, (0,-1)/e, (1,-1)/f, (2,-2)/g, (3,-3)/h, (3,-4)/i, (4,-4)/j, (6,-4)/k, (3,-5)/l, (4,-5)/m, (1,-6)/n, (5,-6)/o, (3,-7)/p, (5,-7)/q}
    \node [vertex] (\name) at \pos {$\name$};

    \foreach \source/\dest  in {a/e, a/f, b/e, b/f, c/j, d/g, d/h, d/k, d/o, f/g, f/n, f/p, g/h, h/i, h/j, h/k, i/l, j/m, k/q, l/q, m/o, m/n, o/p,o/q}
    \path[edge] (\source)-- (\dest);
    
    \end{tikzpicture}
        \caption{Example BN with evidence map $evidMap=\{e:e_s, p:p_s\}$.}
        \label{fig:ex-BN}
    \end{subfigure}%
    \hfill
    \begin{subfigure}[t]{0.65\textwidth}
    \begin{tikzpicture}[thick,scale=0.8, every node/.style={transform shape}]
    \node (ctf1) at (0,2) {\color{blue}$\mathbf{CTF_1:}$};
    \foreach \pos/\name in {(2,2)/abf, (3.5,2)/abe, (2,0.5)/fgdh, (4,0.5)/fdhm, (6,0.5)/hmj, (6,2)/chj, (7.5,2)/hi, (9,2)/il, (2,-1)/fmn, (4,-1)/dhk, (6,-1)/dmo}
    \node [selected vertexb] (\name) at \pos {$\name$};
    \foreach \source/\dest/\weight  in {abe/abf/$ab$,  fgdh/fdhm/$fdh$, fdhm/hmj/$hm$, chj/hi/$h$, hi/il/$i$}
    \path[edge2] (\source) -- node [above,midway ] {\color{gray} \weight} (\dest);
    \foreach \source/\dest/\weight  in {fdhm/dhk/$dh$, fdhm/dmo/$dm$, chj/hmj/$jh$, abf/fgdh/$f$}
    \path[edge2] (\source) -- node [right,midway ] {\color{gray} \weight} (\dest);
    \foreach \source/\dest/\weight  in {fdhm/fmn/$fm$}
    \path[edge2] (\source) -- node [left,midway ] {\color{gray} \weight} (\dest);

    \foreach \pos/\name in {(2,-2.25)/fmoh, (4,-2.25)/ohkl, (6,-2.25)/kloq, (2,-3.5)/fop}
    \node [selected vertexb] (\name) at \pos {$\name$};
    \foreach \pos/\name\label in {(4,-3.5)/dhk/dhk2}
    \node [selected vertexb] (\label) at \pos {$\name$};
    \foreach \source/\dest/\weight  in {fmoh/ohkl/$ho$, ohkl/kloq/$lok$}
    \path[edge2] (\source) -- node [above,midway ] {\color{gray} \weight} (\dest);
    \foreach \source/\dest/\weight  in {fmoh/fop/$fo$, ohkl/dhk2/$hk$}
    \path[edge2] (\source) -- node [right,midway ] {\color{gray} \weight} (\dest);

    \node (ctf2) at (0,-2.25) {\color{blue}$\mathbf{CTF_2:}$};

    \node (im) at (0,-5.5) {\color{magenta}$\mathbf{IM:}$};
        
    \node (x) at (3,-5.5) {\begin{tabular}{|r|c|l|}\hline $mo$ & $:$ &${dmo,fmoh}$\\ \hline $fhm$ &$:$ &${fdhm,fmoh}$\\ \hline $dhk$&$:$&${dhk,dhk}$ \\ \hline $hl$&$:$&${il,ohkl}$ \\ &&${hi, ohkl}$ \\ \hline\end{tabular}};

    \node (x1) at (8.5,-5.5) {\setlength\tabcolsep{1pt}\begin{tabular}{|>{\columncolor{gray!15}}r>{\columncolor{gray!15}}c>{\columncolor{gray!15}}l|}\hline  
         \multicolumn{3}{|l|}{\cellcolor{gray!15}\textbf{SLCTF}}\\
    \color{blue} $\boldsymbol{L_{CTF}}$ &\color{blue} $\boldsymbol{=}$ & \color{blue} $\boldsymbol{[CTF_1,CTF_2]}$\\
    \color{magenta} $\boldsymbol{L_{IM}}$ & \color{magenta} $\boldsymbol{=}$ & \color{magenta} 
    $\boldsymbol{[IM]}$\\ $I_E$ &$=$&$2$ \\ \hline \end{tabular}};

    \end{tikzpicture}
        \caption{Corresponding SLCTF obtained using Algorithm~\ref{alg:inferenceAlgo}.}
        \label{fig:exOut}
    \end{subfigure}
    \caption{Construction of sequence of linked CTFs (SLCTF) for a sample BN using IBIA. The maximum clique size constraints, $mcs_p$ and $mcs_{im}$ are set to 4 and 3 respectively. All variables in this BN are assumed to be binary.
    }
\end{figure}

We will use the BN shown in Figure~\ref{fig:ex-BN} as a running example to explain the steps used in various algorithms proposed in this work. All variables are assumed to be binary in this BN. Figure~\ref{fig:exOut} shows the corresponding SLCTF obtained using Algorithm~\ref{alg:inferenceAlgo} with clique size constraints $mcs_p$ and $mcs_{im}$ set to 4 and 3 respectively. In the following sections, we will use this example to illustrate the steps involved in the algorithms used to convert the DAG in Figure~\ref{fig:ex-BN} to the SLCTF in Figure~\ref{fig:exOut}. We will also show how this SLCTF is used for approximate inference.

\section{Construction of the SLCTF}
As mentioned in the overview, the construction of the SLCTF for a DAG $G$ involves a series of incremental build, infer and approximate steps until all the variables of a DAG are added to it. This is followed by a function to update the interface links. This section contains the methods used in these algorithms and some propositions based on the algorithms. Detailed proofs are discussed in Appendix A.

The goal is to build the SLCTF for a graph $G$ of the BN. As explained in section \ref{sec:overview}, the SLCTF is basically a triplet that contains a list of CTFs ($L_{CTF}$), a list of interface maps ($L_{IM}$) and the index of the last CTF to which evidence variables are added ($I_E$).

\begin{algorithm}[!h]
	\scriptsize
	\caption{ConstructSLCTF($G,S_{eg},E,mcs_p,mcs_{im}$)}
	\label{alg:cSLCTF}
	\begin{algorithmic}[1]
		\Require ~$G$: A DAG in the BN \newline
                \indent$S_{eg}$: Set of evidence states in G\newline
                \indent$E$: A boolean variable indicating presence of evidence variables\newline
		\indent$mcs_{p}$: Maximum clique size limit for CTFs in SLCTF\newline
		\indent$mcs_{im}$: Maximum clique size limit for the approximated $CTF$
        \State Initialize: 
                $CTF=Graph()$\Comment{{\color{teal!70} \scriptsize Clique Tree Forest}}\newline
                \indent\indent~$L_{CTF}=[~],~L_{IM}=[~], I_E=0$\Comment{{\color{teal!70} \scriptsize Initialize SLCTF}} \newline
                \indent\indent~$k=1$ \Comment{{\color{teal!70} \scriptsize Index of the current CTF}}
		
		\While {$G$.size $\not=$ 0}
        \State $CTF,IV,G,S_{eg}\gets$BuildCTF($CTF,G,S_{eg},mcs_p$)\Comment{{\color{teal!70} \scriptsize Add variables to $CTF$ and remove from $G,S_{eg}$}}
        \State $CTF \gets$  CalibrateCTF \Comment{{\color{teal!70}Use Belief Propagation to calibrate the CTF}}
        \State $L_{CTF}$.append($CTF$)\Comment{{\color{teal!70} \scriptsize Add $CTF$ to the sequence}}
        \State $CTF, IM\gets$  ApproximateCTF($CTF, ~IV, ~mcs_{im},E$) 
        \Comment{{\color{teal!70} $IM$ has links between input and approximate CTFs}}
        \If {$E$} 
        \State $L_{IM}$.append($IM$)  
        \IIf{$S_{eg} == \varnothing$ \&\& $I_E == 0$} $I_E= k$ \EndIIf 
        \Comment{{\color{teal!70}Set $I_E$ once all evidence variables have been added.}}
        \EndIf
        \State $k=k+1$
	\EndWhile
        \IIf {$E$ \&\& $query==MAR_e$}  $L_{IM}\gets$ UpdateInterfaceMap($L_{CTF}, L_{IM}$) \EndIIf
        \Comment{{\color{teal!70}Add links between adjacent CTFs}}
        \State \Return ($L_{CTF}, L_{IM},I_E$)
 \end{algorithmic}
  \end{algorithm}

Algorithm \ref{alg:cSLCTF} details the steps involved in the construction of the SLCTF for DAG $G$. 
The algorithm repeatedly calls $BuildCTF$, $CalibrateCTF$ and $ApproximateCTF$ until all variables in $G$ have been added to the SLCTF. 
The function $BuildCTF$  (described in Section 4.1) incrementally builds a CTF by adding variables from $G$. 
Variables and evidence states that have been added to the CTF are removed from $G$ and $S_{eg}$. It returns a CTF with maximum clique size $mcs_p$ and the set of interface variables (IV, see Definition~\ref{def:iv}).
The CTF returned by $BuildCTF$ is calibrated using the function $CalibrateCTF$ (explained in Section 4.2).
The function $ApproximateCTF$ (described in Section 4.3) uses the calibrated CTF, $IV$ and $mcs_{im}$ to construct an approximate CTF and an interface map ($IM$) that has links between cliques in the input and approximate CTFs.
Once all the evidence variables are added ($S_{eg}$ becomes empty), $I_E$ stores the index of the corresponding CTF. 
After all CTFs are constructed, $IM$ is updated to link adjacent CTFs by the algorithm $UpdateInterfaceMap$.
Since these links are needed only for inference of posterior beliefs, it is done only if the query is $MAR_e$ and $E$ is true.
The lists $L_{CTF}$ and $L_{IM}$ store the CTFs and the interface maps.

We now describe the techniques used incrementally build and approximate the CTFs and find the links between adjacent CTFs.
\subsection{BuildCTF: Algorithm for incremental construction of CTF}
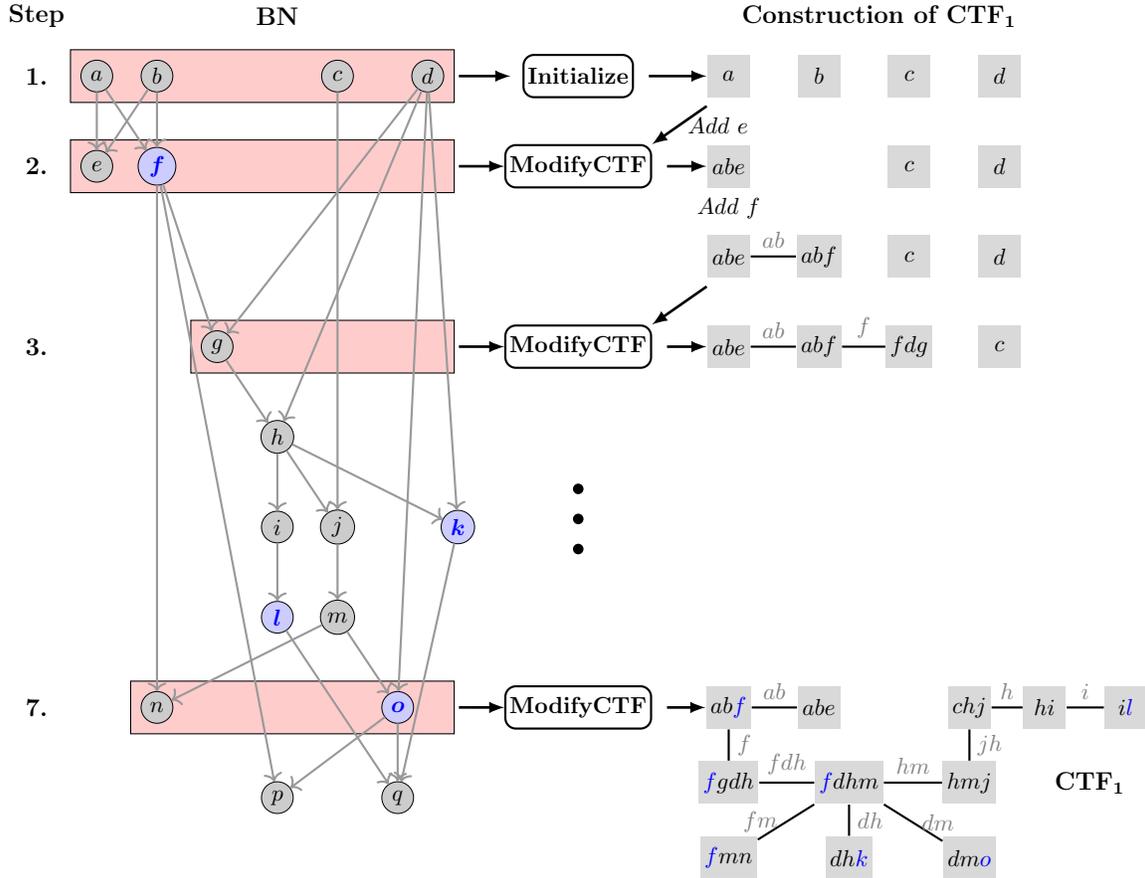
\begin{figure}[!htb]
    \centering
    \begin{tikzpicture}[thick,scale=0.8, every node/.style={transform shape}]
        \node (bn) at (3,1) {\large \textbf{BN}};
        \node (step) at (-1,1) {\large \textbf{Step}};
        \node (step1) at (-1,0) {\large \textbf{1.}};
        \node (step2) at (-1,-1.5) {\large \textbf{2.}};
        \node (step3) at (-1,-4.5) {\large \textbf{3.}};
        \node (step7) at (-1,-10.5) {\large \textbf{7.}};
        \node (ctf1) at (13,1) {\large \bf Construction of $\mathbf{CTF_1}$};
        \foreach \pos/\label/\name in {(0,0)/bna/a, (1,0)/bnb/b, (4,0)/bnc/c, (5.5,0)/bnd/d, (0,-1.5)/bne/e, (2,-4.5)/bng/g, (3,-6)/bnh/h, (3,-7.5)/bni/i, (4,-7.5)/bnj/j, (4,-9)/bnm/m, (1,-10.5)/bnn/n, (3,-12)/bnp/p, (5,-12)/bnq/q}
        \node [vertex, draw=black, ultra thin] (\label) at \pos {$\name$};
        \foreach \pos/\label/\name in { (1,-1.5)/bnf/{\color{blue}\boldsymbol f},  (6,-7.5)/bnk/{\color{blue}\boldsymbol k}, (3,-9)/bnl/{\color{blue}\boldsymbol l}, (5,-10.5)/bno/{\color{blue}\boldsymbol o}}
        \node [vertex, draw=black, ultra thin, fill=blue!20] (\label) at \pos {$\name$};

        \foreach \source/\dest  in {bna/bne, bna/bnf, bnb/bne, bnb/bnf, bnc/bnj, bnd/bng, bnd/bnh, bnd/bnk, bnd/bno, bnf/bng, bnf/bnn, bnf/bnp, bng/bnh, bnh/bni, bnh/bnj, bnh/bnk, bni/bnl, bnj/bnm, bnk/bnq, bnl/bnq, bnm/bno, bnm/bnn, bno/bnp, bno/bnq}
        \path[edge, black!40] (\source)-- (\dest);

        
        \foreach \pos/\name/\label in {(8,0)/{Initialize}/init, (8,-1.5)/{ModifyCTF}/m1, (8,-4.5)/{ModifyCTF}/m2, (8,-10.5)/{ModifyCTF}/m6}
        \node [rounded corners,vertexrnofill] (\label) at \pos {\bf \name};
        \node [circle, fill, minimum size=5pt,inner sep=0pt, outer sep=0pt] (c1) at ([yshift=-2cm]m2.south) {};
        \node [circle, fill, minimum size=5pt,inner sep=0pt, outer sep=0pt] (c2) at ([yshift=-2.5cm]m2.south) {};
        \node [circle, fill, minimum size=5pt,inner sep=0pt, outer sep=0pt] (c3) at ([yshift=-3cm]m2.south) {};

        \draw[edge5] ([xshift=0.25cm]bnd.east)--([xshift=-0.2cm]init.west);
        \draw[edge5] ([xshift=0.25cm,yshift=-1.5cm]bnd.east)--(m1);
        \draw[edge5] ([xshift=0.25cm,yshift=-4.5cm]bnd.east)--(m2);
        \draw[edge5] ([xshift=0.25cm,yshift=-10.5cm]bnd.east)--(m6);
        
        \foreach \pos/\name in {(10.5,0)/a, (12,0)/b, (13.5,0)/c, (15,0)/d}
        \node [vertexr] (\name) at \pos {$\name$};
        \draw [edge5] ([yshift=-0.5cm]a.west)--node[right, midway]{$Add~e$}(m1.north east);
        \draw [edge5] ([xshift=0.25cm]init.east)--(a);

        \foreach \pos/\name/\label in {(10.5,-1.5)/abe/abe1, (13.5,-1.5)/c/c1, (15,-1.5)/d/d1}
        \node [vertexr] (\label) at \pos {$\name$};
        \draw [edge5] ([xshift=0.25cm]m1.east)--(abe1);
        \foreach \pos/\name in {(10.5,-3)/abe, (12,-3)/abf, (13.5,-3)/c, (15,-3)/d}
        \node [vertexr] (\name) at \pos {$\name$};
        \path [edge2] (abe)--node [above, midway] {\color{gray} $ab$} (abf);
        \draw [edge5] ([yshift=-0.5cm]abe.west)--(m2.north east);
        \node [align=center,above=0.125 of abe] {$Add~f$};

        \foreach \pos/\name in {(10.5,-4.5)/abe, (12,-4.5)/abf, (13.5,-4.5)/fdg, (15,-4.5)/c}
        \node [vertexr] (\name) at \pos {$\name$};
        \foreach \source/\dest/\weight  in {abe/abf/$ab$, abf/fdg/$f$}
        \path[edge2] (\source) -- node [above,midway ] {\color{gray} \weight} (\dest);
        \draw [edge5] ([xshift=0.25cm]m2.east)--(abe);

        \foreach \pos/\label/\name in {(10.5,-10.5)/abf/$ab\textcolor{blue}{f}$, (12,-10.5)/abe/$abe$, (10.5,-11.75)/fgdh/$\textcolor{blue}{f}gdh$, (12.5,-11.75)/fdhm/$\textcolor{blue}{f}dhm$, (14.5,-11.75)/hmj/$hmj$, (14.5,-10.5)/chj/$chj$, (15.75,-10.5)/hi/$hi$, (17.1, -10.5)/il/$i\textcolor{blue}{l}$, (10.5, -13)/fmn/$\textcolor{blue}{f}mn$, (12.5,-13)/dhk/$dh\textcolor{blue}{k}$, (14.5,-13)/dmo/$dm\textcolor{blue}{o}$}
        \node [vertexr] (\label) at \pos {\name};
        \foreach \source/\dest/\weight  in {abe/abf/$ab$,  fgdh/fdhm/$fdh$, fdhm/hmj/$hm$, chj/hi/$h$, hi/il/$i$}
        \path[edge2] (\source) -- node [above,midway ] {\color{gray} \weight} (\dest);
        \foreach \source/\dest/\weight  in {fdhm/dhk/$dh$, fdhm/dmo/$dm$, chj/hmj/$jh$, abf/fgdh/$f$}
        \path[edge2] (\source) -- node [right,midway ] {\color{gray} \weight} (\dest);
        \foreach \source/\dest/\weight  in {fdhm/fmn/$fm$}
        \path[edge2] (\source) -- node [left,midway ] {\color{gray} \weight} (\dest);
        \draw [edge5] ([xshift=0.25cm]m6.east)--(abf);
        \node (ctf11) at (16.5,-11.75) {\large \bf $\mathbf{CTF_1}$};
    
      mybackground/.style={execute at end picture={
        \begin{scope}[on background layer]
            \draw [fill=red!20] ([xshift=-0.25cm,yshift=0.25cm]bna.north west) rectangle ([xshift=0.25cm,yshift=-0.25cm]bnd.south east);
            \draw [fill=red!20] ([xshift=-0.25cm,yshift=0.25cm]bne.north west) rectangle ([xshift=0.25cm,yshift=-1.75cm]bnd.south east);
            \draw [fill=red!20] ([xshift=-0.25cm,yshift=0.25cm]bng.north west) rectangle ([xshift=0.25cm,yshift=-4.75cm]bnd.south east);
            \draw [fill=red!20] ([xshift=-0.25cm,yshift=0.25cm]bnn.north west) rectangle ([xshift=0.25cm,yshift=-10.75cm]bnd.south east);
        \end{scope}
    }},
    \end{tikzpicture}
    
    \caption{Construction of the first CTF, $CTF_1$, in the SLCTF for the example with $mcs_p$ set to 4. Incoming arrows indicate inputs to the functions and outgoing arrows indicate the output. At each step, the variables that are ready for addition to the CTF (active nodes) are shown in red boxes. Addition of variables $p,q$ yields a CTF with max-clique size is greater than $mcs_p$ (set to 4) and are thus deferred. The interface variables are marked in blue.}
    \label{fig:build1}
\end{figure}
In this section, we describe the methods used to incrementally construct clique trees. We first explain the overall procedure using the running example shown in Figure~\ref{fig:ex-BN}. Figure~\ref{fig:build1} illustrates the steps used by $BuildCTF$ (Algorithm~\ref{alg:BuildCTF}) to construct the first CTF ($CTF_1$) in the SLCTF for the example.
Recall that the primary inputs (PIs) to the BN are variables that do not have any parents (see Definitions~\ref{def:pi} and~\ref{def:parent}, section~\ref{sec:background}). For our example, the PIs are $a,b,c$ and $d$. As shown in the figure, the CTF is first initialized using these inputs. A node is ready for addition to the CTF if its parents are already present in the CTF. We call these nodes the \textit{active nodes}. The active nodes in each step are shown in red boxes in the figure. Once the PIs are added, the active nodes are $e$ and $f$. In the next step, these two nodes are added to the CTF using the function $ModifyCTF$. Subsequently, $g$ becomes active and is added to the CTF. This process is continued until the maximum clique size reaches $mcs_p$, which is $4$ in this case. Since the addition of variables $p$ and $q$ is not possible without increasing the clique size beyond $mcs_p$ (set to 4), they are deferred for addition to the next CTF in the sequence. The parents of $p$ and $q$ ($f,o,l$ and $k$) are the interface variables (IV, see Definition~\ref{def:iv}, section~\ref{sec:definitions}). They are highlighted in blue. 
$CTF_1$ is the first CTF in the sequence. 
Subsequent CTFs are built in a similar manner, except that the starting point for the incremental construction is the approximated CTF.

\subsubsection{Incremental construction of CTF}\label{sec:incrMod}
We propose a novel technique for addition of new variables to an existing CTF. For clarity, we use the running example to explain how the method works when a single variable $v$ is added to the CTF. The addition of a set of variables is a simple extension.

Consider the addition of a new variable $v$ whose parents $pa_v=p_1, \hdots, p_m$ are already present in the existing CTF. This will introduce moralizing edges between all pairs of parent variables, resulting in an additional clique $C_{v}$ containing the variables $[p_{1},\hdots p_m, v]$ and an associated factor $P(v|p_1, \hdots p_m)$. 
The existing CTF now needs to be modified so that $C_v$ can be added to it while ensuring that the CTF remains valid.
The portion of the CTF that is affected by the addition of $C_v$ is the minimal subgraph that connects the cliques containing the parent variables ($MSG[Pa_v]$, see Definition~\ref{def:msg}). We denote this subgraph as $SG_{min}$.
Depending on the location of the parents, $SG_{min}$ can either be a connected subtree or a set of disjoint subtrees. 
Based on this, we identify three possible cases: Case~1: $SG_{min}$ has a single node, Case~2: $SG_{min}$ is a set of disconnected cliques and Case~3: $SG_{min}$ is fully or partially connected.
Let $ST'$ denote the modified subgraph obtained after addition of the new variable.

\begin{figure}
	\small
	\centering
	\captionsetup[subfigure]{position=b}
    \begin{subfigure}[t]{0.15\textwidth}
        \vskip 0pt
		\centering
        \begin{tikzpicture}[thick,scale=0.8, every node/.style={transform shape}]
            \foreach \pos/\name/\label in {(0,-1)/{\bf $\mathbf{v}$}/av, (0,-2)/{$f~(Case~1)$}/vf, (0,-3.5)/{$g~(Case~2)$}/vg}
                \node (\label) at \pos {\name};
        \end{tikzpicture}
    \end{subfigure}%
    \hfill
    \begin{subfigure}[t]{0.07\textwidth}
        \vskip 0pt
		\centering
        \begin{tikzpicture}[thick,scale=0.8, every node/.style={transform shape}]
            \foreach \pos/\name/\label in {(0,-1)/{$\mathbf{Pa_v}$}/av, (0,-2)/{$a,b$}/vf, (0,-3.5)/{$f,d$}/vg}
                \node (\label) at \pos {\name};
        \end{tikzpicture}
    \end{subfigure}%
    \hfill
    \begin{subfigure}[t]{0.25\textwidth}
        \vskip 0pt
		\centering
        \begin{tikzpicture}[thick,scale=0.8, every node/.style={transform shape}]
            \node (sgmin) at (1,-1) {\bf Existing $\mathbf{CTF}$};
            \foreach \pos/\name/\label in {(0,-2)/{$\boldsymbol{\color{blue}ab}e$}/abe1, (1.25,-2)/{$c$}/c, (2.5,-2)/$d$/d,  (0,-5)/{$ab\boldsymbol{\color{blue}f}$}/abf, (2.5,-3.5)/{$\boldsymbol{\color{blue}d}$}/d, (0, -3.5)/{$abe$}/abe2, (1.25,-3.5)/{$c$}/c2}
                \node [vertexr] (\label) at \pos {\name};
            \draw[edge2] (abe2)--node[midway,right]{\color{gray}$ab$}(abf);
        \end{tikzpicture}
    \end{subfigure}%
    \hfill
    \begin{subfigure}[t]{0.15\textwidth}
        \vskip 0pt
		\centering
        \begin{tikzpicture}[thick,scale=0.8, every node/.style={transform shape}]
            \node (sgmin) at (0,-1) {$\mathbf{SG_{min}}$};
            \foreach \pos/\name/\label in {(0,-2)/{$abe$}/abe, (0,-3.5)/{$abf$}/abf, (1.25,-3.5)/{$d$}/d}
                \node [vertexr] (\label) at \pos {\name};
        \end{tikzpicture}
    \end{subfigure}%
    \hfill
    \begin{subfigure}[t]{0.38\textwidth}
        \vskip 0pt
		\centering
        \begin{tikzpicture}[thick,scale=0.8, every node/.style={transform shape}]
            \node (stp) at (0,-1) {$\mathbf{ST'}$};
            \foreach \pos/\name/\label in {(0,-2)/{$abe$}/abe1, (1.5,-2)/{$abf$}/abf1, (0,-3.5)/{$abf$}/abf2, (1.5,-3.5)/{$d$}/d2, (0,-5)/{$fdg$}/fdg2, (5,-3.5)/{$abf$}/abf3, (5,-5)/{$fdg$}/fdg3}
                \node [vertexr] (\label) at \pos {\name};
            \foreach \source/\dest/\sepset  in {abe1/abf1/ab}
                \path[edge2] (\source) --node[midway,above]{\color{gray}\sepset} (\dest);
            \foreach \source/\dest/\sepset  in {abf2/fdg2/f, abf3/fdg3/f}
                \path[edge2] (\source) --node[midway,left]{\color{gray}\sepset} (\dest);
            \foreach \source/\dest/\sepset  in {fdg2/d2/d}
                \path[edge2] (\source) --node[midway,right]{\color{gray}\sepset} (\dest);
            \node (ds) [left=0.1 of abe1] {$C$};
            \node (ds) [right=0.1 of abf1] {$C_v$};
            \node (ds) [below=0.1 of d2] {$C'$};
            \node (ds) [right=0.1 of fdg3] {$C_v$};
            \node (ds) [right=0.1 of fdg2] {$C_v$};
            \draw[edge5] ([xshift=0.5cm]d2.east)--node[above, midway]{$Remove$} node [below, midway]{$non\mbox{-}maximal$}  node [below, midway, yshift=-0.35cm]{$clique, ~C'$}([xshift=-0.5cm]abf3.west);
        \end{tikzpicture}
    \end{subfigure}%
    \caption{Addition of a variable $v$ to an existing CTF for two cases. 
    $SG_{min}$ is the minimal subgraph connecting parents of variable $v$ ($Pa_v$) and $ST'$ is the modified subtree that replaces $SG_{min}$.
    Case 1: Both parents present in same clique, Case 2: Parents are present in disconnected cliques.
    }
    \label{fig:mCF12}
\end{figure}
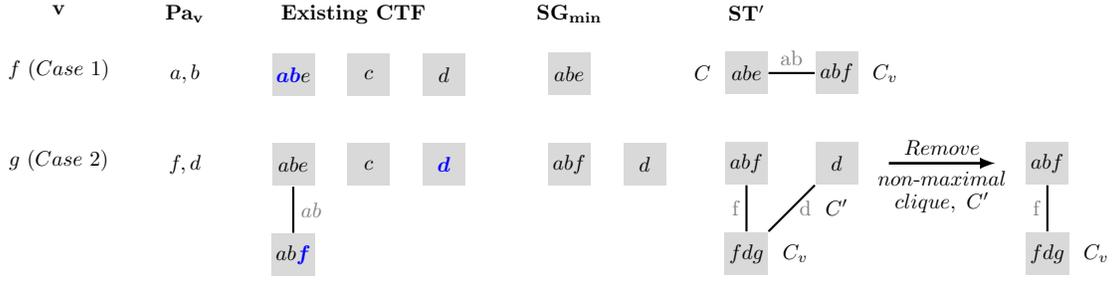
We first explain how the three cases are handled using the running example (refer to Figure~\ref{fig:build1} for the BN).
Figure~\ref{fig:mCF12} shows the addition of variables $f$ and $g$, corresponding to Cases 1 and 2 respectively. Consider addition of the variable $f$ to the CTF. The new clique $C_v$ contains $f$ and its parents, $a$ and $b$. Since both parents are present in the same clique, $SG_{min}$ is a single clique $C$ containing variables $a$, $b$ and $e$. $ST'$ is formed by connecting $C_v$ to $C$, with sepset containing the parents. When $g$ is added (Case2), the parents $f$ and $d$ belong to disjoint cliques. Therefore, $SG_{min}$ has the two cliques containing the parents. The new clique $C_v$ connects the two cliques. The result has a non-maximal clique $C'$ which has a single variable $d$ that is contained in $C_v$. $ST'$ is obtained after removing $C'$. The factor associated with $C'$ is re-assigned to $C_v$. 

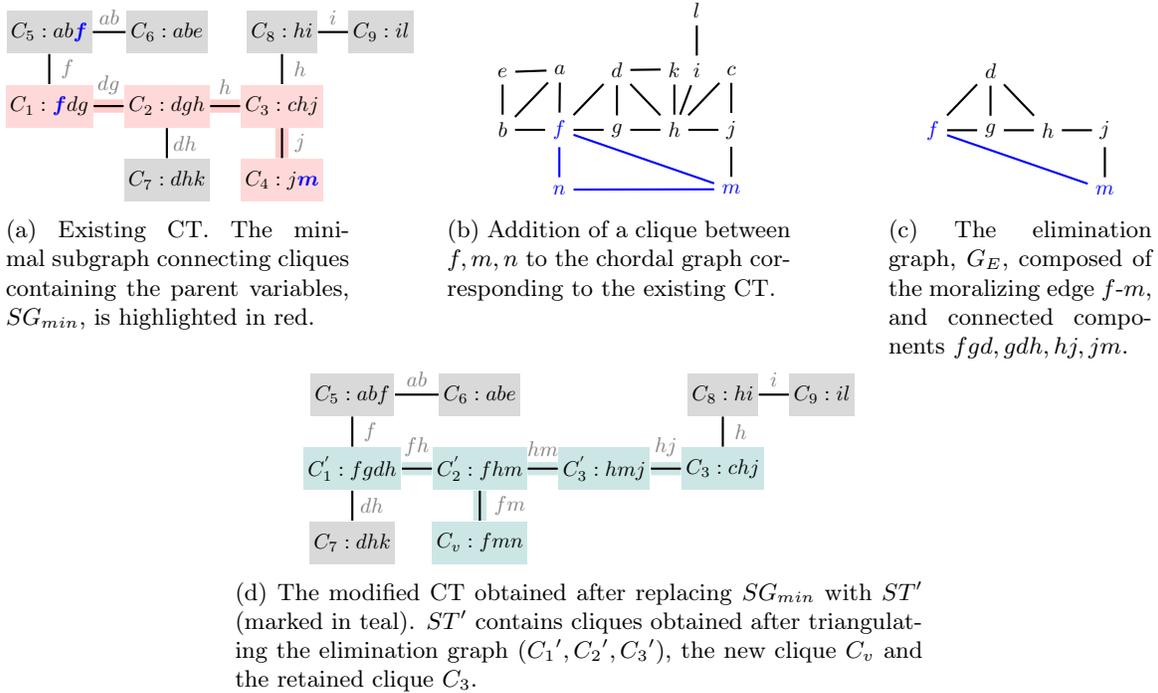
\begin{figure}[t]
	\small
	\centering
	\captionsetup[subfigure]{position=b}
	\begin{subfigure}[t]{0.3\textwidth}
		\centering
		\begin{tikzpicture}[thick, scale=0.8, every node/.style={transform shape}]
		\node (c1) [selected vertex] {$C_1: {\color{blue}\boldsymbol{f}}dg$};
		\node (c2) [selected vertex, right=0.5 of c1] {$C_2: dgh$};
		\node (c3) [selected vertex, right=0.5 of c2] {$C_3: chj$};
		\node (c4) [selected vertex, below=0.5 of c3] {$C_4: j{\color{blue}\boldsymbol{m}}$};
		\node (c5) [vertexr, above=0.5 of c1] {$C_5: ab{\color{blue}\boldsymbol{f}}$};
		\node (c6) [vertexr, above=0.5 of c2] {$C_6: abe$};
		\node (c7) [vertexr, below=0.5 of c2] {$C_7: dhk$};
		\node (c8) [vertexr, above=0.5 of c3] {$C_8: hi$};
		\node (c9) [vertexr, right=0.5 of c8] {$C_9: il$};
		\node (dummy)[below=0.7 of c2]{};
		\draw [loop1] (c1)--(c2) node[midway, above]{$\color{gray}dg$};
		\draw [loop1] (c2)--(c3)node[midway, above]{$\color{gray}h$};
		\draw [loop1] (c3)--(c4)node[midway, xshift=0.3cm]{$\color{gray}j$};
		\draw (c5)--(c1) node [midway, xshift=0.3cm] {${ \color{gray} f}$};
		\draw (c5)--(c6) node [midway, above] {${ \color{gray}ab}$};
		\draw (c8)--(c3) node [midway, xshift=0.3cm] {${ \color{gray}h }$};
		\draw (c8)--(c9) node [above,midway] {${ \color{gray}i }$};
		\draw (c1)--(c2);
		\draw (c2)--(c3);
		\draw (c2)--(c7) node [midway, xshift=0.3cm] {${ \color{gray}dh }$};
		\draw (c3)--(c4);
		\end{tikzpicture}
        \caption{Existing CT. The minimal subgraph connecting cliques containing the parent variables, $SG_{min}$, is highlighted in red.}
		\label{fig:mCF3_org}
	\end{subfigure}\hfill
	\begin{subfigure}[t]{0.3\textwidth}
		\centering
		\begin{tikzpicture}[thick, scale=0.8, every node/.style={transform shape}]
		\node (f) [align=center] {$\color{blue}f$};
		\node (g) [align=center, right=0.5 of f] {$g$};
		\node (h) [align=center, right=0.5 of g] {$h$};
		\node (j) [align=center, right=0.5 of h] {$j$};
		\node (m) [align=center, below=0.5 of j] {$\color{blue}m$};
		\node (n)[align=center, below=0.5 of f] {$\color{blue}n$};
		\node (d) [align=center, above=0.5 of g] {$d$};
		\node (a) [align=center, left=0.5 of d] {$a$};
		\node (b) [align=center, left=0.5 of f] {$b$};
		\node (c) [align=center, above=0.5 of j] {$c$};
		\node (i) [align=center, left=0.175 of c] {$i$};
		\node (l) [align=center, above=0.5 of i] {$l$};
		\node (k) [align=center, above=0.5 of h] {$k$};
		\node (e) [align=center, above=0.5 of b] {$e$};
        \foreach \source/\dest in {a/b, b/e, a/e, a/f, b/f, f/d, d/h, h/j, j/m, h/c, c/j, h/i, h/g, f/g, d/g, k/d, k/h, i/l}
            \draw (\source) -- (\dest);
        \foreach \source/\dest in { f/m, f/n, m/n}
            \draw[blue] (\source) -- (\dest);
		
		\end{tikzpicture}		
		\caption{Addition of a clique between $f,m,n$ to the chordal graph corresponding to the existing CT.
		}
		\label{fig:mCF3_cg}
	\end{subfigure}\hfill
	\begin{subfigure}[t]{0.23\textwidth}
		\centering
		\begin{tikzpicture}[thick, scale=0.8, every node/.style={transform shape}]
		\node (f) [align=center] {$\color{blue}f$};
		\node (g) [align=center, right=0.5 of f] {$g$};
		\node (h) [align=center, right=0.5 of g] {$h$};
		\node (j) [align=center, right=0.5 of h] {$j$};
		\node (m) [align=center, below=0.5 of j] {$\color{blue}m$};
        \node (d) [align=center, above=0.5 of g] {$d$};
        \foreach \source/\dest in {f/d, d/h, h/j, j/m, g/f, g/d, g/h}
            \draw (\source) -- (\dest);
		\draw [blue](f)--(m);
		\end{tikzpicture}		
        \caption{The elimination graph, $G_E$, composed of the moralizing edge $f\mbox{-}m$, and connected components $fgd, gdh, hj, jm$.}
		\label{fig:mCF3_eg}
	\end{subfigure}
	\begin{subfigure}[t]{0.6\textwidth}
		\centering
		\begin{tikzpicture}[thick, scale=0.8, every node/.style={transform shape}]
		\node (c1p) [selected vertext] {$C_1^{'} : fgdh $};
		\node (c2p) [selected vertext,right=0.5 of c1p] {$C_2^{'} : fhm$};
		\node (c3p) [selected vertext, right=0.5 of c2p] {$C_3^{'} : hmj$};	
		\node (c3) [selected vertext, right=0.5 of c3p] {$C_3: chj$};	
		\node (cn) [selected vertext, below =0.5 of c2p] {$C_v: f m n$};
		\node (c5) [vertexr, above=0.5 of c1p] {$C_5: abf$};
		\node (c6) [vertexr, above=0.5 of c2p] {$C_6: abe$};
		\node (c8) [vertexr, above=0.5 of c3] {$C_8: hi$};
		\node (c9) [vertexr, right=0.5 of c8] {$C_9: il$};
		\node (c7) [vertexr, below=0.5 of c1p] {$C_7: dhk$};
		\draw [loop2] (c1p)--(c2p) node [midway, above] {${{\color{gray}fh}}$};
		\draw [loop2] (c2p)--(c3p) node [midway, above] {${ \color{gray}hm}$};
		\draw [loop2] (c3p)--(c3) node [midway, above] {${ \color{gray}hj}$};
		\draw (c5)--(c1p) node [midway, xshift=0.3cm] {${\color{gray} f}$};
		\draw (c5)--(c6) node [midway, above] {${ \color{gray}ab}$};
		\draw (c8)--(c3) node [midway, xshift=0.3cm] {${ \color{gray}h }$};
        \draw (c8)--(c9) node [above,midway] {${ \color{gray}i }$};
        \draw (c7)--(c1p) node [right,midway] {${ \color{gray}dh}$};
		\draw [loop2] (c2p)--(cn) node [midway, xshift=0.5cm] {${\color{gray} f m }$};
		\draw (c1p)--(c2p);
		\draw (c2p)--(c3p);
		\draw (c3p)--(c3);
		\draw (c2p)--(cn);
		\end{tikzpicture}
        \caption{The modified CT obtained after replacing $SG_{min}$ with $ST'$ (marked in teal). $ST'$ contains cliques obtained after triangulating the elimination graph (${C_1}',{C_2}',{C_3}'$), the new clique $C_v$ and the retained clique $C_3$.}
		\label{fig:mCF3_new}
	\end{subfigure}	
    \caption{\textit{\underline{Case 3}}: Addition of a new variable $n$, with parents ($f,m$) in connected cliques.}
	\label{fig:mCF3}
\end{figure} 
Figure~\ref{fig:mCF3} shows an example for Case~3 where the parents $f$ and $m$ of the new variable $n$, are present in connected cliques. 
The cliques shaded in red in Figure~\ref{fig:mCF3_org} form $SG_{min}$.  
The addition of variable $n$ results in a new clique $C_v$ containing variables $f,~m$ and $n$. 
The goal is to replace $SG_{min}$ with a modified subtree $ST'$ that contains $C_v$ while ensuring that CT remains valid.
As shown in Figure \ref{fig:mCF3_cg}, when the moralizing edge between the parents $f$ and $m$ is added to the chordal graph corresponding to the existing CT, chordless loops $f\mbox{-}g\mbox{-}h\mbox{-}j\mbox{-}m\mbox{-}f$ and $f\mbox{-}d\mbox{-}h\mbox{-}j\mbox{-}m\mbox{-}f$ are introduced. Therefore, retriangulation is needed to get back a chordal graph. 
However, only a subgraph of the modified chordal graph needs to be re-triangulated.
As shown in Figure~\ref{fig:mCF3_cg}, there are several fully connected components corresponding to cliques in the existing CT. Using variable elimination to form the cliques, it can be seen that the elimination order $O=[e,a,b,l,i,k,c]$ gives maximal cliques in set $S_O=\{C_6,C_5,C_9,C_8,C_7,C_3\}$ all of which are present in the existing CT. The subgraph $G_E$ shown in Figure~\ref{fig:mCF3_eg} is obtained after eliminating these variables and deleting the corresponding edges. 
This is the subgraph that needs re-triangulation. We call it the elimination graph.

Comparing Figures~\ref{fig:mCF3_org} and~\ref{fig:mCF3_eg}, it can be seen that $G_E$ contains only the parent variables of the node $n$ and variables in the sepsets of $SG_{min}$. This makes sense because the chordless loops are introduced by the moralizing edge between the parents. Therefore, any variable other than the parents and the sepset variables can be eliminated without introducing fill-in edges, which means that the resulting cliques will be present in the existing CT.
As seen in Figure~\ref{fig:mCF3_eg}, $G_E$ includes the moralizing edge between parents and  fully connected components between parent and sepset variables contained in each clique in $SG_{min}$.
On triangulating $G_E$, we get cliques ${C_1}',{C_2}'$ and ${C_3}'$ shown in Figure~\ref{fig:mCF3_new}. 
$C_v$ is connected to ${C_2}'$ since it contains both parent variables $f$ and $m$.
Amongst the cliques contained in set $S_O$, clique $C_3$ is also present in $SG_{min}$. It is obtained after elimination of variable $c$, which is neither a parent nor a sepset variable. We call such cliques as \textit{retained cliques}. 
We connect $C_3$ to clique ${C_3}'$ in ${ST'}$ via the sepset variables $h$ and $j$.

Finally, $ST'$ shown in teal in Figure~\ref{fig:mCF3_new}, contains the new clique $C_v$, the retained clique $C_3$ and the cliques obtained after triangulating $G_E$.
$ST'$ replaces $SG_{min}$ in the existing CT. The connection is done via cliques $C_5,C_7$ and $C_8$ that were adjacent to $SG_{min}$ with the same sepsets.
Since cliques $C_1, C_2, C_4$ are no longer present in the modified CT, the associated factors are re-assigned to corresponding containing cliques in $ST'$. The factors associated with $C_1$ and $C_2$ are re-assigned to $C_1^{'}$ and that associated with $C_4$ is re-assigned to $C_3^{'}$.

The formal steps in the procedure to add a variable $v$ with corresponding clique $C_v=\{v, Pa_v\}$ are as follows.
Based on the minimal subgraph connecting the cliques containing the parent variables $SG_{min}$, we identify three possible cases.\\
\textit{\underline{Case 1}:} $SG_{min}$ has a single node. This will happen if all parents are contained in the same clique $C$. In this case, we connect $C_v$ to $C$.
In case $C$ is contained in $C_v$, $C$ is replaced by $C_v$ and the factor associated by $C$ is re-assigned to $C_v$.
\\ 
\textit{\underline{Case 2}:} $SG_{min}$ is a set of disconnected cliques. If the parents belong to disconnected cliques $C_1, ..,C_j$ in CTF, $C_v$ gets connected to each of the cliques. Any non-maximal clique in $\{C_1, ..,C_j\}$ is removed and its neighbors are connected to $C_v$ and factors are re-assigned.
\\ 
\textit{\underline{Case 3}:} $SG_{min}$ is fully or partially connected. 
We use the following steps to first obtain the modified subtree $ST'$ (Steps
1-3) and then replace $SG_{min}$ by $ST'$ to obtain the modified CTF (Steps 4-5).
\begin{enumerate} [leftmargin=\widthof{[Step~3]}+\labelsep]
    \item [Step~1:] Construct a set $S$ containing parents of the new variable and all variables in the sepsets in $SG_{min}$.
	\item[Step~2:] Find cliques in $SG_{min}$ that contain variables that do not belong to set $S$.
	These cliques are called \textit{retained cliques}.
	\item[Step~3:] Find the modified clique tree $ST'$ using the following steps. \label{step:constructSTp}
    \begin{enumerate} [leftmargin=\widthof{[St]}+\labelsep]
	\item[Step~3.1:]  Construct a graph $G_E$ as follows. The nodes of the graph are the variables contained in the set $S$. The edges are obtained as follows.
	For each clique $C$ in $SG_{min}$, a fully connected component corresponding to the variable set $C \cap S$ is added to $G_E$.  Finally,  moralizing edges are introduced between the parent variables. We call this graph the {\it elimination graph}. \label{step:elimGraph}
    \item[Step~3.2:] Triangulate $G_E$ and obtain the modified clique tree $ST'$. Note that the $ST'$ is not unique and depends on the elimination order used for re-triangulation.\label{step:triangElimGraph}
    \item[Step~3.3:] Connect the new clique $C_v$ to the clique $C'$ in $ST'$ that contains all parents. If $C'$ is a subset of $C_v$, replace $C'$ with $C_v$ and re-assign factors.\label{step:connectNew}
	\item[Step~3.4:] For each retained clique $C$, identify clique $C'$ in $ST'$ that contains $C\cap S$.
	Connect $C$ to $C'$.
    If $C'$ is a subset of $C$, replace $C'$ with $C$ and re-assign factors. \label{step:connectRetained}
    \item[Step~3.5:] The factors associated with cliques that are not retained are re-assigned to cliques in $ST'$ containing the entire scope of factors.  
	\end{enumerate}
	\item[Step~4:] Remove the impacted subgraph $SG_{min}$ from CTF.\label{step:removeST}
	\item[Step~5:] Connect $ST'$ to CTF via the set of cliques adjacent to $SG_{min}$ in the existing CTF.
	Cliques in this adjacency set are reconnected to cliques in $ST^{'}$ that contain the corresponding sepset in the input CTF. 
        \label{step:connectAdj}
	\end{enumerate}
The three cases described above show how a single active variable is added to the existing CTF. 
In the first two cases, the size of the existing cliques does not change and variables corresponding to these cases are added sequentially.  
Although sequential addition can also be done for variables corresponding to Case 3, it is more efficient to add variables that have overlapping minimal subgraphs together since it avoids repeated formation and retriangulation of the same subgraph.
Therefore, we group the new variables into subsets such that each subset contains variables with overlapping minimal subgraphs. All variables in a subset are added together.
The set $S$ will now contain parents of all variables in the subset as well as all the sepsets in the impacted subtree.
\begin{algorithm}[!h]
	\scriptsize
    \caption{BuildCTF~($CTF,G,S_{eg},mcs_p$)}
	\label{alg:BuildCTF}
	\begin{algorithmic}[1]
		\Require ~$CTF$: Existing CTF \newline
		\indent$G$: Input DAG\newline
		\indent$S_{eg}$: Set of evidence variables\newline	
		\indent$mcs_{p}$: Maximum clique size limit for $CTF$ 
		\Ensure $CTF$: Modified $CTF$\newline
		\indent$IV$: Set of interface variables\newline
        \indent$G$: Modified input DAG\newline
        \indent$S_{eg}$: Modified set of evidence variables
        \If {$CTF.isEmpty()$}
                \State $PI\gets$ Primary inputs in $G$
                \State $CTF$ = \{Disjoint cliques over $PI$\} \Comment{{\color{teal!70} \scriptsize Clique Tree Forest}}
                \State $G$.remove($PI$) \Comment{{\color{teal!70} \scriptsize Remove PIs and their outgoing edges from $G$}}
                \State $S_{eg}$.remove($PI\cap S_{eg}$) \Comment{{\color{teal!70} \scriptsize Remove evidence variables added to $CTF$}}
        \EndIf
	    \State $N_a \gets \varnothing$ \Comment{{\color{teal!70} \scriptsize  Set of variables added to $CTF$}} 
		\State $IV\gets \varnothing$ \Comment{{\color{teal!70} \scriptsize  Interface variables}}
        \State $N_{act} \gets $ \{Variables in $G$ with zero in-degree\}  \Comment{{\color{teal!70} \scriptsize  Active variables}} 
		\While{ $N_{act}\neq \varnothing$}
		\State $N_i \gets N_{act}$.pop variables with min. topological level
		\State $CTF, N_a\gets$ ModifyCTF ($CTF, ~N_i, ~mcs_p,~S_{eg}$) \Comment{{\color{teal!70} \scriptsize  $N_a$ is the subset of variables in $N_i$ that are added to $CTF$}}
        \State $G$.remove($N_a$) \Comment{{\color{teal!70} \scriptsize Delete $N_a$ and outgoing edges in G}}
		\State $S_{eg}$.remove($N_a\cap ~S_{eg}$) \Comment{{\color{teal!70} \scriptsize Remove evidence variables that are added to the $CTF$}}
		\State $IV$.add parents of variables in $N_i\setminus N_a$  \Comment{{\color{teal!70} \scriptsize Update interface variables}}
        \State $N_f \gets$ Children of $N_a$ with zero in-degree in $G$ \Comment{{\color{teal!70} Identify new active variables}}
		\State $N_{act}$.push($N_f$) \Comment{{\color{teal!70} \scriptsize Update the list $N_{act}$}}
		\EndWhile		
        \State \Return $CTF,~IV,~G,~S_{eg}$
	\end{algorithmic}
\end{algorithm}

\noindent\textbf{Algorithm $BuildCTF$}\\
Inputs to this function (Algorithm~\ref{alg:BuildCTF}) are an existing CTF to which variables are to be added, the DAG $G$, the set of evidence variables in $G$ ($S_{eg}$) and the maximum clique size bound $mcs_p$. $BuildCTF$ incrementally adds variables to the input CTF until the clique size bound is reached. The main steps are as follows. If it is the first CTF in the sequence, the CTF is initialized as disjoint cliques containing the primary inputs. Every time a variable is added to a CTF, it is removed from $G$ and also from $S_{eg}$ if it is an evidence variable. 
Therefore, if a node has zero in-degree in $G$, it means that all its parents have been added to the CTF and it is an active node. All active nodes are identified and stored in the list $N_{act}$. 
As long as there are nodes in the list, $BuildCTF$ attempts to add it to the CTF. The set of active nodes is divided into subsets based on the topological level. The function $ModifyCTF$ adds each subset to the CTF, provided the clique size does not exceed $mcs_p$. 
The parents of nodes that could not be added to the CTF without violating clique size bounds are included in the list of interface variables, $IV$.
\begin{algorithm}[!h]
\scriptsize
	\caption{ModifyCTF~($CTF, ~N_i,~mcs_p,~S_{eg}$)}
	\label{alg:modifyCTF}
	\begin{algorithmic}[1]
		\Require ~$CTF$: Existing CTF \newline
		\indent$N_i$: Set of active variables to be added\newline
		\indent$mcs_{p}$: Maximum clique size limit for $CTF$ \newline
		\indent$S_{eg}$: Set of evidence variables	
		\Ensure $CTF$: Modified $CTF$\newline
		\indent~$N_a$: Subset of $N_i$ added to $CTF$
		\State \textbf{Initialize:} $N_a = [~]$
        \LineComment{{\color{teal!70} First add variables whose parents are contained in the same clique (Case~1)  or disconnected cliques (Case~2).}}
		\For {$v \in N_i$}
            \State $Pa_v\gets $ Parents of $v$ in BN
            \State $SG_{min} \gets MSG[Pa_v]$ \Comment{{\color{teal!70} Minimal subgraph connecting variables in $Pa_v$}}
            \If{$SG_{min}$ corresponds to Case~1 or Case~2}
                \State Connect $C_v=\{v,Pa_v\}$ to parent cliques; Check Maximality
                \State $N_a$.add($v$)
			\State $N_i$.remove($v$)
            \EndIf 
            	\EndFor
		\LineComment{{\color{teal!70} Add remaining Case 3 variables after grouping variables with connected overlapping subgraphs}}
        \State $L_{grp}\gets $ Group variables in $N_i$ with overlapping $SG_{min}$
        \While{$L_{grp}\neq \varnothing$}
        \State $N_g = L_{grp}.$pop()
		\State $SG_{min}\gets$ $CTF$.subgraph($\cup_{v\in N_g} MSG[Pa_v]$) 
        \Comment{{\color{teal!70} Find the union of the overlapping subgraphs}}
        \State $ST'\gets$ Find modified subtree by adding variables in $N_g$ using steps 1-3 in Section~\ref{sec:incrMod}
		\LineComment{{\color{teal!70}  Add variables if clique size constraints are met; else choose a subset for addition}}
		\If {$ST'.$max\_clique\_size$ ~\leq mcs_p$}
        \State $CTF \gets$ Replace $SG_{min}$ in CTF by $ST'$ using steps 4-5 in Section~\ref{sec:incrMod} 
		\State $N_a.add(N_g)$
		\Else
		\State $N_s = $ Choose a subset of $N_g$ for addition; Prioritize $N_g\cap S_{eg}$
        \State Group variables in $N_s$ with overlapping subgraphs and add to $L_{grp}$
		\EndIf
		\EndWhile
		\State \Return{$CTF, N_a$}
	\end{algorithmic}
\end{algorithm}
\\
\textbf{Algorithm $ModifyCTF$}\\
Procedure $ModifyCTF$ (Algorithm \ref{alg:modifyCTF}) shows the main steps involved in modifying the CTF to add a set of active variables under a given maximum clique size constraint $mcs_p$.
Corresponding to each new variable $v$, we identify the minimal subgraph that connects the cliques containing the parent variables, $SG_{min}=MSG[Pa_v]$ (see Definition \ref{def:msg}).
It is the minimal subgraph that is impacted by the addition of the new variable. 
We sequentially add variables belonging to Cases 1 and 2.
Variables belonging to Case 3 are grouped into subsets such that variables in each subset have some overlap in the corresponding $SG_{min}$.   These subsets are added to the list $L_{grp}$. All variables in a subset $N_g$ are added together. 
The overall $SG_{min}$ is the union of the subgraphs corresponding to all variables in $N_g$. The modified subtree $ST'$ is obtained using Steps~1-3 described previously. 
If the maximum clique size in $ST'$ is less than $mcs_p$, $SG_{min}$ is replaced by $ST'$ using Steps~4-5.
Otherwise, we choose a smaller subset $N_s$ for addition and defer the remaining variables for addition to the next CTF in the sequence. We prioritize evidence variables while choosing variables in $N_s$. The reason for this is discussed in more detail in Section~\ref{sec:pm}. Variables in $N_s$ are re-grouped based on overlapping subgraphs, and the new groups are added to $L_{grp}$.
The iteration ends when $L_{grp}$ becomes empty, that is, no variable can be added without violating the clique size bound.

\subsubsection{Soundness of the algorithm}
$BuildCTF$ (Algorithm \ref{alg:BuildCTF}) repeatedly uses function $ModifyCTF$ (Algorithm \ref{alg:modifyCTF}) to incrementally add variables. Therefore, if Algorithm~\ref{alg:modifyCTF} returns a valid CTF, Algorithm~\ref{alg:BuildCTF} will also return a valid CTF. The following propositions show that if the input CTF to Algorithm \ref{alg:modifyCTF} is valid, it returns a valid CTF that satisfies the properties in Definition \ref{def:vct}.
The proofs for these propositions are included in Appendix \ref{app:proofs}.

\begin{proposition}\label{pr:pr1}
    The modified CTF obtained using $ModifyCTF$ (Algorithm \ref{alg:modifyCTF}) contains (possibly disjoint) trees i.e., no loops are introduced by the algorithm.
\end{proposition}

\begin{proposition} \label{pr:pr2} After addition of a variable $v$, the modified CTF contains only maximal cliques. 
\end{proposition}
\begin{proposition} \label{pr:pr4} All CTs in modified CTF satisfy the running intersection property (RIP).
\end{proposition}
\begin{proposition}\label{pr:pr3} Product of factors in the modified CTF gives the correct joint distribution.
\end{proposition}
\begin{theorem}\label{thm:thm1}
    The CTF constructed by Algorithm~\ref{alg:BuildCTF} is a valid CTF.
  \begin{proof}
      Algorithm~\ref{alg:BuildCTF} starts the construction starts with a set of disjoint cliques corresponding to the PIs, which is a valid CTF. This is the first input to the function $ModifyCTF$.
    Based on Propositions \ref{pr:pr1} - \ref{pr:pr4}, if the input is a valid CTF, the modified CTF built using  Algorithm~\ref{alg:modifyCTF} is also a valid CTF since it satisfies all the properties needed to ensure that the CTF contains a set of valid CTs.
   \end{proof}
\end{theorem}

\subsection{CalibrateCTF: Algorithm to infer clique beliefs}
$BuildCTF$ (Algorithm \ref{alg:BuildCTF}) returns a CTF in which one or more cliques have the maximum allowed size $mcs_p$.  Let this CTF be denoted as $CTF_{in}$. 
We wish to do a factor based approximation of $CTF_{in}$ that is based on the clique beliefs and not on the structure of the CTF or the BN. In order to do this, we have an \textit{infer} phase (line 4, Algorithm~\ref{alg:cSLCTF}) in which the function $CalibrateCTF$ calibrates $CTF_{in}$ using the standard belief propagation algorithm for exact inference~\cite{Lauritzen1988,Koller2009}. The algorithm performs two rounds of message passing, after which the clique and sepset beliefs ($\beta$ and $\mu$) beliefs are available.
In the presence of evidence, the BP algorithm uses the un-normalized clique beliefs (see Definition~\ref{def:pr}, section~\ref{sec:background}). After calibration, the  normalization constant of all clique and sepset beliefs are identical.
The beliefs from the calibrated $CTF_{in}$ are used for the approximation. 
\subsection{ApproximateCTF: Algorithm to approximate the CTF}
\label{sec:interfaceModel}
The next phase is the approximate phase. The inputs are a calibrated CTF, denoted $CTF_{in}$ and the set of interface variables ($IV$, see Definition~\ref{def:iv}) in $CTF_{in}$.  
Variables in $CTF_{in}$ that are not interface variables are denoted as {\it{non-interface}} variables (NIV).
In this step, the goal is to obtain a CTF, denoted $CTF_a$, in which the maximum clique size is $mcs_{im}$, which is lower than $mcs_p$. This allows for addition of new variables to $CTF_a$ to form the next CTF in the sequence.
Since the new variables to be added are all successors of variables in $IV$, $CTF_a$ must contain all variables in the set. The accuracy of the joint beliefs of the new variables depends on the accuracy of the joint beliefs of the interface variables in $CTF_a$. Ideally, we would like $CTF_a$ to be a structure that contains only the interface variables and preserves their joint beliefs exactly. 

In the following subsections, we discuss our approximation strategy, the metrics and heuristics used for the approximation and the algorithm.
For clarity, we explain the steps assuming that the clique sizes can be reduced to exactly $mcs_{im}$. In practice, it could be larger or smaller depending on the size of the CPDs of the variables that are removed.

\subsubsection{Approximation Strategy}\label{sec:approxStrat}

Our strategy to obtain $CTF_a$ is to reduce the size of cliques in $CTF_{in}$ by removing some variables and marginalizing the clique beliefs by summing over the states of these variables. An \textit{exact marginalization} of the clique beliefs over the non-interface variables will preserve the joint distribution of the interface variables. 
However, exact marginalization corresponding to a variable can only be done after collapsing all the cliques containing the variable into a single clique, finding the joint belief of the collapsed clique and then marginalizing the belief by summing over the states of the variable to be removed. This process becomes expensive or infeasible as the size of the collapsed clique increases.
An alternative is to do a \textit{local marginalization} in which variables are removed from the large sized cliques and individual clique beliefs are marginalized. This has to be done carefully so that the resulting CTs are valid CTs satisfying RIP. 

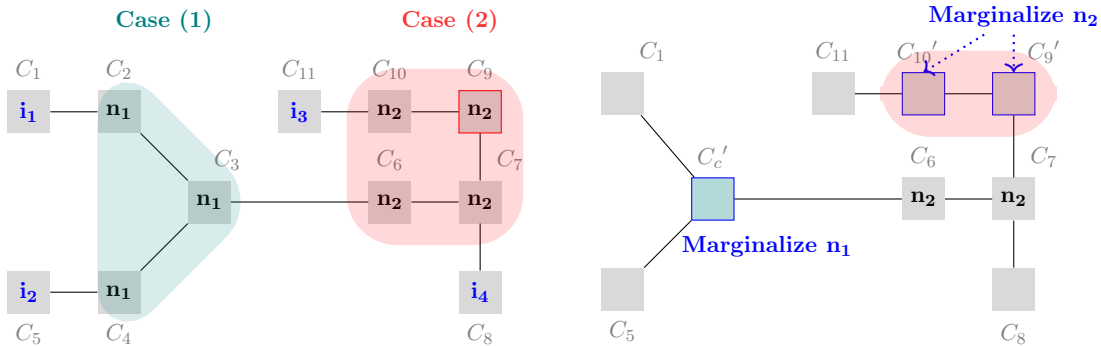
\begin{figure}[!htb]
	\centering
	\begin{subfigure}[t]{0.5\textwidth}
		\centering
	    \begin{tikzpicture}[scale=0.8, every node/.style={scale=0.8},highlightfill/.style={fill,rounded corners=1em, line width=1.5em,opacity=0.15,cap=round},]
	        \node[vertexr] (c2) at (0,0){\color{black}$\mathbf{n_1}$};
            \draw (c2)++(0.75,1.5) node{\bf \color{teal} Case (1)};
	        \node (d2) at (0,0.7){\color{gray}$C_2$};
            \node[vertexr] (c1) at (-1.5,0){\color{blue}$\mathbf{i_1}$};
	        \node (d1) at (-1.5,0.7){\color{gray}$C_1$};
	        \node[vertexr] (c3) at (1.5,-1.5){\color{black}$\mathbf{n_1}$};
	        \node (d3) at (1.8,-0.8){\color{gray}$C_3$};
	        \node[vertexr] (c4) at (0,-3){\color{black}$\mathbf{n_1}$};
	        \node (d4) at (0,-3.7){\color{gray}$C_4$};
            \node[vertexr] (c5) at (-1.5,-3){\color{blue}$\mathbf{i_2}$};
	        \node (d5) at (-1.5,-3.7){\color{gray}$C_5$};
	        \node[vertexr] (c7) at (4.5,-1.5){\color{black}$\mathbf{n_2}$};
	        \node (d6) at (4.5,-0.8){\color{gray}$C_6$};
	        \node[vertexr] (c8) at (6,-1.5){\color{black}$\mathbf{n_2}$};
	        \node (d6) at (6.5,-0.8){\color{gray}$C_7$};
            \node[vertexr] (c9) at (6,-3){\color{blue}$\mathbf{i_4}$};
	        \node (d6) at (6,-3.7){\color{gray}$C_8$};
	        \node[vertexr,draw=red] (c10) at (6,0){\color{black}$\mathbf{n_2}$};
	        \node (d6) at (6,0.7){\color{gray}$C_9$};
	        \node[vertexr] (c11) at (4.5,0){\color{black}$\mathbf{n_2}$};
            \draw (c11)++(1,1.5) node{\bf \color{red!80} Case (2)};
	        \node (d6) at (4.5,0.7){\color{gray}$C_{10}$};
            \node[vertexr] (c12) at (3,0){\color{blue}$\mathbf{i_3}$};
	        \node (d6) at (3,0.7){\color{gray}$C_{11}$};
	        \draw (c1)--(c2);
	        \draw (c2)--(c3);
	        \draw (c3)--(c4);
	        \draw (c4)--(c5);
	        \draw (c3)--(c7);
	        \draw (c7)--(c8);
	        \draw (c8)--(c9);
	        \draw (c8)--(c10);
	        \draw (c10)--(c11);
	        \draw (c11)--(c12);
	        \drawoutside[highlightfill, teal]{(c2.north) -- (c3.east) -- (c4.south)  -- cycle}
            \drawoutside[highlightfill, red]{(c11.north west) -- (c7.south west) -- (c8.south east) -- (c10.north east) -- cycle}
	    \end{tikzpicture}
        \caption{A CT containing interface variables $i_1,i_2,i_3,i_4$ and non-interface variables $n_1, n_2$. Cliques used for exact and local marginalization are highlighted in teal and red colors respectively. Clique size for $C_9$ is greater than $mcs_{im}$.}
	\label{fig:ctfin}
	\end{subfigure}
    \hfill%
	\begin{subfigure}[t]{0.48\textwidth}
		\centering
	    \begin{tikzpicture}[scale=0.8, every node/.style={scale=0.8},highlightfill/.style={fill,rounded corners=1em, line width=1.5em,opacity=0.15,cap=round}]
	        \node[vertexr, fill=teal!30, draw=blue] (cn2) at (0,-1.5) {};
	        \node (tn2) [align=center, below=0.1 of cn2, xshift=0.9cm] { \bf \color{blue} Marginalize  $\mathbf{n_1}$};
	        \node (d) at (0,-0.8){\color{gray}${C_{c}}'$};
	        \node[vertexr] (c1) at (-1.5,0.25){};
	        \node (d) at (-1,0.95){\color{gray}${C_{1}}$};
	        \node[vertexr] (c5) at (-1.5,-3){};
	        \node (d) at (-1.5,-3.7){\color{gray}${C_{5}}$};
	        \node[vertexr] (c7) at (3.5,-1.5){\color{black} $\mathbf{n_2}$};
	        \node (d) at (3.5,-0.8){\color{gray}${C_{6}}$};
	        \node[vertexr] (c8) at (5,-1.5){\color{black} $\mathbf{n_2}$};
	        \node (d) at (5.5,-0.8){\color{gray}${C_{7}}$};
	        \node[vertexr] (c9) at (5,-3){};
	        \node (d) at (5,-3.75){\color{gray}${C_{8}}$};
	        \node[vertexr] (c12) at (2,0.25){};
	        \node (d) at (2,0.95){\color{gray}${C_{11}}$};
	        \node[vertexr, draw=blue] (c10) at (5,0.25){};
	        \node (d) at (5.5,0.95){\color{gray}${C_{9}}'$};
	        \node[vertexr, draw=blue] (c11) at (3.5,0.25){};	
	        \node (d) at (3.4,0.95){\color{gray}${C_{10}}'$};
	        \draw (c1)--(cn2);
	        \draw (cn2)--(c5);
	        \draw (cn2)--(c7);
	        \draw (c7)--(c8);
	        \draw (c8)--(c9);
	        \draw (c8)--(c10);
	        \draw (c10)--(c11);
	        \draw (c11)--(c12);
	        \node (tm) [align=center,above=0.5 of c10] {\color{blue} \bf Marginalize $\mathbf{n_2}$};
	        \draw [->, dotted, blue, thick](tm)--(c10.north);
	        \draw [->, dotted, blue, thick](tm)--(c11.north);
            \drawoutside[highlightfill, red]{(c11.north west) -- (c11.south west) -- (c10.south east) -- (c10.north east) -- cycle}
	    \end{tikzpicture}
        \caption{Approximate CT. The collapsed clique generated after exact marginalization of $n_1$ is shown in teal and the modified cliques obtained after local marginalization of $n_2$ are highlighted in red.}
	\label{fig:ctfa}
	\end{subfigure}
	\caption{
        Illustration of exact and local marginalization. The relevant interface and non-interface variables are shown in each clique. 
    }
\label{fig:approxCTF}
\end{figure}

We use a combination of exact and local marginalization to obtain $CTF_a$. Exact marginalization is used whenever a variable is present in a single clique or the size of the collapsed cliques is at most $mcs_{im}$. For all other cases, local marginalization is used. The process is illustrated in Figure \ref{fig:approxCTF}.  In the figure, variables $i_1, \cdots i_4$ are interface variables and $n_1$ and $n_2$ are non-interface variables. 
Figure~\ref{fig:ctfin} shows the two cases. In Case (1) (highlighted in teal), the sizes of cliques $C_2$, $C_3$ and $C_4$ are such that they can be collapsed to a single clique $C_c$ with size less than or equal to $mcs_{im}$. Following this, a new clique ${C_c}'$ with reduced size is obtained from $C_c$ after removing $n_1$ and computing its belief as $\beta({C_c}') = \sum_{n_1.states}\beta(C_c)$.
In Case (2) (highlighted in red), we attempt to remove $n_2$ which is contained in $C_6,C_7,C_9$ and $C_{10}$. Assume that clique $C_9$ has size greater than $mcs_{im}$. A new clique $C_9'$ is obtained from $C_9$ by removing $n_2$ and setting  $\beta(C_9') = \sum_{n_2.states}\beta({C_{9}})$. Therefore, the joint beliefs of the remaining variables in $C_9'$ is preserved, but the joint beliefs of variables present in different cliques is not. We call this process local marginalization. However, the problem is that the clique tree is not a valid tree since $n_2$ is present in $C_6, C_7$ and $C_{10}$, but will not be present in the intermediate clique $C_9'$, violating RIP. 
To satisfy RIP, one possibility is to locally marginalize the variable from all cliques and sepsets in which it is present. However, the interface variables are present in many different cliques and our aim is to preserve their joint beliefs as much as possible. Therefore, we locally marginalize the variable $n_2$ from a minimum number of cliques so that the size constraint is not violated and RIP is satisfied. In the figure, $n_2$ is also removed from $C_{10}$ to give $C_{10}'$ with belief $\beta(C_{10}') = \sum_{n_2.states}\beta({C_{10}})$. It is also similarly removed from the sepset between $C_9$ and $C_{10}$. It is retained in $C_6$ and $C_7$.  
\tikzstyle{circ}=[
circle,
minimum size=0.75cm,
draw=black,
inner sep=1pt,
]

If we only want to reduce clique sizes, strictly speaking, exact marginalization is not needed.
However, exact marginalization reduces the number of cliques and non-interface variables in $CTF_a$, while preserving the joint beliefs exactly. This in turn reduces the computational effort involved in adding new variables to $CTF_a$ by reducing the number of cliques separating parents of variables to be added, leading to smaller elimination graphs.

More formally, our technique uses the following steps.
We initialize $CTF_a$ to $MSG[IV]$ (see Definition~\ref{def:msg}) which is the minimal subgraph that connects cliques containing the interface variables. This is followed by exact and local marginalization as described below. 
\begin{enumerate}
    \item[1.] \textit{Exact marginalization:} All cliques containing a variable $v$ are collapsed if the size of the collapsed clique is less than $mcs_{im}$. Let $ST_v$ be the subtree of $CTF_a$ that has all the cliques containing $v$. A new clique $C_c$ is obtained after collapsing all cliques in $ST_v$ and removing $v$. The clique belief for $C_c$ is obtained after marginalizing the joint probability distribution of $ST_v$ over the states of variable $v$, as follows.
\begin{equation*}
  \beta(C_c) = \sum_{v.states} \left ( \frac{\prod_{C\in {ST_v}} \beta(C)}{\prod_{SP \in {ST_v}}\mu(SP)} \right), \quad SP~ {\rm{~denotes~sepsets~in}} ~ST_v
\end{equation*}
This is continued until further collapsing is not possible without violating size constraints.
  Since the marginalization is exact, the joint distribution of the remaining variables in $CTF_a$ is preserved.
\item[2.] \textit{Local marginalization:} Cliques in $CTF_a$ with size greater than $mcs_{im}$ are identified and a variable is locally marginalized from the smallest possible subset of cliques and sepsets containing the variable, so that the resulting CT remains a valid CT satisfying RIP. If $v$ is the variable to be locally marginalized from two adjacent cliques $C_i$ and $C_j$ with sepset $S_{i,j}$, then local marginalization results in two approximate cliques $C_i' = C_i\setminus v$ and $C_j'=C_j\setminus v$ with sepset $S_{i,j}' = S_{i,j}\setminus v$. The corresponding beliefs are
        \begin{equation}\label{eqn:localMarg}
          \beta(C_i') = \sum\limits_{v.states}\beta(C_i), \quad \beta(C_j')  = \sum\limits_{v.states}\beta(C_j), \quad \mu(S_{i,j}') =  \sum\limits_{v.states}\mu(S_{i,j})
        \end{equation}
\end{enumerate}

\subsubsection{Choice of variables for local marginalization}\label{sec:choice}
Since our aim is to preserve the joint beliefs of the interface variables as much as possible, we would like to choose variables that have the least impact on this joint belief for local marginalization. 
We need a metric that measures this influence and is inexpensive to compute. 
Towards this end, we propose a heuristic technique based on pairwise mutual information (MI) between variables. The MI between two variables $x$ and $y$ is defined as 
\begin{equation*}
          MI(x;y) = \sum\limits_{s \in D_x, w \in D_y} p(s,w) \log \frac{p(s,w)}{p(s)p(w)}
\end{equation*}

We define two metrics, \textit{Maximum Local Mutual Information ($MLMI$)} and \textit{Maximum Mutual Information ($maxMI$)}, as follows.
Let $IV_C$ denote the set of interface variables in a clique $C$. The $MLMI$ of a variable $v$ in clique $C$ is defined as
\begin{equation}\label{eq:mlmi}
    MLMI_{v,C} = \max\limits_{\forall x\in {IV_C\setminus v}} MI(v;x)
  \end{equation}

The $maxMI$ for a variable $v$ is defined as the maximum $MLMI$ over all cliques.
    \begin{align}\label{eq:maxmi}
	maxMI_v=\max_{\forall C\in CTF ~s.t. ~v\in C} MLMI_{v,C}
	\end{align}
As seen in Equation~\ref{eq:mlmi}, if $v$ is an interface variable, $MLMI$ is the maximum MI between $v$ and the other interface variables in the clique. If $v$ is a non-interface variable, it is the maximum MI between $v$ and all the interface variables in the clique. Since $maxMI$ of $v$ is the maximum $MLMI$ over all cliques, it is a measure of the maximum influence that a variable $v$ has on interface variables that are present in cliques that contain $v$. A low $maxMI$ means that $v$ has a low $MI$ with interface variables in all the cliques in which it is present and is therefore assumed to have a lower impact on the joint distribution of the interface variables. 

Note that we do not compute the MI between $v$ and all interface variables in the CTF. This is because if $v$ and an interface variable $i_k$ are present in different cliques, computation of $MI$ becomes expensive. However, our heuristic gives good results for the following reason. If $MI(v;i_k)$ is large and the two are present in different cliques, it can only happen via a sepset variable $w$ that has a large $MI$ with both $v$ and $i_k$. Even if $v$ happens to have a low $maxMI$ and is locally marginalized, $w$ will have a large $MLMI$ and is likely to be retained.

We prioritize non-interface variables with the least $maxMI$ for local marginalization.
If it is not possible to reduce clique sizes by removing non-interface variables, we locally marginalize over interface variables with least $maxMI$. However, we make sure that the interface variable is retained in atleast one clique. 

If evidence variables are present, we require a connected CT to remain connected. 
If the CTs get disconnected, the product of the normalization constants of the disjoint CTs will not be preserved leading to erroneous inference of PR and posterior beliefs.
Therefore, if a sepset contains a single variable, this variable is not chosen for marginalization even if it has low MI. This constraint is not needed if only prior beliefs are required.

\subsubsection{Example}
\begin{figure}
    \centering

\begin{tikzpicture}[thick,scale=0.8, every node/.style={transform shape}]


    \foreach \pos/\name/\label in {(-1.5,-1)/$C_8:ab\textcolor{red}{f}$/c1, (1.25,-1)/$C_9:abe$/c2, (-1.5,-2.25)/$C_{10}:\textcolor{red}{f}gdh$/c6, (-1.5,-3.75)/$C_{11}:fmn$/c9}
    \node[vertexr] (\label) at \pos {\name};
    
    \foreach \pos/\name/\label in {(4.25,-1)/$C_6:chj$/c3, (7,-1)/$C_4:hi$/c4, (7,-2.25)/$C_5:i\textcolor{red}{l}$/c5, (1.25,-2.25)/$C_2:\textcolor{red}{f}dhm$/c7, (4.25,-2.25)/$C_7:hmj$/c8, (4.25,-3.75)/$C_3:dh\textcolor{red}{k}$/c10, (1.25, -3.75)/$C_1:dm\textcolor{red}{o}$/c11}
    \node[selected vertexr] (\label) at \pos {\name};

    \foreach \source/\dest/\weight  in {c1/c2/$ab$, c6/c7/$fdh$, c3/c4/$h$, c7/c8/$hm$}
    \path[edge2] (\source) -- node [above,midway ] {\color{gray} \weight} (\dest);

    \foreach \source/\dest/\weight  in {c1/c6/$f$, c7/c10/$dh$, c7/c11/$dm$, c8/c3/$jh$, c4/c5/$i$}
    \path[edge2] (\source) -- node [right,midway] {\color{gray} \weight} (\dest);

    \foreach \source/\dest/\weight  in {c7/c9/$fm$}
    \path[edge2] (\source) -- node [left,midway] {\color{gray} \weight} (\dest);
    
    \draw [dashed] ([xshift=-0.25cm,yshift=0.25cm]c3.north west) rectangle ([xshift=0.25cm,yshift=-0.25cm]c8.south east);
    \draw [dashed] ([xshift=-0.25cm,yshift=0.25cm]c4.north west) rectangle ([xshift=0.25cm,yshift=-0.25cm]c5.south east);

    \foreach \pos/\name/\label in {(12,-2.25)/${C_2}:fdhm$/fdhm, (14.5,-2.25)/${C_4}':hl$/hl, (12,-3.75)/${C_1}:dmo$/dmo, (14.5,-3.75)/${C_3}:dhk$/dhk}
    \node [selected vertexr] (\label) at \pos {\name};
    \node (ctfp) [above=0.2 of fdhm] {\color{blue}$\mathbf{{CTF_1}'}$};
    
    \foreach \source/\dest/\weight  in {fdhm/hl/h}
    \path[edge2] (\source) -- node [above,midway ] {\color{gray} $\weight$} (\dest);

    \foreach \source/\dest/\weight  in {fdhm/dhk/dh, fdhm/dmo/dm}
    \path[edge2] (\source) -- node [right,midway ] {\color{gray} $\weight$} (\dest);
    \path[draw,edge5] ([xshift=0.25cm]c5.east)--node[above,midway]{\bf Exact Marg.} node[below, midway]{\bf $c,j,i$} ([xshift=-0.2cm]fdhm.west);

    \foreach \pos/\label/\name in {(14.5,-8.5)/dhk2/${C_3}':dhk$, (12,-7)/fhm/${C_2}':fhm$, (14.5,-7)/hl2/${C_4}':hl$, (12,-8.5)/mo/${C_1}':mo$}
    \node[selected vertexr] (\label) at \pos {\name};
    \node (ctfp) [above=0.2 of fhm] {\color{blue}$\mathbf{{CTF_{1,a}}}$};
    
    \foreach \source\dest\weight in {fhm/hl2/$h$}
    \path[edge2] (\source) -- node[above, midway] {\color{gray}\weight} (\dest);
    \foreach \source\dest\weight in {dhk2/fhm/$h$,fhm/mo/$m$}
    \path[edge2] (\source) -- node[right, midway] {\color{gray}\weight} (\dest);
    
    \path[draw,edge5] ([yshift=-0.25cm, xshift=1cm]dmo.south)--node[right,midway]{\bf Local Marg.} node[right, midway, xshift=1cm, yshift=-0.5cm] {$d$}  ([yshift=0.2cm, xshift=1cm]fhm.north);
    
    \foreach \pos/\name/\label in {(-3,-5.5)/{\color{blue}$\mathbf{IM_1}$}/im,(-2,-6)/{${C'\in CTF_{1,a}}$}/ca, (0.5,-6)/{${C\in CTF_1}$}/c, (-2,-7)/$~~~{C_2}':fhm$/fhm, (0.5,-7)/$~~C_2:fdhm$/fdhm, (-2,-6.5)/$~{C_1}':mo$/mo, (0.5,-6.5)/$C_1:dmo$/dmo, (-2,-7.5)/$~~{C_3}':dhk$/dhk1, (0.5,-7.5)/${C_3}:dhk$/dhk2, (-2,-8)/${C_4}':hl$/hl, (0.5,-8)/${C_4:hi}$/hil, (0.5,-8.5)/${C_5:il}$/il, (-3,-1)/{\color{blue}$\mathbf{CTF_{1}}$}/ctfin}
    \node (\label) at \pos {\name};
    \path [edge2] (-3,-6.25)--(2,-6.25);
    \path [edge2] (-0.75,-5.75)--(-0.75,-8.75);             
    \path [edge2] (2,-5.75)--(2,-8.75);             
    \end{tikzpicture}
    \caption{Approximation of $CTF_1$ constructed for the running example. The blue cliques in $CTF_1$ form the minimal subgraph containing interface variables $f,~k,~o$ and $l$ (marked in red). $CTF_{1,a}$ is obtained after exact marginalization of non-interface variables $c,j,i$ and  local marginalization of variable $d$. The interface map $IM_1$ that contains links between cliques in $CTF_{1}$ and $CTF_{1,a}$ is also shown.}
    \label{fig:approxCTF1}
\end{figure}
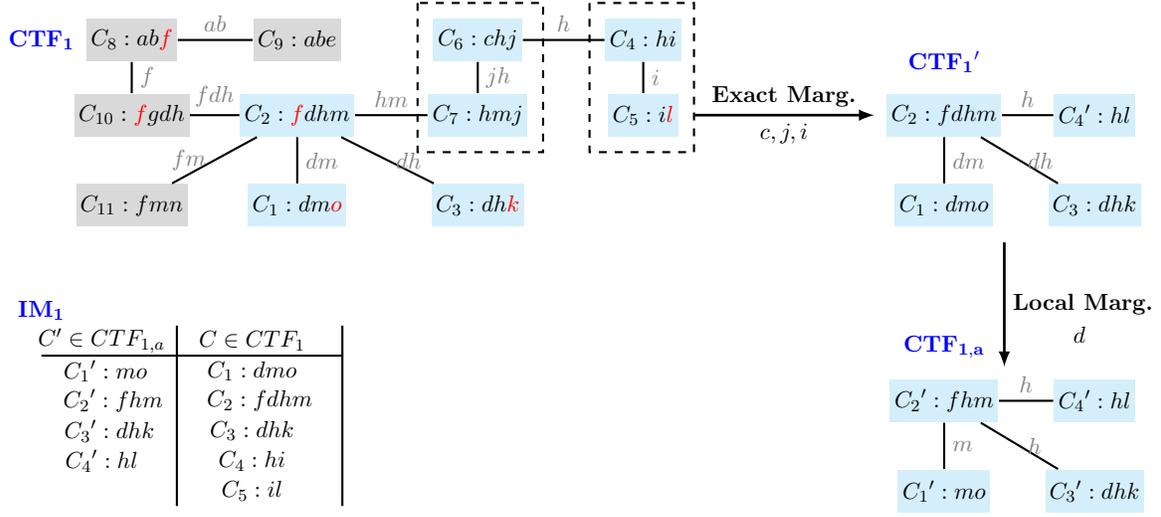
Figure~\ref{fig:approxCTF1} shows approximation of first CTF of the SLCTF ($CTF_1$) for the running  example. In the example, $mcs_{im}$ is set to 3 and $IV=\{f,l,k,o\}$ (shown in Figure~\ref{fig:build1}).
As explained, for the next CTF in the sequence, we only need the joint beliefs of the interface variables. 
Therefore, $CTF_{1,a}$ is initialized to $MSG[IV]$.
In Figure~\ref{fig:approxCTF1}, the interface variables are marked in red and $CTF_{1,a}$ is initialized to the part of $CTF_1$ highlighted in blue.
From the definition of $MSG[IV]$, it is clear that we do not need to include $C_8$ and $C_9$ which also contain the interface variable $f$, since $f$ is contained in $C_2$.
Our aim is to make sure that all cliques in $CTF_{1,a}$ have a maximum size of 3 ($mcs_{im}=3$) and reduce the number of non-interface variables and cliques in $CTF_{1,a}$ as much as possible. As discussed, for this we use a combination of exact and local marginalization.

As seen in the figure, the variable $c$ is present in a single clique $C_6$ and is a non-interface variable. Since $C_6$ has size equal to $mcs_{im}$, $c$ is removed from $C_6$ and the belief is marginalized. Once this is done, $C_6$ will contain only $h$ and $j$, both of which are also present in $C_7$. Since $C_6$ is a non-maximal clique, it is removed and its neighbour $C_4$ is connected to $C_7$. In $C_7$, $j$ is a non-interface variable, present in a single clique. We can follow a similar process of marginalization and removal of a non-maximal clique, leaving only $C_1,C_2,C_3, C_4$ and $C_5$ in $CTF_{1,a}$.
We can reduce the number of non-interface variables further.
Collapsing $C_4$ and $C_5$, gives a new clique containing the variables $h,i$ and $l$. Once again, since $i$ is a non-interface variable present in only one clique, it is removed and the beliefs marginalized to give a new clique $C_4'$. The CTF obtained after the exact marginalization steps is denoted ${CTF_1}'$.

After exact marginalization, there is one clique $C_2$, with size greater than $mcs_{im}$. To reduce its size, we do a local marginalization. In ${CTF_1}'$, the choice of variables for local marginalization are $f, d, h$ and $m$. Since $f$ is an interface variable that is present in a single clique with size greater than $mcs_{im}$, it is not considered for marginalization. Variable $h$ is also not considered, because removal of $h$ from cliques $C_2$ and ${C_4}'$ will disconnect the clique tree, since the sepset between them contains only $h$. 
We then sort $d$ and $m$ in ascending order of $maxMI$. Assuming $d$ has least $maxMI$, it is removed from $C_2$ and it must be removed from either $C_1$ or $C_3$ to satisfy RIP. Assuming that the MLMI of $d$ in $C_3$ is greater than in $C_1$, it is retained in $C_3$ and removed from $C_1$. The beliefs of $C_1$ and $C_2$ are marginalized. The resulting approximated CTF, $CTF_{1,a}$, is used as a starting point for the next CTF in the sequence.

Figure~\ref{fig:approxCTF1} has the interface map $IM_1$ which links cliques in $CTF_1$ and $CTF_{1,a}$. In our example, ${C_1}'$ and ${C_2}'$ are obtained after local marginalization of $C_1$ and $C_2$ respectively.
It is seen from Figure~\ref{fig:approxCTF1}, that the corresponding cliques are linked in $IM_1$.
${C_4}'$ is obtained after collapsing $C_4$ and $C_5$ and is therefore linked to both in $IM_1$.

\subsubsection{Algorithm}
\begin{algorithm}[!h]
	\scriptsize
	\caption{ApproximateCTF~($CTF_{in}, ~IV, ~mcs_{im},~E$)}
	\label{alg:ApproximateCTF}	
	\begin{algorithmic}[1]
		\Require~$CTF_{in}$: Input clique tree forest \newline
		\indent$IV$: Interface variables\newline
		\indent$mcs_{im}$: Maximum clique size limit for interface $CTF$
		\newline\indent$E$: Boolean variable indicating presence of evidence
		\Ensure$CTF_a$: Approximate $CTF$\newline
		\indent$IM$: Dictionary linking cliques $CTF_{a}$ to $CTF_{in}$ 
		\LineComment{{\color{teal!70}Initialize $CTF_a$ as the minimal subgraph that connects the $IV$}}
        \State $CTF_a \gets MSG[IV]$ 
                \LineComment{{\color{teal!70} Step1: Exact marginalization}}
		    \State $NIV\gets $ Variables $\in CTF_a\setminus IV$ \Comment{{\color{teal!70}\scriptsize Identify non-interface variables in $CTF_a$}}
                \State Sum out NIVs present in a single clique in $CTF_a$
		\State Remove resultant non-maximal cliques, reconnect neighbors
        \State Collapse cliques and marginalize NIVs as long as size of the collapsed clique is less than $mcs_{im}$
		\LineComment{{\color{teal!70}Step 2: Local marginalization}} 
        \color{black}
		\State $L_c\gets$ List of cliques with size $> mcs_{im}$
		\State Compute metrics $maxMI$, $MLMI$ for variables in all cliques in $L_c$  
		\State $NIV\gets$ Variables $\in L_c\setminus IV$ \Comment{{\color{teal!70}\scriptsize Find non-interface variables in $L_c$}}
        \State $N\gets [Sort(NIV),Sort(IV),metric=maxMI]$
        \Comment{{\color{teal!70}\scriptsize Sort NIV and IV in ascending order of $maxMI$}}
        \While{$L_c\neq \varnothing$ \&\& $N\neq \varnothing$} 
        \State $v=N.$pop() \Comment{{\color{teal!70} \scriptsize Get the variable with the least $maxMI$}}
		\State $ST_v\gets$ Subgraph of $CTF_a$ over cliques containing $v$
        \LineComment{{\color{teal!70}Identify a subtree $ST_r$ in which $v$ can be retained}}
        \State $C_{rv}\gets$ Clique $\in ST_v$ with the largest $MLMI_v$, clique-size $\leq mcs_{im}$
        \State $ST_r\gets $ Connected subtree $\in ST_v$ containing $C_{rv}$ s.t. max-clique-size $\leq mcs_{im}$
		\State $L_m\gets$ Cliques $\in ST_v \setminus$Cliques $\in ST_r$
        \Comment{{\color{teal!70} \scriptsize Use these cliques for local marginalization.}}
        \If{$E$}
		\State  $ms\gets$ Minimum sep-set size for cliques in $L_m$
        \LineComment{{\color{teal!70}Ignore $v$ if the minimum sepset size is 1 or if it is an interface variable present only in large cliques.}}
        \IIf{$ms==1~||~(ST_r.isEmpty() ~\&\&~ v\in IV)$} continue 
        \Comment{{\color{teal!70} \scriptsize Ignore $v$ to keep CT connected}}
        \Else
		    \If{$ST_r.isEmpty() ~\&\&~ v\in IV$}  \Comment{{\color{teal!70} \scriptsize All cliques containing $v$ have size $> mcs_{im}$}}
                \State Add an independent clique $C$ with $C=\{v\}$  \Comment{{\color{teal!70} \scriptsize Do this if the CT can be disconnected}}
            \EndIf            
        \EndIf
		\State Remove $v$ and locally marginalize beliefs of all cliques in $L_m$. 
		\State Remove resultant non-maximal cliques, reconnect neighbors
		\State $L_c$.remove($C$)  if $C$.size $\leq mcs_{im}$, $\forall ~C\in L_m$  
		\EndWhile 
        \LineComment{{\color{teal!70} \scriptsize Step 3: Create Interface Map: Map between cliques in $CTF_{a}$ and $CTF_{in}$ }}
		\State $IM(C')= $ List of constituent cliques in $CTF_{in}$,  $ ~\forall ~C' \in CTF_a~ $  \Comment{{\color{teal!70} \scriptsize Link cliques in $CTF_a$ and $CTF_{in}$ }}
		\LineComment{{\color{teal!70}Step 4: Re-assign clique factors}}
		\For{$CT \in CTF_a$}
		\State root $\gets$ Pick a node at random from CT
        \State $\psi_{root} = \beta_{root}$; $L_v = [root]$ 
        \Comment{{\color{teal!70}Belief of root node is the clique belief.}}
        \LineComment{{\color{teal!70}Assign conditional beliefs to other cliques.}}
		\For {$C_i\in L_v$}
		\FFor {$C_j \in \{CT$.neighbors($C_i$) $\not\in L_v\}$~} $\psi_{C_j} = \beta_{j}/\mu_{ij}$; $L_v$.add($C_j$)			
		\EndFor
		\EndFor
	\State \Return $CTF_a, ~IM$
	\end{algorithmic}
\end{algorithm}

$ApproximateCTF$ (Algorithm~\ref{alg:ApproximateCTF}) is the function used to approximate the CTF.  The inputs are $CTF_{in}$ and the set of interface variables $IV$. 
$CTF_a$ is initialized to $MSG[IV]$.
The algorithm performs four main steps namely, exact marginalization, local marginalization, finding links between cliques in $CTF_{in}$ and $CTF_a$ and finally re-assigning clique factors to re-parameterize the joint distribution. 
The steps involved are as follows.

\step{\it Exact marginalization:}
We first marginalize out non-interface variables present in a single clique and update beliefs. 
We then perform exact marginalization by collapsing cliques and marginalizing out non-interface variables as long as the clique sizes are less than $mcs_{im}$.
If any clique obtained after marginalization is a non-maximal clique, then it is removed from $CTF_a$, and its neighbors are connected to the containing clique.

\step{\it Local marginalization:}
If $CTF_a$ obtained after the first step contains cliques with size greater than $mcs_{im}$, we perform local marginalization. We first get a list $L_c$ that contains cliques that have size greater than $mcs_{im}$ and compute metrics $maxMI$ and $MLMI$ for all variables in these cliques. The set of interface and non-interface variables is sorted in the ascending order of $maxMI$. We prioritize non-interface variables for marginalization. 
Let $v$ be the variable with the least $maxMI$ and $C_{rv}$ be a clique of size atmost $mcs_{im}$, in which $v$ has the largest MLMI. We retain $v$ in a subtree containing $C_{rv}$ such that the maximum clique size in the subtree is atmost $mcs_{im}$ (denoted $ST_r$). Variable $v$ is locally marginalized from all other cliques (contained in $L_m$).
If any clique obtained after local marginalization is a non-maximal clique, then it is removed from $CTF_a$, and its neighbors are connected to the containing clique. 

If $E$ is true (that is, BN has evidence and query is $MAR_e/PR$), we need to ensure that a connected CT remains connected, as discussed in section~\ref{sec:choice}.
Therefore, we ignore variables for which the minimum sepset size ($ms$) is equal to one.
In this case, we also ignore interface variables that are only present in cliques with size greater than $mcs_{im}$ ($ST_r$ is empty). If $E$ is False, we do not need to keep the CT connected. Therefore, if $ST_r$ is empty, an interface variable can be removed from all cliques and added as an independent clique. 



\step{\it Create Interface Map (IM):} In this step, we link cliques in $CTF_{in}$ and $CTF_a$.
For each clique $C'$ in $CTF_a$, we add a link to a clique or a set of cliques in $CTF_{in}$ in $IM$ as follows
\begin{itemize}
  \item If $C'$ is obtained after collapsing a set of cliques $\{C_1, \cdots C_m\}$ in $CTF_{in}$,  links are added from $C'$ to each of $\{C_1, \cdots C_m\}$.
    \item If $C'$ is obtained from $C$ in $CTF_{in}$ after a local marginalization, a link is added from $C'$ to $C$.
  \item If $C'$ is same as clique $C$ in $CTF_{in}$, a link is added from $C'$ to $C$.
\end{itemize}
$IM$ is implemented as a Python dictionary, with cliques $C'$ in $CTF_a$ used as keys and the corresponding cliques in $CTF_{in}$ as values i.e. $IM[C']=\{C\}$.

\step{\it Re-assignment of clique factors:} ~\label{sec:reassignFactor}
In the following section (section 4.3.5), we prove that all CTs in $CTF_a$ are calibrated (see Proposition~\ref{pr:approx1}). Therefore, the joint distribution can be obtained from the clique and sepset beliefs (see Definition~\ref{def:calCT}). 
$CTF_a$ forms the initial CTF for the construction of the next CTF in the sequence. In order to calibrate the next CTF, the product of the factors in $CTF_a$ must give a valid joint distribution. Therefore, we re-parameterize the joint beliefs of $CTF_a$ to satisfy this constraint. This is done as follows.
For each CT in the $CTF_a$, a root node is chosen at random. 
The factor for the root node is the same as the clique belief.
All other nodes are assigned factors by iterating through them in pre-order, i.e., from the root node to the leaf nodes. 
An un-visited neighbor ${C_{j}}'$ of a node ${C_{i}}'$ in $CTF_a$ is assigned the conditional belief $\beta({C_j}'|{C_i}')=\dfrac{\beta({C'_j})}{\mu({S'_{i,j}})}$ as a factor.
This ensures that the product of the factors is the joint distribution of variables in $CTF_a$.

\subsubsection{Properties of the approximated CTF}
The resulting approximate CTF, denoted $CTF_a$ satisfies the following properties.
The proofs for these properties are included in Appendix \ref{app:proofs}.


\begin{proposition} \label{pr:approx1}
  All CTs in $CTF_a$ are valid CTs that are calibrated.
\end{proposition}

\begin{proposition} \label{pr:approx2}
  Algorithm \ref{alg:ApproximateCTF} preserves the  normalization constant and the within-clique beliefs of all cliques in $CTF_a$.
    \end{proposition}
Based on this proposition, the joint belief of variables present within any clique in $CTF_a$ is the same in both $CTF_a$ and $CTF_{in}$. In other words, the within-clique beliefs are consistent in $CTF_{in}$ and $CTF_{a}$.
However, the joint beliefs of variables present in different cliques of a CT in $CTF_a$  may not be preserved.
This is because variables that are locally marginalized from a clique are also locally marginalized from the corresponding sepsets.

\begin{proposition} \label{pr:approx6}
  If the clique beliefs are uniform, then the beliefs obtained after local marginalization is exact
\end{proposition}

\subsection{Updating links between adjacent CTFs}\label{sec:links}
The build, infer and approximate steps are used repeatedly until all variables are added to some CTF in the SLCTF. At this point, we have a sequence of CTFs and interface maps between each CTF and its approximation.
In this section, we discuss the method to update the interface map $IM$ so that it contains links between adjacent CTFs. As will be seen in Section~\ref{sec:pm}, these links are needed for computation of the posterior marginals, $MAR_e$.

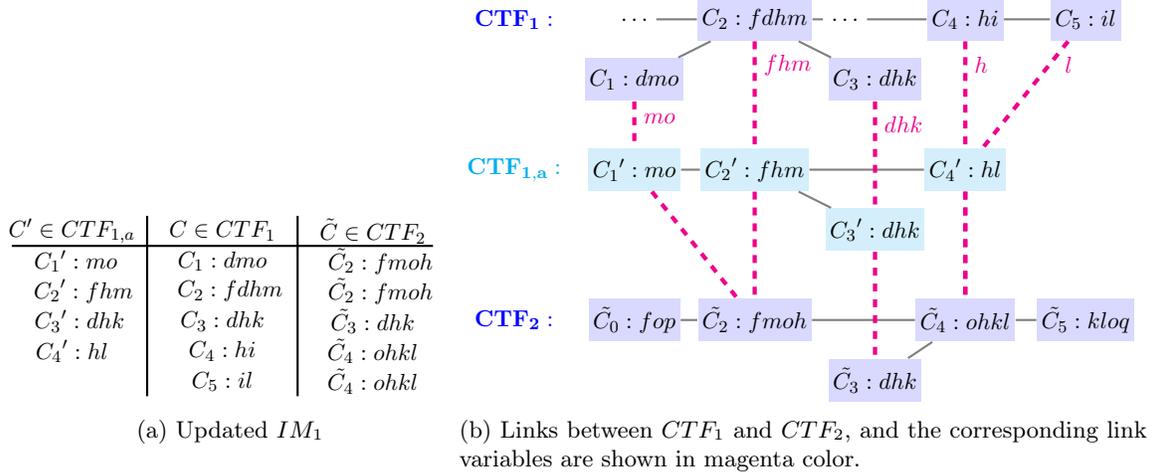
\begin{figure}[htb]
\centering
    \begin{subfigure}[t]{0.4\textwidth}
\begin{tikzpicture}[thick,scale=0.8, every node/.style={transform shape}]
    \foreach \pos/\name/\label in {(0,0)/{${C'\in CTF_{1,a}}$}/ca, (2.5,0)/{${C\in CTF_1}$}/c, (0,-1)/$~~~{C_2}':fhm$/fhm, (2.5,-1)/$~~C_2:fdhm$/fdhm, (0,-0.5)/$~{C_1}':mo$/mo, (2.5,-0.5)/$C_1:dmo$/dmo, (0,-1.5)/$~~{C_3}':dhk$/dhk1, (2.5,-1.5)/$C_3:dhk$/dhk2, (0,-2)/${C_4}':hl$/hl, (2.5,-2)/${C_4:hi}$/hil,(5,0)/{${\tilde{C}\in CTF_2}$}/ct,(5,-0.5)/$~~\tilde{C_2}:fmoh$/fmoh1,(5,-1)/$~~\tilde{C_2}:fmoh$/fmoh2, (5,-1.5)/$\tilde{C_3}:dhk$/dhk3, (5,-2)/$\tilde{C_4}:ohkl$/ohkl, (2.5,-2.5)/${C_5:il}$/il, (5,-2.5)/${\tilde{C_4}:ohkl}$/ohkl1}
    \node (\label) at \pos {\name};
    \path [edge2] (-1,-0.25)--(6,-0.25);
    \path [edge2] (1.25,0.25)--(1.25,-2.75);             
    \path [edge2] (3.75,0.25)--(3.75,-2.75);             
\end{tikzpicture}
        \caption{Updated $IM_1$}
        \label{fig:updatedIM}
    \end{subfigure}%
    \begin{subfigure}[t]{0.6\textwidth}
    \begin{tikzpicture}[thick,scale=0.8, every node/.style={transform shape}]
    \node (ctf1) at (-3,0) {\color{blue}$\mathbf{CTF_1:}$};
        \node [selected vertexb](dhk3) at (3,-6) {$\tilde{C_3}:dhk$}; 
        \node [selected vertexr](dhk2) at (3,-3.5) {${C_3}':dhk$}; 
        \node [selected vertexb](dhk1) at (3,-1) {$C_3:dhk$}; 

        \foreach \pos/\name/\label in {(1,0)/{$C_2:fdhm$}/fdhm, (4.5,0)/{$C_4:hi$}/hi, (6.5,0)/{$C_5:il$}/il, (-1,-1)/{$C_1:dmo$}/dmo}
    \node [selected vertexb] (\label) at \pos {\name};
    \node (hd) at (2.5,0) {$\hdots$};
    \node (hd1) at (-1,0) {$\hdots$};
    \path[edge2, gray] (hd1)--(fdhm)--(hd)--(hi)--(il);
    
    \foreach \source/\dest in {fdhm/dmo, fdhm/dhk1}
    \path[edge2, gray] (\source)-- (\dest);

        \node (ctfa) at (-3,-2.5) {\color{cyan}$\mathbf{CTF_{1,a}:}$};
        \foreach \pos/\name/\label in {(-1,-2.5)/{${C_1}':mo$}/mo, (1,-2.5)/{${C_2}':fhm$}/fhm, (4.5,-2.5)/{${C_4}':hl$}/hl}
    \node [selected vertexr] (\label) at \pos {\name};
    \foreach \source/\dest in {fhm/mo, fhm/dhk2, fhm/hl}
    \path[edge2,gray] (\source)-- (\dest);

    \node (ctfa) at (-3,-5) {\color{blue}$\mathbf{CTF_2:}$};
        \foreach \pos/\name/\label in {(-1,-5)/{$\tilde{C_0}:fop$}/fop, (1,-5)/{$\tilde{C_2}:fmoh$}/fmoh, (4.5,-5)/{$\tilde{C_4}:ohkl$}/ohkl, (6.5,-5)/{$\tilde{C_5}:kloq$}/kloq}
    \node [selected vertexb] (\label) at \pos {\name};
    \foreach \source/\dest in {fmoh/ohkl, ohkl/dhk3}
    \path[edge2,gray] (\source)-- (\dest);
    \path [edge2, gray] (ohkl) --(kloq);
    \path [edge2, gray] (fmoh) --(fop);

    \foreach \source/\destx/\desty/\weight in { fhm/fdhm/fmoh/$fhm$, hl/hi/ohkl/$h$, dhk2/dhk1/dhk3/$dhk$}
    \path[udashededge] (\destx) -- node[right,midway, yshift=0.5cm]{\weight} (\source)-- (\desty);
    \path[udashededge] (dmo) -- node[right,midway, yshift=0.1cm]{$mo$} (mo)-- (fmoh);
    \path[udashededge] (il) -- node[right,midway, yshift=0.5cm, xshift=0.5cm]{$l$} (hl)-- (ohkl);
    \end{tikzpicture}
    \caption{Links between $CTF_1$ and $CTF_2$, and the corresponding link variables are shown in magenta color.}
    \label{fig:ex-links}
    \end{subfigure}
    \caption{The SLCTF for the running example, which has two CTFs, $CTF_1$ and $CTF_2$, and an interface map $IM_1$. The updated $IM_1$ is obtained using Algorithm~\ref{alg:updateInterfaceMap}. 
    }
    \label{fig:links}
\end{figure}
We first show the process using the running example. As shown in Figure~\ref{fig:approxCTF1}, we have added links between cliques in $CTF_1$ and $CTF_{1,a}$ to $IM_1$. 
Figure~\ref{fig:links} shows the two CTFs, $CTF_1$ and $CTF_2$ in the SLCTF obtained for the running example as well as the updated $IM_1$. 
$CTF_2$ is incrementally built by adding the variables $p$ and $q$ to $CTF_{1,a}$. Therefore, each clique $C'$ in $CTF_{1,a}$ will be contained in some clique $\tilde{C}$ in $CTF_2$. As shown in Figure~\ref{fig:links}, we add a link between $C'$ and $\tilde{C}$ in $IM_1$. For example, the variables in both ${C_1}'$ and ${C_2}'$ are contained in $\tilde{C_2}$. Hence, both cliques are linked to $\tilde{C_2}$. 


Algorithm \ref{alg:updateInterfaceMap} describes the procedure used to update the links. 
$CTF_{k}$ and $CTF_{k,a}$ denote the $k^{th}$ CTF in the sequence and its approximation respectively. $CTF_{k+1}$ is the next CTF in the sequence. 
$IM_k$ is the interface map that contains links between cliques in $CTF_k$ and $CTF_{k,a}$. 
Since $CTF_{k+1}$ is constructed by incrementally adding variables to $CTF_{k,a}$,  each clique ${C_j}'$ in $CTF_{k,a}$ is contained in some clique $\tilde{C_j}$ in $CTF_{k+1}$. 
We update the interface map by adding a clique $\tilde{C_j}$ for each  ${C_j}'$ in the dictionary $IM_{k}$ such that ${C_j}'$ is a subset of $\tilde{C_j}$. 
At the end of this step, each  ${C_j}'$ in $IM_{k}$ has links to one or more cliques in the $CTF_{k}$ and a link to a clique in $CTF_{k+1}$.
The variables in $CTF_{k,a}$ are called the link variables (Definition~\ref{def:lv}, section~\ref{sec:definitions}). The link variables associated with the link between between cliques $C_j\in CTF_k$ and $\tilde{C_j}\in CTF_{k+1}$ is the set $L=C_j\cap {C_j}'$.
\begin{algorithm}[h]
	\scriptsize
	\caption{UpdateInterfaceMap($L_{CTF}, L_{IM}$)}
	\label{alg:updateInterfaceMap}
	\begin{algorithmic}[1]
		\Require~$L_{CTF}$: List containing sequence of CTFs corresponding to G \newline
		\indent$L_{IM}$: List of interface maps for adjacent CTFs 
        \Ensure~$L_{IM}$: Updated list of interface maps for adjacent CTFs
        \For{$k = 1$ to len($L_{CTF}$)}
		        \State $CTF_{k+1} = L_{CTF}[k+1]$; ~$IM_{k} = L_{IM}[k]$;
                \LineComment{{\color{teal!70} $IM_k$ is a dictionary with approximate cliques as keys}}
                \State $C_a = IM_{k}.keys()$
                \LineComment{{\color{teal!70} Add links between approximate cliques in $CTF_{k,a}$ and cliques in $CTF_{k+1}$ }}
			    \State $IM_{k}({C_j}')$.add($\tilde{C_j}$),  $ ~\forall ~{C_j}' \in C_a$~ such that ${C_j}' \subseteq \tilde{C_j} \in CTF_{k+1}$  
	    \EndFor
        \State \Return $L_{IM}$
	\end{algorithmic}
\end{algorithm}


\section{Approximate Inference of the partition function and marginals}\label{sec:answerQueries}

In this section, we discuss how the constructed SLCTF can be used for approximate inference of three probability queries namely, Prior marginals ($MAR_p$), Partition function ($PR$), Posterior marginals ($MAR_e$). We first discuss propositions that are required to answer each of these queries and the implications of the proposition on the running example. Following this, we describe the algorithm $InferPRandMAR$.
The proofs of all propositions and theorems are included in the Appendix~\ref{app:proofs}.

\subsection{Prior marginals ($MAR_p$)}
All CTFs in the sequence $L_{CTF}$ are calibrated in the corresponding infer phase. 
A consequence of Proposition~\ref{pr:approx2} is the following proposition which shows how the prior marginals can be obtained using the calibrated clique beliefs.
\begin{proposition} \label{pr:prior}
    In the absence of evidence, the estimate of prior singleton marginal of a variable can be obtained from any of the CTFs in which it is present.
\end{proposition}

As seen from the proof, this is because in the absence of evidence the within-clique beliefs are preserved across CTFs. For our running example, as seen in Figure~\ref{fig:links}, variables $f,h,m,o,d,k$ and $l$ are present in both $CTF_1$ and $CTF_2$. This proposition guarantees that the prior marginals are the same in both CTFs.

\textbf{Note:} The prior marginals estimated using IBIA are exact for variables that belong to the first CTF in the sequence.

\subsection{Partition function ($PR$)}\label{sec:pr}
Simplification of the BN could give a set of disjoint DAGs. If exact join-tree based inference is used, we will get a single CT corresponding to each DAG. After calibration, the normalization constant of all the clique and sepset beliefs in the CT is the same and is the probability of the evidence variables present in the DAG. The partition function is the product of the normalization constants of CTs corresponding to each DAG. 

In our method, we have a sequence of calibrated CTFs corresponding to each DAG. 
Evidence variables can be added to any of the CTFs in the sequence. Therefore, the normalization constant of each CTF (obtained using a product of the normalization constants of the CTs in the CTF) could be different. 
Since Algorithm~\ref{alg:ApproximateCTF} keeps a connected CT connected, the last CTF in the sequence will contain a single CT. 
Proposition~\ref{pr:pf1} and Theorem~\ref{pr:pf2} show how PR can be obtained from the SLCTF.

\begin{proposition}\label{pr:pf1}
    The normalization constant of a CT in $CTF_{k}$ is the estimate of probability of all evidence states added to it in the current and all preceding CTFs $\{CTF_1,\hdots, CTF_k\}$.
\end{proposition}
\begin{theorem} \label{pr:pf2}
The product of the normalization constants of the CTs corresponding to the last CTF in the sequence for all DAGs in the BN is the estimate of Partition Function (PR).
\end{theorem}

In our running example, there are two evidence variables $e$ and $p$ with corresponding states $e_s$ and $p_s$. As seen in Figure~\ref{fig:approxCTF1}, variable $e$ is present in $CTF_1$.   Therefore, the normalization constant obtained after summing any clique belief in $CTF_1$ corresponds to $P(e=e_s)$. Based on Proposition~\ref{pr:approx2}, we know that the approximation algorithm preserves the normalization constant. Therefore, if no new evidence variables are added in $CTF_2$, the normalization constants for $CTF_1$ and $CTF_2$ are equal. However, as seen in Figure~\ref{fig:links}, the second evidence variable $p$ is added to $CTF_2$. Using Proposition~\ref{pr:pf1}, when the new evidence state $p=p_s$ is added, the normalization constant obtained after summing any clique belief in $CTF_2$ is an estimate of the partition function $P(e=e_s,p=p_s)$. 
      
\textbf{Note:} PR obtained with IBIA is exact if all evidence variables are added to the first CTF in the sequence.

\subsection{Posterior Marginals ($MAR_e$)} \label{sec:pm}
Besides the sequence of CTFs, the SLCTF contains links between cliques in adjacent CTFs in $L_{IM}$ and index $I_E$ of the last CTF in the sequence in which new evidence variables have been added. 
Based on Proposition~\ref{pr:approx2}, we know that if no new evidence is added, within-clique beliefs of the link variables are preserved in adjacent CTFs.
Therefore, similar to prior beliefs, once all evidence variables have been added, the beliefs of variables do not change in subsequent CTFs.
Thus, as proved in Theorem \ref{th:post1}, we can estimate of posterior marginals of all variables in CTFs $\{CTF_{k}, k \geq I_E\}$ from the calibrated clique beliefs. 
\begin{theorem} \label{th:post1}
  The singleton posterior marginals of variables in CTFs $\{CTF_{k}, k \geq I_E\}$ are preserved and can be computed from any of these CTFs.
\end{theorem}

However, when new evidence variables are added to a CTF, the posterior belief of a variable changes and is not the same as its belief in the earlier CTFs in the sequence. 
Hence, we need to do a belief update in CTFs  $\{CTF_k, ~k< I_E\}$ to make it consistent with all evidence states. 
This is done using links between cliques in adjacent CTFs stored in the interface maps. 

We first explain how the belief update is done using the running example. As mentioned in Section~\ref{sec:pr}, $CTF_2$ (shown in Figure~\ref{fig:links}) is consistent with both evidence variables $e$ and $p$, but $CTF_1$ (shown in Figure~\ref{fig:approxCTF1}) needs a belief update to account for $p$. 
As seen in Figure~\ref{fig:links}, $IM_1$ links clique $C$ in $CTF_1$ and $\tilde{C}$ in $CTF_2$ via link variables in $C'$ in $CTF_{1,a}$. Consider the cliques $C_2$ and $\tilde{C_2}$ which are linked via the link variables $f,h$ and $m$. After adding the evidence variable $p$, the beliefs of the link variables are not the same in $C_2$ and $\tilde{C_2}$.
We first update $\beta(C_2)$ to make sure they are consistent. This is done as shown in equation (\ref{eq:bup}).
\begin{equation}
  \beta(C_2) = P(f,h,d,m) = \left (\frac{\beta(C_2)}{\sum\limits_{d.states}\beta({C_2})}\right ) \sum\limits_{o.states}\beta(\tilde{C_2}) = P(d|f,h,m)\tilde{P}(f,h,m)
  \label{eq:bup}
\end{equation}
Here, $\tilde{P}$ denotes the beliefs of the link variables in $\tilde{C}_2$ and $d$ and $o$ are variables other than the link variables in $C_2$ and $\tilde{C_2}$ respectively.
Beliefs of all other cliques in $CTF_1$ are then updated using a single round of message passing from $C_2$ to all other cliques. 
However, as seen in Figure \ref{fig:links}, there are five links in $IM$ that can be used for belief update. We use some heuristics to choose links that are used for the update procedure.

More formally, we propose a heuristic back-propagation algorithm for belief update. Starting with $k$ equal to $I_E-1$, we successively update beliefs in $CTF_{k}$.
This is done using the interface map $IM_k$ which contains links between cliques in $CTF_k$ and $CTF_{k+1}$. As seen in Section~\ref{sec:links}, each set of variables $C'$ in $IM_k$ links cliques $C$ in $CTF_k$ and $\tilde{C}$ in $CTF_{k+1}$ via link variables in set $L=C\cap C'$. 
The belief of a clique $C$ is updated via $L$ as follows.
\begin{align} \label{eq:bu}
    \beta(C) &= \left( \frac{\beta(C)}{\sum\limits_{C\setminus L}\beta(C)} \right)  \sum \limits_{{\tilde{C}} \setminus {L}} \beta(\tilde{C})
\end{align}
This is followed by one round of message passing from $C$ to all other cliques in the CT containing $C$. After this step, the beliefs of the link variables in $CTF_{k}$ agree with the beliefs in $CTF_{k+1}$ only over link variables in $L$. Belief update of other variables is approximate.


There are multiple links via which the beliefs $CTF_{k}$ can be updated. To ensure that beliefs of all CTs in $CTF_{k}$ are updated, at least one link must be chosen for each CT. It is also clear that more than one link may be required since variables that have low correlations in $CTF_{k}$ could become tightly correlated in $CTF_{k+1}$ (due to the additional evidence added, for example).
Empirically, we have found that updating via all links gives the best result. But it is expensive, since each update via a link requires a round of message passing in $CTF_{k}$. Based on results over many benchmarks, we use the following heuristics to choose and schedule links for backward belief update. 
\begin{enumerate}[label=\arabic*.,leftmargin=*,ref=\arabic*]
\item We first find the difference in the normalized marginals of all the link variables in both $CTF_{k}$ and $CTF_{k+1}$.  Link variables for which this difference is less than a threshold are discarded. For each remaining link variable, we find the  clique $C$ in $CTF_{k}$ that contains the maximum number of link variables. The minimum set of cliques covering these remaining link variables is chosen for belief update. \label{st:cl1}
\item The updated beliefs depend on the order in which the links are used for update. Based on the difference in marginals, we form a priority queue with the cliques containing link variables that have the lowest change in marginals having the highest priority. This is to make sure that large belief updates do not get over-written by smaller ones. This could happen for example, if two variables, $v_1$ and $v_2$, that are highly correlated in $CTF_k$ become relatively uncorrelated in $CTF_{k+1}$ due to the approximation. Assume that evidence added to  $CTF_{k+1}$ affects $v_1$ but not $v_2$. A belief update via the link containing $v_1$ will make sure that its belief is consistent in $CTF_k$ and $CTF_{k+1}$. Later, if we perform a belief update using a link containing $v_2$, the previous larger belief update of $v_1$ will be overwritten by something smaller since the belief of $v_2$ is not very different in the two CTFs. 
\end{enumerate}
The computational effort for belief update increases with $I_E$. Therefore, we prioritize addition of evidence variables while construction of CTs so that they are added as early as possible while building the CTFs. 

\subsection{Algorithm for approximate inference}

\begin{algorithm}[!h]
	\scriptsize
    \caption{InferPRandMAR($L_D, query$)}
	\label{alg:answerQueries}
	\begin{algorithmic}[1]
		\Require ~$L_D$: List of SLCTFs corresponding to all DAGs in $BN$ \newline
        \indent$query$: Prior marginals $MAR_p$, Partition function $PR$, Posterior marginals $MAR_e$
		\State Initialize: \indent$MAR = <>$ \Comment{{\color{teal!70} \scriptsize Map $<variable: margProb>$}}
        \newline\indent\indent\indent$PR = 1.0$ \Comment{{\color{teal!70} \scriptsize Partition Function}}
        \For {$L_{CTF},L_{IM}, I_E \in L_D$}
        \State$G\gets $ DAG corresponding to $L_{CTF}$
        \LineComment{{\scriptsize \color{teal!70}Infer prior marginals}}
        \If {$query==MAR_p$}
            \For {$v\in G$}
                \State Choose $CTF_j$ s.t. $\exists ~C\in CTF_j~s.t.~v\in C$
                \Comment{{\color{teal!70} \scriptsize Choose a clique containing $v$ from any CTF}}
                \State $MAR[v] = \sum\limits_{x\in C\setminus v}~\sum\limits_{s\in x.states}~\beta_C$
                \Comment{{\color{teal!70} \scriptsize Get a marginal of $v$}}
            \EndFor
        \LineComment{{\scriptsize \color{teal!70}Infer partition function}}
        \ElsIf {$query==PR$}
            \State $C\gets$ Choose any clique in $L_{CTF}[-1]$ \Comment{{\color{teal!70} Choose a clique from the last CTF in the sequence}}
            \State PR = PR $\times \sum\limits_{v\in C}~\sum\limits_{s\in v.states} ~\beta_C$  \Comment{{\color{teal!70} \scriptsize Sum over all beliefs in the clique to get the normalization constant}}
        \LineComment{{\scriptsize \color{teal!70}Infer posterior marginals}}
        \ElsIf {$query==MAR_e$}
            \LineComment{{\color{teal!70} \scriptsize Infer marginals of variables in CTFs in  $\{CTF_k, k\geq I_E\}$}} 
            \FFor {$k \geq I_E$} $MAR[v] \gets$ Infer marginals of variables in these CTFs using any containing clique \EndFFor
            \LineComment{{\color{teal!70} \scriptsize Perform belief update for CTFs in $\{CTF_k, k<I_E\}$}}
            \For {$k \in I_E-1$~ down~ to~ 1}
                \State $IM_{k} \gets L_{IM}[k]$ \Comment{{\color{teal!70} \scriptsize Get interface map for $CTF_k$}}
                \State $CTF_{k} \gets L_{CTF}[k]$; $CTF_{k+1} = L_{CTF}[k+1]$
                \State $L_l \gets $ Ordered list of links in $IM_{k}$ selected using heuristics in section \ref{sec:pm}
                \LineComment{{\scriptsize \color{teal!70}Update belief using back-propagation via all selected links}}
                \For {$(C, C', \tilde{C}) \in L_l$} 
                    \State Update $\beta(C)\in CTF_{k}$ based on $\beta(\tilde{C}) \in CTF_{k+1}$  using link variables $L=C\cap C'$ as in Equation~\ref{eq:bu}
                    \State Update belief of all other cliques in $CTF_{k}$ using single pass message passing  with C as the root node
                \EndFor
            \EndFor
             \LineComment{{\color{teal!70} \scriptsize Infer marginals of variables in CTFs $\{CTF_k, k<I_E\}$ using the first CTF in which $v$ is added }}
            \State $MAR[v]\gets$ Find marginal of $v$ from $CTF_j$ s.t. $v\in CTF_{k}, v \not\in CTF_{k-1}~~~~~\forall v\in\{CTF_{1},\hdots,CTF_{I_E-1}\}$
        \EndIf
        \EndFor
    \end{algorithmic}
\end{algorithm}

Algorithm~\ref{alg:answerQueries} describes the approximate inference algorithms. 
Based on Proposition~\ref{pr:prior}, the prior marginal of a variable $v$ is computed by finding a clique containing the variable and summing the clique belief over the states of all the other variables in the clique. 
Based on Theorem \ref{pr:pf2}, the partition function is  computed as the product of normalization constants of the last CTFs in the sequence corresponding to each DAG in the BN. The posterior beliefs are obtained in two steps. Using Theorem~\ref{th:post1}, the marginals of all variables in $\{CTF_k, k\geq I_E\}$ is obtained similar to prior beliefs. As explained, beliefs of CTFs {$\{CTF_k, k< I_E\}$ need to be updated in order to find the marginals of the variables contained in these CTFs. Starting with $k=I_E-1$, we update beliefs of $CTF_k$  based on beliefs in $CTF_{k+1}$. Links for belief update are chosen based on heuristics described in Section~\ref{sec:pm}. For each link, the belief of the corresponding clique $C$ in $CTF_k$ is updated using Equation~\ref{eq:bu}. Beliefs of all other cliques in $CTF_k$ are updated using single pass message passing with clique $C$ as the root node. 

\textbf{Note:} If all evidence variables are present in the first CTF, belief update is not needed and posterior beliefs are the same across CTFs. Otherwise, belief update is needed before finding the marginals. The posterior marginals of variables obtained after belief update are {\it{not}} the same in all CTFs in which they are present. In our algorithm, they are inferred in the first CTF in which they are introduced. This is to make sure that they are inferred after all their successors have been updated.

\section{Solution guarantees and complexity} 
\subsection{Solution guarantees}\label{sec:vallim}
In this section, we address the following question - Is IBIA guaranteed to give a solution?
IBIA uses two user-defined clique size constraints, $mcs_p$ and $mcs_{im}$. 
$mcs_p$ is the bound on the maximum clique size in any CTF in the sequence. 
$mcs_{im}$ is the bound on the maximum clique size in the approximate CTF that serves as the starting point for the next CTF.
The clique size in the approximate CTF should be such that atleast one new variable can be added to it without exceeding the constraint $mcs_p$. 
Each variable in the BN is associated with a CPD whose size is dependent on the number of parent variables and the corresponding domain sizes.
Since each CPD is assigned to a clique in the CTF, the clique size constraints should be ``CPD-size'' aware, by which we mean the following.
\begin{itemize}
	\item $mcs_p$ is large enough to accommodate the variable with the largest number of states in the CPD.
    \item $mcs_{im}$ is a soft constraint that can be reduced to add variables with large number of states in the CPD.
\end{itemize}

When no evidence variables are present, we do not have the constraint that a connected CT should remain connected after the approximation, as discussed in~\ref{sec:choice}. Therefore, it is always possible to reduce clique sizes to $mcs_{im}$ using local marginalization, independent of the minimum sepset size.
If we have large cliques containing only interface variables, the interface variables with least $maxMI$ are removed from all cliques and are added to the CTF as independent nodes (lines 24-27, Algorithm~\ref{alg:ApproximateCTF}).
Therefore, in this case, IBIA is guaranteed to give a result if the clique size constraints are CPD-size aware, as described above.



In the presence of evidence, no such guarantees can be provided.
This is because, in this case, a connected CT should remain connected while constructing the CTFs.
This means that we cannot locally marginalize variables for which the minimum sepset size is one. Similarly, it is not possible to remove interface variables and add them as independent cliques (lines 20-23, Algorithm~\ref{alg:ApproximateCTF}). The algorithm will therefore fail to give a solution under the following condition - there are cliques of size $mcs_p$ that contain only interface variables and none of these interface variables are present in any clique that has size $mcs_{im}$ or lower. 
Since it is enough for the interface variables to be retained in a single clique, it is possible to circumvent this failure by modifying the algorithm so that it locally marginalizes a set of interface variables from each of the cliques of size $mcs_p$.
However, we have not implemented it, as it happens very rarely. 
Out of the 500+ benchmarks we tested, it occurs only in 5 testcases.
\subsection{Complexity}
Let $T$ be the number of topological levels in the $BN$ and $N_{CTF}$ be the number of CTFs in the SLCTF generated by our algorithm. To construct each CTF, we perform three main steps. We now discuss the worst-case complexity of each of these steps.  
\begin{enumerate}
	\item Build CTF:
        In each step, we add active variables at a particular topological level (as shown in Figure~\ref{fig:build1}). A new clique is added to the CT for variables belonging to Cases 1 and 2 in Algorithm \ref{alg:BuildCTF}, which is an O(1) operation. Variables that belong to Case 3 are grouped into subsets, with each subset requiring the modification of disjoint subgraphs of the CT.
        The complexity of modification depends on the number of subgraphs and the cost of re-triangulating each subgraph.  
        In the worst case, we get a single subgraph that contains all the cliques in the CTF and there are no retained cliques. The cost of re-triangulation ($Cost_{R}$) using any of the greedy search methods is polynomial in the number of variables in CTF~\cite<Chapter 9 of>{Koller2009}. Hence, the worst-case complexity is upper bounded by {$O(N_{CTF}\cdot T\cdot Cost_R)$}. Generally, the number of computations required  is much lower since there are many retained cliques  and different subsets of variables to be added impact disjoint subgraphs of the existing CTs.  There is also considerable scope for parallelism here since each impacted subgraph can be processed independently.
    \item Inference and Approximation: Since we use exact inference to calibrate the clique-tree, the complexity of inference in each CTF is $O( 2^{mcs_p})$. Approximation involves summing out variables from a belief table. Once again, this is  $O(2^{mcs_p})$. The overall complexity is therefore $O(N_{CTF} \cdot 2^{mcs_p})$.

    \item Complexity of belief update via back-propagation: Each step in this involves updating belief of a single clique via the link variables, followed by one round of message passing to re-calibrate the entire clique tree. Let $N_l$ is the maximum number of links between adjacent CTFs. In the worst case, we back-propagate beliefs from the last to the first CTF via $N_l$ links. Therefore, the worst-case complexity is $O(N_{CTF} \cdot N_l\cdot 2^{mcs_p})$. 
\end{enumerate}

\section{Comparison of IBIA with related methods}
\subsection{Comparison with related work on incremental construction of CTs}
Incremental methods for CT modification have been explored in some previous works~\cite{Draper1995,Darwiche1998,Flores2002}. 
In \citeA{Draper1995}, incremental addition of links is performed by first forming a cluster graph using a set of rules and then converting the cluster graph into a junction tree. 
Although several heuristic-based graph transformations are suggested, a difficulty is to choose a set of heuristics so that clique size constraints are met. Also, there is no specific algorithm to construct the CT. 
A preferable method would be to make additions to an existing CT. Dynamic reconfiguration of CTs is explored~\cite{Darwiche1998}, but it is specific to evidence and query based simplification.  
A more general approach using the Maximal Prime Subgraph Decomposition (MPD) of the BN is discussed in \citeA{Flores2002}. In this method, the CT is converted into another graphical representation called the MPD join tree which is based on the moralized graph. When variables are added, the minimal subgraph of the moralized graph that needs re-triangulation is identified using the MPD tree. The identified subgraph is re-triangulated, and both the CT and MPD join trees are updated. In contrast, our method,
\begin{itemize}
    \item Requires a lower effort for re-triangulation. This is because the minimal subgraph that is re-triangulated is not the modified moralized graph, but a portion of the modified chordal graph corresponding to the CT (which we have denoted as the elimination graph). Moreover, as opposed to \citeA{Flores2002}, the subgraph identified using our method need not always contain all variables present in the impacted cliques of the CT. 
	\item Eliminates the memory and runtime requirements for maintaining additional representations like the moralized graph and the MPD join tree. Our method identifies the minimal subgraph to be re-triangulated directly from the CT, triangulates it and updates the CT. No other representation of the BN is needed.
\end{itemize}

\subsection{Comparison with related work on approximate inference}\label{sec:relatedApprox}
We compare our algorithm with existing methods in terms of (a) bounded cluster sizes, (b) consistency of beliefs after approximation and (c) trade-off between accuracy and runtime. 

%

    There are multiple methods that have been proposed in the literature to bound clique sizes. One method is a  straightforward scope or structure-based partitioning that is used in mini-bucket elimination and some region graph based methods. An alternative is a factor based technique, which is  an iterative process that involves multiple rounds of structure/region reconstruction and approximate belief estimation until ``good'' regions are identified. Either approximate factorized messages are used or iterative BP algorithms are run on relaxed structures or local message passing/relative entropy computations are used to get approximate beliefs.     
Our method approaches the problem in an entirely different way. Unlike these methods,  we focus on the clique trees in our method, constructing them incrementally. Our method simply keeps adding variables to the clique tree until the bound is reached. We do not explicitly try to identify ``good regions'' or nodes for duplication (as in the relax-compensate-recover method). 
In IBIA,  nodes are added to an existing CTF until the maximum permissible clique size bound is reached. Following this, we calibrate the CTF using the standard two-pass BP algorithm and use calibrated beliefs for approximation. This is done once for each CTF in the sequence. 

Another approach to bound clique sizes is to use thin junction tree based methods~\shortcite{Bach2001,Elidan2008,Dafna2009,Scanagatta2018} that simplify the network by identifying ``dominant" features (nodes and edges in the graph). The remaining features are ignored. 
Feature selection is based on the gain in KL divergence of the overall distribution or using the pairwise mutual information between variables.
In contrast, in IBIA, a sequence of CTFs is constructed considering all nodes and edges in the BN.
After each CTF is built, it is approximated to obtain a CTF with reduced clique sizes that forms the starting point for the next CTF. 
To get good accuracies, the approximate CTF should preserve the joint distribution over the interface variables as much as possible. Therefore, we choose variables that have the least pairwise mutual information with the interface variables for approximation.

Exact inference using multiple CTs corresponding to overlapping partitions of the BN has been explored in  \citeA{Xiang1993}, \citeA{Lin1997} and \citeA{Xiang2003}.
In~\citeA{Lin1997}, inference is performed over multiple overlapping subnetworks of the BN obtained using relevance based decomposition. 
In \citeA{Xiang1993}, the BN is partitioned into multiple sections and clique trees for each section are built using co-operative triangulation.  
Similar ideas have been used in~\citeA{Bilmes2010} for dynamic graphical models. 
In all these methods, since the partitioning is performed at the network level, clique sizes can possibly be reduced, but it is not
possible to guarantee bounds on the sizes. Finding partitions that meet the clique size bounds just based on the structure of the BN, would involve repeated re-triangulations of a large number of candidate sections, which is computationally infeasible. 

    The idea of constructing a sequence of CTs is also used in the interface algorithm~\cite{Murphy2002} for inference in the 2T-BN representation of the dynamic Bayesian networks (DBN), which models Markov processes. 
    This means that the children of nodes in time slice $t$ are only present in time slice $t+1$.
Variables in time-slice $t$ that have children in time-slice $t+1$ are referred to as interface nodes. 
In contrast, IBIA is used for inference in static BNs where the children of interface nodes can be present in any subsequent CTFs, which makes the problem harder.
In the interface algorithm, clique trees are constructed for each time slice.
For exact inference, all interface variables must be contained in a single clique, which makes it intractable for most networks. In the factored frontier (FF)~\cite{Murphy2001} method, the joint distribution over the interface variables is approximated as the product of marginals of individual variables. An extension is the Boyen-Koller (BK) approximation~\cite{Boyen1998}, where the joint distribution is modeled using clusters containing interface variables.
The main problem with both BK and FF approximations is that they do not guarantee that the clusters containing the interface variables form a connected CT. Therefore, these approaches cannot be used within the IBIA framework for inference of PR or posterior marginals since evidence variables could be present in any CTF. 
Moreover, the approximate CTF in IBIA contains both interface and non-interface variables, which allows for a better approximation of the joint distribution of the interface variables. 
The IBIA framework can be easily extended for dynamic BNs, which could help improve the accuracy of the solution obtained.


A big advantage of our method is the ease of approximating a clique tree for use as the starting point for the next CTF, in a way that consistency conditions are satisfied. Since we always marginalize beliefs by summing out variables, the approximation preserves within-clique beliefs (see Proposition~\ref{pr:approx2}). In the absence of evidence, this automatically ensures consistency of singleton marginals. Marginal consistency is ensured via edge parameters in \citeA{Choi2006}. In our algorithm, not only the singleton beliefs but the joint beliefs of all link variables present in a clique are consistent across adjacent CTFs in the SLCTF. 
Our method can also be viewed in terms of clique duplication (instead of node duplication). However, instead of deleting edges we reduce the size of the clique by summing over the states of some of the variables present in cliques with size greater than a threshold. 
The other consistency condition in \citeA{Choi2006} is to take care of ``lost evidence''. They fix this by adding auxiliary evidence nodes, whose posterior marginal is specified in terms of posterior beliefs of evidence, given the clone node. We do this in a different way. In our algorithm, ``lost evidence'' for a CTF would be the evidence variables added in subsequent CTFs. We do not add auxiliary evidence nodes. Rather, we update beliefs via back-propagation. We use inferred marginals in adjacent CTFs to guide the choice and sequence of links via which this updating is done. Our belief update method, however is heuristic and approximate. We do not have consistency of posterior marginals across CTFs for all variables. The posterior beliefs are inferred in the first CTF in which the variable is added in order to make sure it is inferred after its successors are updated. 

Like some of the other methods, we can trade-off time and space complexity for accuracy. Many of the factor based clustering techniques to improve accuracy require multiple iterations involving candidate region selection and belief estimation. The number of such iterations performed controls the accuracy of the method.
In IJGP and WMB, the trade-off is based on a single parameter making it very convenient. Similarly, the trade-off is easy to achieve in IBIA, since we have only two parameters. 
Clique sizes are limited to $mcs_p$ in each CTF and factor based information is used to reduce the size of the cliques to $mcs_{im}$ during approximation. The reduction in clique sizes is done after exact join tree based inference, but \textit{within a CTF containing bounded clique sizes}. Higher values of $mcs_p$ result in better accuracies but increased complexity of exact inference in each CTF due to larger clique sizes. $mcs_{im}$ controls the amount of approximation. Lower values of $mcs_{im}$ imply more aggressive approximations, but generally, a smaller number of CTFs in the sequence. As explained in Section~\ref{sec:interfaceModel}, it is not used as a hard constraint in our algorithm.
Lower or higher values are also possible depending on the size of the CPDs of variables.



\section{Results}\label{sec:results}
In this section, we evaluate the algorithms in IBIA. We first evaluate the performance of the proposed incremental CT construction method and discuss the trade-off between runtime and accuracy due to the two user-defined parameters $mcs_p$ and $mcs_{im}$. We then evaluate our method against other methods for approximate inference. 
All experiments were carried out on a 3.7-GHz Intel i7-8700 Linux system with 64-GB memory. 
\subsection{Evaluation setup}
In this section, we summarize the datasets, error metrics and the pre-processing steps performed on the input BN. 
\subsubsection{Datasets}\label{sec:dataset}
Table~\ref{tab:benchmark} lists the benchmark sets used in this work. Benchmark sets $BN$, $Pedigree$, $Relational$, $Grid\mbox{-}BN$ have been included in several UAI approximate inference challenges~\cite{UAI2010,UAI2014} and the Probabilistic Inference Challenge~\cite{PIC}. The input files were downloaded from websites hosted by~\citeA{IhlerURL} and~\citeA{MpeURL}. $Bnlearn$ is a set of miscellaneous discrete BNs provided with the bnlearn package for Bayesian learning and inference~\cite{BnlearnURL}. Table~\ref{tab:benchmark} contains the number of instances and some statistics for each set, namely,  the average number of variables, average maximum variable domain cardinality and the average maximum factor scope size. For each set, the checkmarks show the queries that are possible. Priors could not be inferred for benchmark $Pedigree$ since the available BNs have already been simplified based on evidence states. 
For the benchmarks $Grid\mbox{-}BN$ and $Bnlearn$, only priors can be inferred since no standard evidence files are available. 
Benchmark set $Relational$ includes four subsets of networks namely, $mastermind$, $blockmap$, $students$, and $fs$.
Most of these benchmarks have a significant degree of determinism. 
\begin{table}[!htp]\centering
    \caption{Number of instances (\#Inst), average number of variables ($N_v$), the average maximum variable domain cardinality ($D_{max}$), and average maximum factor scope size ($S_{max}$) for different benchmark sets. Checkmarks are used to indicate queries that are evaluated for each benchmark set.
    }
    \label{tab:benchmark}
\scriptsize
\begin{tabular}{ccccccccc}\toprule
    &\multirow{2}{*}{\#Inst} & \multirow{2}{*}{Avg $N_v$} & \multirow{2}{*}{Avg $D_{max}$} &\multirow{2}{*}{Avg $S_{max}$} & \multicolumn{3}{c}{Query} \\\cmidrule{6-8}
    &&&&&$PR$&$MAR_e$ & $MAR_p$ \\\midrule
    BN\_UAI (BN)*  &119 &754 &11.6 &7.2& \cmark & \cmark & \cmark\\
    Pedigree  &22 &917 &5 &4.5 & \cmark & \cmark & \\
    Relational  &395 &16351 &2 &3 & \cmark & \cmark & \cmark\\
    Grid-BN  &32 &649 &2 &3 & &  & \cmark\\
    Bnlearn  &26 &256 &16.1 &4.2 & &  & \cmark\\
\bottomrule
\end{tabular}\\
{\footnotesize*We refer to the set BN as BN\_UAI in our discussion since we use BN for Bayesian Networks.}
\end{table}


\subsubsection{Error Metrics}
Wherever possible, the exact probabilities were estimated using the ACE, which implements the Weighted Model Counting technique~\cite{Chavira08}.
The following error metrics were used to evaluate our algorithm.
\begin{itemize}
    \item Query: Partition Function ($PR$)
     \begin{itemize}
        \item \textit{Absolute error in log PR}: $\Delta_{Alg}= |\log PR_{Alg} - \log PR_{ACE}|$
    \end{itemize}
        \item Query : Posterior and prior marginals $MAR_e,~MAR_p$\\
    For instances where the exact solution can be computed using ACE, we compute the following.
	\begin{itemize}
		\item \textit{Maximum Error}(max-error): Modulus of the maximum error in probability over all non-evidence variables and states.
		\item \textit{Root Mean Square Error} (RMSE): Root mean square error in the probability of all states of all non-evidence variables.
        \item \textit{Mean KL distance} ($KL_{mean}$): 
        It is computed as the average KL distance between the exact (P) and approximate (Q)  marginal probabilities as follows.
		\begin{align*}
            KL_{mean}=\frac{1}{N_s} \sum\limits_{s\in N_s} P(s)~\log \frac{P(s)}{Q(s)}
		\end{align*}
		where, $N_s$, is the total number of states over all variables.
    The average KL distance is better than RMSE when the dynamic range of the probabilities is large. On the other hand, the KL distance marks a state as accurate if $P(s) = 0$ and $Q(s) \not= 0$. For states where the $Q(s)=0$ and $P(s)\not=0$, the distance becomes infinite. If this happens for even one of the states of the benchmark, the entire benchmark will be assigned a score of zero. To avoid this, for these states, we set  $Q(s)=10^{-16}$ while computing KL distance. RMSE does not have this problem. 
        \item \textit{Maximum KL distance} ($KL_{max}$):
        It is computed as the maximum KL distance between the exact (P) and approximate (Q)  marginal probabilities as follows.
		\begin{align*}
            KL_{max}=\max\limits_{s\in N_s} P(s)~\log \frac{P(s)}{Q(s)}
		\end{align*}
		where, $N_s$, is the total number of states over all variables.
        \item \textit{Score}~\cite{Mateescu2010}:  It is defined as $Score=10^{-KL_{mean}}$.
		It increases with the accuracy of estimation and is equal to one when the estimated marginals are exact. If a solver is not able to solve the problem within a time limit, it is assigned a score of zero. To evaluate performance over a benchmark set, we used the method in \citeA{Mateescu2010}. The sum of the scores ($SumScore$)  over all networks in the set is computed as a function of time. It is reflective of both runtimes and the error.
	\end{itemize}
    For instances where exact solutions could not be obtained using ACE, we use IBIA as the baseline. The metrics are a measure of the difference in the results obtained using IBIA and another approximate method. The two metrics used are
	\begin{itemize}
        \item \textit{Maximum absolute difference} ($Max~\Delta_{MAR}$): It is the modulus of the maximum difference in marginal probabilities over all variables and states.
    \item \textit{Average absolute difference} ($Avg~\Delta_{MAR}$): It is the average absolute difference in marginal probabilities over all variables and states. 
	\end{itemize}
\end{itemize}

We have used ACE and methods in Merlin and libDAI tools for comparison with IBIA.
Each tool reports the marginals using a different number of precision digits.\footnote{The format used by various tools for MAR is as follows.
ACE - number format with 17 decimal places.
libDAI - scientific format with 3 decimal places (X.XXXE-Y).
Merlin - number format with 6 decimal places
}
We use the values reported as is and compute various metrics. While reporting, we round off max-error, RMSE, $KL_{mean}$ and $KL_{max}$ to three decimal places if the corresponding value is greater than $0.001$. Otherwise, we report the results in scientific format (X$\times10^{-Y}$).
    
\subsubsection{Simplification of the network}\label{sec:simplifyBN}
All tools perform some simplification to remove some of the determinism in the network, especially in the presence of evidence.
Both determinism and local structure have been exploited very efficiently in ACE~\cite{Chavira2006,Chavira2005}.
libDAI has a switch $surgery$, which reduces the factor graph, when set to a non-zero value. However, the exact reduction methods used in these tools are unknown.
Similarly, it is not clear what simplification methods are included in Merlin, although \citeA{Mateescu2010} indicate that IJGP performs SAT-based variable domain pruning.

For a fair comparison, we also simplified the available input BN using the following steps. The first step was to simplify the network based on the fixed state of the evidence and other variables. 
For each such variable, we removed the outgoing edges and simplified the CPDs of the variable and its children. If the reduced CPD of the variable contains all ones, it does not affect the product. Moreover, the variable is independent of its parents. Therefore, all incoming edges to this variable were removed.
If the reduced CPD fixes the state of the parent variables, the same procedure is repeated for the parent variables. For example, if the reduced CPD of a variable with two parents is (0,0,0,1), this implies that the parent variables are fixed to state 11. This simplification is repeatedly performed until there are no fixed variables in the BN.
Additionally, if a variable $X$ is conditionally independent of its parent $P$ given all other parents, the edge between variable $X$ and its parent $P$ is removed and the CPD of $X$ is reduced.

We also did structural simplification for variables with a single parent where the CPD enforces either equality (buffer) or negation (inverter) between the variable and its parent. All such nodes in the BN were collapsed and the children were connected to the source node. The  CPDs of the children are modified to take care of possible inversion. Even though this step results in an increased out-degree for the source node, it can potentially reduce the re-convergence depth (cycle length), leading to a lower tree-width. 

In all our experiments, we used the available network files directly as input to all tools. The runtimes indicated include the time required for simplification.

\subsection{Evaluation of Algorithms in IBIA}
In this section, we first evaluate our technique for incremental construction of clique trees. We also study the trade-off between accuracy and runtime required for estimation of PR and marginals. Exact solutions were obtained using ACE~\cite{Chavira08}. The error metrics used in this evaluation are max-error and RMSE for $MAR_e$ and $MAR_p$ and absolute error in $\log(PR)$. Based on this evaluation, we fix the values of $mcs_p$ and $mcs_{im}$ to be used for further comparisons.

\subsubsection{Evaluation of Incremental CT construction}
We first evaluated our algorithm for incremental construction of the CT in terms of the maximum clique size. We used the following method for evaluation.
For a given $mcs_p$, we used Algorithm \ref{alg:BuildCTF} to construct the first CTF in the SLCTF. For comparison, we used a CTF obtained using full compilation of all the variables in this CTF. This is done as follows. We first identified the subgraph of BN over all variables in the first CTF and compiled the entire subgraph using variable elimination \cite{Zhang1996,Koller2009}. The elimination order is found using the `min-fill' metric, and the metric `min-neighbors' is used in case of a tie~\cite{Koller2009}. Re-computing the number of fill-in edges each time a variable is eliminated increases the execution time. Therefore, we adopted the methodology suggested in~\shortciteA{Kalev2011} to compute only the change in the number of fill-in edges.
 The input BN was used without any evidence-based simplifications.
\begin{table}[htb]\centering
    \caption{Comparison of the maximum clique size obtained after incremental construction ($mcs_{ibia}$) of the first CTF in the SLCTF with that obtained after full compilation of the corresponding BN subgraph ($mcs_{f}$). $N_v$ indicates the number of variables in the first CTF. For each benchmark, we show results for CTFs obtained using different clique size constraints. Entries are marked with `-' if $mcs_p$ is less than maximum CPD size. $\Delta=mcs_{ibia}-mcs_f$
    } 
\label{tab:incrVStriangAll}
\scriptsize
\setlength\tabcolsep{2pt}
\resizebox{\textwidth}{!}{%
\begin{tabular}{|c|ccc|>{\columncolor{gray!30}}c|ccc|>{\columncolor{gray!30}}c|ccc|>{\columncolor{gray!30}}c|ccc|>{\columncolor{gray!30}}c|}\toprule
		\multirow{2}{*}{\textbf{Network}} &\multicolumn{4}{c|}{$\mathbf{mcs_p=10}$} &\multicolumn{4}{c|}{$\mathbf{mcs_p=15}$} &\multicolumn{4}{c|}{$\mathbf{mcs_p=20}$} &\multicolumn{4}{c|}{$\mathbf{mcs_p=25}$} \\\cmidrule{2-17}
    &$\boldsymbol{N_v}$ &$\mathbf{mcs_{ibia}}$ &$\mathbf{mcs_{f}}$ &$\mathbf{\Delta}$ &$\boldsymbol{N_v}$ &$\mathbf{mcs_{ibia}}$ &$\mathbf{mcs_{f}}$ &$\mathbf{\Delta}$ &$\boldsymbol{N_v}$ &$\mathbf{mcs_{ibia}}$ &$\mathbf{mcs_{f}}$ &$\mathbf{\Delta}$ &$\boldsymbol{N_v}$ &$\mathbf{mcs_{ibia}}$ &$\mathbf{mcs_{f}}$ &$\mathbf{\Delta}$ \\\midrule
BN\_21 &- &- &- &- &624 &13.2 &15.2 &-1.9 &627 &17.2 &16.2 &1 &2832 &20.2 &21.2 &-1 \\
BN\_49 &183 &10 &10 &0 &350 &14 &14 &0 &461 &20 &19 &1 &540 &25 &25 &0 \\
BN\_60 &201 &10 &10 &0 &257 &15 &14 &1 &323 &20 &18 &2 &341 &25 &25 &0 \\
BN\_62 &173 &10 &10 &0 &343 &14 &14 &0 &455 &20 &19 &1 &517 &24 &23 &1 \\
BN\_66 &101 &8 &8 &0 &157 &15 &15 &0 &196 &18 &17 &1 &242 &25 &25 &0 \\
BN\_51 &183 &10 &10 &0 &357 &15 &15 &0 &467 &20 &19 &1 &541 &25 &25 &0 \\
BN\_55 &217 &10 &10 &0 &254 &15 &15 &0 &294 &20 &20 &0 &309 &25 &23 &2 \\
mastermind\_03\_08\_03-0010 &376 &10 &11 &-1 &579 &15 &16 &-1 &743 &20 &20 &0 &817 &22 &19 &3 \\
mastermind\_03\_08\_04-0010 &372 &10 &10 &0 &748 &15 &15 &0 &1051 &20 &21 &-1 &1204 &25 &24 &1 \\
mastermind\_03\_08\_05-0010 &339 &6 &6 &0 &591 &15 &15 &0 &1135 &18 &18 &0 &1503 &25 &24 &1 \\
mastermind\_04\_08\_03-0010 &425 &10 &11 &-1 &632 &15 &16 &-1 &796 &20 &21 &-1 &997 &25 &25 &0 \\
mastermind\_04\_08\_04-0010 &568 &10 &10 &0 &758 &15 &16 &-1 &1080 &20 &20 &0 &1371 &25 &28 &-3 \\
mastermind\_05\_08\_03-0010 &495 &10 &11 &-1 &630 &15 &16 &-1 &837 &20 &20 &0 &1045 &25 &30 &-5 \\
mastermind\_06\_08\_03-0010 &555 &10 &11 &-1 &728 &15 &16 &-1 &893 &20 &22 &-2 &1111 &25 &25 &0 \\
mastermind\_10\_08\_03-0010 &773 &10 &11 &-1 &941 &15 &18 &-3 &1117 &20 &20 &0 &1323 &25 &27 &-2 \\
blockmap\_10\_01-0005 &477 &10 &10 &0 &1076 &15 &16 &-1 &1602 &20 &20 &0 &2165 &25 &24 &1 \\
blockmap\_15\_01-0006 &1177 &10 &11 &-1 &1873 &15 &15 &0 &3851 &20 &21 &-1 &5001 &25 &24 &1 \\
blockmap\_10\_02-0008 &180 &10 &11 &-1 &318 &15 &15 &0 &382 &19 &19 &0 &590 &25 &25 &0 \\
90-42-5 &292 &10 &10 &0 &399 &15 &13 &2 &531 &20 &16 &4 &583 &25 &18 &7 \\
90-23-5 &171 &10 &10 &0 &206 &15 &12 &3 &235 &20 &14 &6 &268 &24 &15 &9 \\
50-18-5 &131 &10 &10 &0 &194 &15 &13 &2 &229 &20 &18 &2 &251 &24 &22 &2 \\
75-25-5 &190 &10 &9 &1 &209 &13 &11 &2 &252 &20 &18 &2 &276 &25 &20 &5 \\
90-50-5 &517 &10 &9 &1 &598 &15 &13 &2 &740 &20 &18 &2 &848 &25 &19 &6 \\
andes &147 &10 &8 &2 &206 &15 &13 &2 &212 &20 &15 &5 &219 &24 &18 &6 \\
munin1 &68 &9.2 &9.2 &0 &74 &13.3 &13.3 &0 &80 &17.6 &17.6 &0 &129 &24 &25 &-1 \\
munin4 &283 &9.6 &9.6 &0 &301 &13.9 &14.2 &-0.3 &634 &19.6 &21.4 &-1.8 &642 &20.4 &21.4 &-1 \\
diabetes &- &- &- &- &186 &13.6 &13.6 &0 &192 &19.2 &18.9 &0.4 &196 &22.7 &18.2 &4.5 \\
\bottomrule
\end{tabular}%
}
\end{table}

Table~\ref{tab:incrVStriangAll} compares the maximum clique size obtained using incremental ($mcs_{ibia}$) and full compilation ($mcs_f$), for various values of $mcs_p$ for a subset of benchmarks. The results for other benchmarks are similar. The number of variables in the corresponding BN subgraph is shown in column $N_v$. As shown in Equation (\ref{eqn:cs}), our definition for clique size is the logarithm (base 2) of the product of the domain sizes. Therefore, it is possible to get decimal values for sizes when cliques contain variables with domain size greater than 2. In general, we find that the difference, $mcs_{ibia} - mcs_f$,  grows as $mcs_p$ increases, which is to be expected. However, as long as $mcs_p$ is limited to about 25,  the difference is less than or equal to 2 in most cases. Even for testcases like blockmap\_15\_01-0006 where the number of variables is 5000, the clique sizes obtained using our method are within $\pm1$ of the size obtained using full recompilation. An exception to this are the Grid-BN benchmarks, where we find that the difference increases more rapidly. This is because these networks have a localized structure that enforces a particular elimination order when the incremental method is used. 

\subsubsection{Impact of $mcs_p$ and $mcs_{im}$ on accuracy and runtime} \label{sec:ecsc}


In this section, we evaluate the results obtained for different values of $mcs_p$, for all three queries.
We also need to choose the value of $mcs_{im}$.
In general, we would like $mcs_{im}$ to be as high as possible for better accuracies. But, we also need a sufficient margin to allow for addition of variables to the next CTF in the sequence.
Based on results from various benchmarks, we have empirically chosen  $mcs_{im}$ to be 5 less than $mcs_p$.
However, as mentioned in Section~\ref{sec:interfaceModel}, $mcs_{im}$ is a soft constraint and depends on the CPD sizes of the variables to be added.

Table~\ref{tab:mcsCmpMAR} compares the required runtime, the maximum error and the RMSE in $MAR_p$ and $MAR_e$. Table~\ref{tab:mcsCmpPR} compares the required runtime and error in $\log(PR)$ ($\Delta_{IBIA}$).  We show the results for a subset of benchmarks. The results for others are similar. The number of CTFs in the constructed SLCTF ($N_{CTF}$) is indicated in brackets. Similar to all BP based techniques, the clique size bounds should be atleast as large as the CPDs. The values are marked with a `-', if inference is not possible for the given $mcs_p$ due to the CPD size.
$mcs_p$ of 10 and 15 is not possible in testcases diabetes and barley, where the maximum variable domain cardinality is 21-67 and in testcases BN\_86, BN\_91, BN\_93, BN\_125, where the maximum in-degree is 16-17. 
\begin{table}[!htbp]\centering
    \caption{Comparison of maximum error, RMSE and required runtime (in seconds) for inference of marginals using various clique size constraints. Number of CTFs in the constructed SLCTF ($N_{CTF}$) is indicated in  brackets. Entries are marked with `-' when $mcs_p$ is less than the maximum CPD size.}\label{tab:mcsCmpMAR}
\scriptsize
\setlength\tabcolsep{2pt}
\resizebox{\textwidth}{!}{%
\begin{tabular}{|c|c|cc>{\columncolor{gray!30}}cc|cc>{\columncolor{gray!30}}cc|cc>{\columncolor{gray!30}}cc|}\toprule
    \textbf{} &&\multicolumn{4}{c|}{\textbf{Maximum Error ($\boldsymbol{N_{CTF}}$)}} &\multicolumn{4}{c|}{\textbf{RMSE}} &\multicolumn{4}{c|}{\textbf{Runtime (s)}} \\\cmidrule{2-14}
	&$\mathbf{(mcs_p, mcs_{im})}$ &\textbf{(10,5)} &\textbf{(15,10)} &\textbf{(20,15)} &\textbf{(25,20)} &\textbf{(10,5)} &\textbf{(15,10)} &\textbf{(20,15)} &\textbf{(25,20)} &\textbf{(10,5)} &\textbf{(15,10)} &\textbf{(20,15)} &\textbf{(25,20)} \\\midrule
        \multirow{21}{*}{$\mathbf{MAR_e}$} &BN\_33 &0.057 (2) &$1\times 10^{\mbox{-}15}$ (1) &$1\times 10^{\mbox{-}15}$ (1) &$1\times 10^{\mbox{-}15}$ (1) &0.005 &$1\times 10^{\mbox{-}16}$ &$1\times 10^{\mbox{-}16}$ &$1\times 10^{\mbox{-}16}$ &1 &1 &1 &1 \\
        &BN\_42 &0.186 (5) &0.131 (5) &0.052 (4) &2E-04 (3) &0.029 &0.015 &0.008 &8e-6 &2 &4 &9 &8 \\
        &BN\_64 &0.066 (8) &0.036 (8) &0.009 (8) &0.008 (7) &0.012 &0.005 &0.001 &0.001 &9 &11 &42 &752 \\ 
&BN\_49 &0.018 (3) &0 (2) &0 (1) &0 (1) &0.002 &0 &0 &0 &2 &2 &1 &1 \\
&BN\_60 &0.403 (7) &0.446 (6) &0.054 (5) &0.008 (5) &0.105 &0.122 &0.004 &$7\times 10^{\mbox{-}4}$ &7 &5 &2 &25 \\
&BN\_66 &0.086 (7) &0.069 (8) &0.064 (7) &$4\times 10^{\mbox{-}4}$ (5) &0.012 &0.005 &0.005 &$8\times 10^{\mbox{-}5}$ &3 &2 &7 &45 \\
&BN\_55 &0.049 (4) &0.019 (3) &0.006 (3) &$1\times 10^{\mbox{-}16}$ (2) &0.006 &0.001 &$9\times 10^{\mbox{-}4}$ &$8\times 10^{\mbox{-}18}$ &3 &1 &7 &6 \\
&BN\_86 & - & - &$6\times 10^{\mbox{-}9}$ (2) &$1\times 10^{\mbox{-}13}$ (1) &- & - &$3\times 10^{\mbox{-}10}$ &$6\times 10^{\mbox{-}15}$ &- & - &3 &3 \\
&BN\_91 &- & - &$5\times 10^{\mbox{-}7}$ (3) &$6\times 10^{\mbox{-}14}$ (1) &- & - &$2\times 10^{\mbox{-}8}$ &$5\times 10^{\mbox{-}15}$ &- & - &3 &4 \\
&BN\_93 &- & - &$6\times 10^{\mbox{-}14}$ (2) &$6\times 10^{\mbox{-}14}$ (1) &- & -&$8\times 10^{\mbox{-}15}$ &$8\times 10^{\mbox{-}15}$ &- & -&2 &3 \\
&pedigree1 &0.056 (5) &0.059 (4) &0.005 (2) &$1\times 10^{\mbox{-}14}$ (1) &0.01 &0.00 &$7\times 10^{\mbox{-}4}$ &$1\times 10^{\mbox{-}15}$ &3 &2 &1 &1 \\
&pedigree23 &0.131 (5) &0.128 (5) &0.099 (3) &0.392 (3) &0.017 &0.019 &0.013 &0.025 &3 &3 &3 &23 \\
&pedigree18 &0.339 (7) &0.186 (7) &0.238 (6) &0.171 (6) &0.046 &0.026 &0.026 &0.021 &23 &24 &31 &274 \\
&pedigree30 &0.266 (6) &0.213 (8) &0.257 (6) &0.249 (6) &0.046 &0.026 &0.021 &0.027 &25 &27 &39 &404 \\
&fs-07 &0.016 (4) &0.008 (4) &0.007 (2) &0.006 (2) &0.007 &0.005 &0.003 &0.003 &5 &3 &3 &11 \\
&mastermind\_04\_08\_04-0003 &0.043 (3) &0.059 (2) &0.008 (2) &$3\times 10^{\mbox{-}16}$ (1) &0.009 &0.008 &$7\times 10^{\mbox{-}4}$ &$6\times 10^{\mbox{-}17}$ &3 &1 &2 &3 \\
&mastermind\_10\_08\_03-0005 &0.013 (5) &0.004 (5) &0.002 (4) &0.001 (3) &0.002 &$6\times 10^{\mbox{-}4}$ &$2\times 10^{\mbox{-}4}$ &$2\times 10^{\mbox{-}4}$ &6 &2 &6 &92 \\
&mastermind\_10\_08\_03-0000 &0.014 (7) &0.015 (7) &0.011 (7) &0.008 (9) &0.003 &0.003 &0.002 &0.001 &26 &47 &24 &467 \\
&mastermind\_06\_08\_03-0000 &0.033 (7) &0.028 (5) &0.021 (4) &0.015 (3) &0.006 &0.004 &0.005 &0.003 &8 &8 &32 &163 \\
&mastermind\_04\_08\_04-0000 &0.049 (7) &0.041 (5) &0.066 (5) &0.051 (5) &0.011 &0.008 &0.014 &0.012 &4 &41 &13 &130 \\
&blockmap\_10\_03-0014 &0.505 (12) &0.088 (7) &0.035 (7) &0.033 (5) &0.054 &0.004 &0.001 &0.001 &7 &6 &12 &85 \\
\rowcolor{gray!30} &\textbf{Mean} &\textbf{0.101} &\textbf{0.080} &\textbf{0.043} &\textbf{0.043} &\textbf{0.018} &\textbf{0.014} &\textbf{0.005} &\textbf{0.004} & & & & \\\midrule
\multirow{27}{*}{$\mathbf{MAR_p}$} &BN\_91 & - & - &$3\times 10^{\mbox{-}12}$ (4) &$1\times 10^{\mbox{-}13}$ (1) & - & - &$2\times 10^{\mbox{-}13}$ &$8\times 10^{\mbox{-}15}$ & - & - &2 &11 \\
&BN\_93 & - & - &$3\times 10^{\mbox{-}11}$ (2) &$8\times 10^{\mbox{-}14}$ (1) & -& - &$1\times 10^{\mbox{-}12}$ &$9\times 10^{\mbox{-}15}$ & - & - &3 &7 \\
&BN\_125 & - & - &0.001 (10) &$2\times 10^{\mbox{-}14}$ (1) & -& - &$3\times 10^{\mbox{-}4}$ &$4\times 10^{\mbox{-}15}$ & - & - &5 &18 \\
&BN\_33 &0.160 (10) &0.002 (12) &$5\times 10^{\mbox{-}5}$ (11) &$2\times 10^{\mbox{-}7}$ (9) &0.009 &$7\times 10^{\mbox{-}5}$ &$2\times 10^{\mbox{-}6}$ &$2\times 10^{\mbox{-}8}$ &2 &2 &4 &55 \\
&BN\_37 &0.061 (11) &0.131 (11) &0.002 (10) &$5\times 10^{\mbox{-}6}$ (8) &0.005 &0.005 &$2\times 10^{\mbox{-}4}$ &$1\times 10^{\mbox{-}7}$ &2 &2 &5 &66 \\
&BN\_40 &0.049 (10) &0.009 (11) &0.001 (10) &$3\times 10^{\mbox{-}7}$ (9) &0.005 &$5\times 10^{\mbox{-}4}$ &$4\times 10^{\mbox{-}5}$ &$2\times 10^{\mbox{-}8}$ &2 &2 &5 &71 \\
&BN\_43 &0.028 (6) &0.017 (5) &0.004 (4) &$1\times 10^{\mbox{-}4}$ (2) &0.004 &0.002 &$2\times 10^{\mbox{-}4}$ &$8\times 10^{\mbox{-}6}$ &1 &1 &2 &16 \\
&BN\_51 &0.077 (5) &0.058 (4) &0.035 (3) &$1\times 10^{\mbox{-}16}$ (2) &0.013 &0.007 &0.003 &$9\times 10^{\mbox{-}18}$ &2 &2 &3 &49 \\
&BN\_55 &0.086 (8) &0.086 (9) &0.026 (7) &$8\times 10^{\mbox{-}3}$ (8) &0.01 &0.007 &0.002 &$7\times 10^{\mbox{-}4}$ &1 &1 &3 &60 \\
&BN\_62 &0.094 (7) &0.050 (4) &0.056 (3) &0.038 (3) &0.013 &0.007 &0.004 &0.002 &2 &2 &3 &25 \\
&BN\_64 &0.115 (7) &0.064 (8) &0.011 (8) &0.005 (8) &0.013 &0.007 &$9\times 10^{\mbox{-}4}$ &$6\times 10^{\mbox{-}4}$ &1 &1 &3 &65 \\
&BN\_66 &0.124 (8) &0.124 (10) &0.113 (10) &0.009 (9) &0.008 &0.007 &0.007 &$6\times 10^{\mbox{-}4}$ &1 &1 &4 &70 \\
&mastermind\_10\_08\_03-0010 &0.042 (10) &0.015 (7) &0.011 (7) &0.008 (9) &0.004 &0.003 &0.002 &0.001 &20 &46 &24 &463 \\
&mastermind\_06\_08\_03-0010 &0.041 (7) &0.026 (4) &0.021 (4) &0.015 (3) &0.007 &0.004 &0.005 &0.003 &8 &7 &32 &162 \\
&mastermind\_04\_08\_03-0010 &0.052 (4) &0.030 (3) &0.036 (3) &0.004 (2) &0.013 &0.007 &0.008 &$3\times 10^{\mbox{-}4}$ &11 &10 &7 &29 \\
&blockmap\_05\_03-0003 &0.603 (9) &0.142 (5) &$1\times 10^{\mbox{-}15}$ (5) &$1\times 10^{\mbox{-}15}$ (3) &0.168 &0.026 &$1\times 10^{\mbox{-}16}$ &$1\times 10^{\mbox{-}16}$ &1 &1 &2 &1 \\
&blockmap\_15\_01-0006 &0.602 (20) &0.485 (13) &0.189 (8) &0.113 (9) &0.193 &0.111 &0.03 &0.018 &33 &41 &77 &794 \\
&blockmap\_10\_01-0005 &0.364 (15) &0.361 (7) &0.222 (7) &$3\times 10^{\mbox{-}16}$ (5) &0.12 &0.075 &0.02 &$8\times 10^{\mbox{-}17}$ &8 &12 &23 &205 \\
&90-42-5 &0.021 (11) &0.001 (11) &0.003 (10) &$6\times 10^{\mbox{-}7}$ (11) &0.002 &$1\times 10^{\mbox{-}4}$ &$1\times 10^{\mbox{-}4}$ &$3\times 10^{\mbox{-}8}$ &2 &2 &5 &67 \\
&90-23-5 &0.011 (7) &$4\times 10^{\mbox{-}8}$ (5) &0.002 (4) &$5\times 10^{\mbox{-}8}$ (4) &0.001 &$6\times 10^{\mbox{-}9}$ &$2\times 10^{\mbox{-}4}$ &$7\times 10^{\mbox{-}9}$ &0.5 &1 &1 &11 \\
&andes &$7\times 10^{\mbox{-}4}$ (4) &$9\times 10^{\mbox{-}6}$ (3) &$1\times 10^{\mbox{-}5}$ (2) &$3\times 10^{\mbox{-}15}$ (1) &$8\times 10^{\mbox{-}5}$ &$9\times 10^{\mbox{-}7}$ &$1\times 10^{\mbox{-}6}$ &$3\times 10^{\mbox{-}16}$ &0.2 &0.3 &0.4 &2 \\
&barley & -& - &$7\times 10^{\mbox{-}4}$ (2) &$7\times 10^{\mbox{-}15}$ (1) & -& - &$4\times 10^{\mbox{-}5}$ &$5\times 10^{\mbox{-}16}$ & -& - &0.1 &0.2 \\
&munin1 &0.142 (4) &0.104 (5) &0.017 (4) &$6\times 10^{\mbox{-}4}$ (2) &0.009 &0.006 &0.002 &$5\times 10^{\mbox{-}5}$ &0.2 &0.2 &0.2 &1 \\
&munin3 &0.041 (3) &0.005 (2) &$5\times 10^{\mbox{-}4}$ (2) &$2\times 10^{\mbox{-}15}$ (1) &0.002 &$4\times 10^{\mbox{-}4}$ &$4\times 10^{\mbox{-}5}$ &$1\times 10^{\mbox{-}16}$ &1 &1 &1 &2 \\
&munin4 &0.088 (4) &0.055 (3) &$2\times 10^{\mbox{-}4}$ (2) &$2\times 10^{\mbox{-}15}$ (1) &0.004 &0.002 &$5\times 10^{\mbox{-}6}$ &$2\times 10^{\mbox{-}16}$ &1 &1 &1 &1 \\
&diabetes & - &0.043 (24) &0.049 (23) &$7\times 10^{\mbox{-}4}$ (11) & - &0.01 &0.005 &$5\times 10^{\mbox{-}5}$ & - &1 &1 &4 \\\midrule
\rowcolor{gray!30} &\textbf{Mean} &\textbf{0.133} &\textbf{0.082} &\textbf{0.032} &\textbf{0.007} &\textbf{0.029} &\textbf{0.013} &\textbf{0.003} &\textbf{0.001} & & & & \\

\bottomrule
\end{tabular}%
}
\end{table}

It is seen from  Table~\ref{tab:mcsCmpMAR} that, as expected,  both the max-error and RMSE reduce or remain almost the same, with increasing $mcs_p$ and $mcs_{im}$.
Except for a few testcases (for example, pedigree18, pedigree30), it is seen that both max-error and RMSE are quite low even when $N_{CTF}$ is large. In the benchmark diabetes for example, even with $N_{CTF}$ of $23$, the maximum error is about 0.05. The RMSE is an order of magnitude lower, indicating that majority of the variables have very low error. With a single CTF, IBIA performs exact inference. In this case, the errors are seen to be of the order of $10^{-14}$ or less, indicating accuracy of our computations. Even with multiple CTFs, the error can be quite low, as for example for error in $MAR_p$  in BN\_33 ($\approx 10^{-5}$) and blockmap\_05\_03-0003 ($10^{-15}$). This will occur if a larger number of exact marginalizations are possible.

\begin{table}[!htp]\centering
    \caption{Comparison of absolute error in $\log(PR)$ estimated with IBIA and required runtime (in seconds) for various clique size constraints. Entries are marked with `-' when $mcs_p$ is less than maximum CPD size. 
    $\Delta_{IBIA}=|\log~PR_{IBIA}-\log_{ACE}|$
    }\label{tab:mcsCmpPR}
\scriptsize
\resizebox{0.75\textwidth}{!}{%
\begin{tabular}{|c|c|cc>{\columncolor{gray!30}}cc|cc>{\columncolor{gray!30}}cc|}\toprule
    &\multirow{2}{*}{} &\multicolumn{4}{c|}{$\mathbf{\Delta_{IBIA}}$} &\multicolumn{4}{c|}{\textbf{Runtime (s)}} \\\cmidrule{2-10}
&$\mathbf{(mcs_p, mcs_{im})}$&\textbf{(10,5)} &\textbf{(15,10)} &\textbf{(20,15)} &\textbf{(25,20)} &\textbf{(10,5)} &\textbf{(15,10)} &\textbf{(20,15)} &\textbf{(25,20)} \\\midrule
\multirow{15}{*}{\bf PR}&BN\_33 &$2\times 10^{\mbox{-}7}$ &$2\times 10^{\mbox{-}7}$ &$2\times 10^{\mbox{-}7}$ &$2\times 10^{\mbox{-}7}$ &1 &1 &1 &1 \\
&BN\_49 &$7\times 10^{\mbox{-}7}$ &$7\times 10^{\mbox{-}7}$ &$7\times 10^{\mbox{-}7}$ &$7\times 10^{\mbox{-}7}$ &1 &1 &1 &1 \\
&BN\_60 &0.08 &0.1 &$1\times 10^{\mbox{-}7}$ &$1\times 10^{\mbox{-}7}$ &1 &1 &2 &25 \\
&BN\_66 &0.005 &0.001 &$5\times 10^{\mbox{-}8}$ &$2\times 10^{\mbox{-}4}$ &1 &1 &3 &36 \\
&BN\_55 &$6\times 10^{\mbox{-}4}$ &$7\times 10^{\mbox{-}7}$ &$7\times 10^{\mbox{-}7}$ &$7\times 10^{\mbox{-}7}$ &0.5 &1 &1 &6 \\
&BN\_86 & -& - &$4\times 10^{\mbox{-}7}$ &$4\times 10^{\mbox{-}7}$ & -& - &3 &3 \\
&BN\_91 & -& - &$1\times 10^{\mbox{-}7}$ &$1\times 10^{\mbox{-}7}$ & -& - &2 &4 \\
&BN\_93 & -& - &$2\times 10^{\mbox{-}7}$ &$2\times 10^{\mbox{-}7}$ & -& - &2 &3 \\
&pedigree1 &$4\times 10^{\mbox{-}3}$ &$3\times 10^{\mbox{-}6}$ &$1\times 10^{\mbox{-}5}$ &$2\times 10^{\mbox{-}7}$ &0.4 &0.4 &0.4 &1 \\
&pedigree23 &0.03 &0.07 &$9\times 10^{\mbox{-}5}$ &$1\times 10^{\mbox{-}5}$ &1 &1 &1 &4 \\
&pedigree18 &0.3 &0.006 &0.05 &0.002 &2 &2 &2 &11 \\
&pedigree30 &0.2 &0.07 &0.01 &0.01 &2 &2 &3 &15 \\
&fs-07 &1 &0.9 &0.9 &0.6 &0.4 &0.4 &1 &3 \\
&mastermind\_04\_08\_04-0003 &$2\times 10^{\mbox{-}7}$ &$2\times 10^{\mbox{-}7}$ &$2\times 10^{\mbox{-}7}$ &$2\times 10^{\mbox{-}7}$ &3 &1 &2 &3 \\
&mastermind\_10\_08\_03-0005 &$2\times 10^{\mbox{-}7}$ &$2\times 10^{\mbox{-}7}$ &$2\times 10^{\mbox{-}7}$ &$2\times 10^{\mbox{-}7}$ &6 &2 &6 &92 \\
&blockmap\_10\_01-0014 &$5\times 10^{\mbox{-}8}$ &$5\times 10^{\mbox{-}8}$ &$5\times 10^{\mbox{-}8}$ &$5\times 10^{\mbox{-}8}$ &6 &7 &13 &68 \\
&blockmap\_10\_03-0014 &0.3 &0.008 &0.007 &0.002 &4 &5 &8 &49 \\
\bottomrule
\end{tabular}%
}
\end{table}

A similar trend in $\Delta_{IBIA}$ is observed while estimating PR, as shown in Table~\ref{tab:mcsCmpPR}. Here, the errors are very low, except for fs-07. This is a relational BN, in which a lot of determinism is present, which we have not fully exploited. Note that the error in the posterior is due to both the approximation and belief update via back-propagation, whereas the error in PR is only due to the approximation technique. Therefore, we get much better accuracies for the PR than for $MAR_e$.

The runtime of IBIA includes the time required for simplification of the network, construction of all SLCTFs and approximate inference of queries.
Overall, we expect the runtime to increase with $mcs_p$. The polynomial time complexity of CT construction is dominant for lower values of $mcs_p$ and the exponential complexity of inference dominates for larger values of $mcs_p$.
This is seen for most of the benchmarks in Tables~\ref{tab:mcsCmpMAR} and~\ref{tab:mcsCmpPR}, where there is a significant jump in the runtimes due to larger inference times when $mcs_p$ is increased from 20 to 25.
There are occasional fluctuations in this trend. This happens typically when the incremental CT building algorithm struggles to add variables to a CTF due to large CPD sizes or the local structure of the network calls for repeated re-triangulation of most of the existing CT. This is seen in the evaluation of $MAR_p$ for a couple of mastermind examples when, $mcs_p$ is increased from 15 to 20.
In some cases like BN\_49, BN\_86 and pedigree1, the runtime flattens out with increasing $mcs_p$.
This is because in these cases, there is only one CTF in the SLCTF when $mcs_p$ is increased beyond a certain point.
While generally comparable, in a few cases the runtime for computation of PR is significantly lower than the run time for posterior marginals (for example in pedigree18 and pedigree30). This is because the evaluation of $MAR_e$ requires additional time for belief update via back-propagation.

It is seen from the tables that increasing $mcs_p$ beyond 20 results in a marginal decrease in error with a large increase in runtime. Based on this,  we chose $mcs_p$ of $20$ for further experiments.

%

\subsection{Comparison with other approximate methods}
In this section, we first discuss the methods used for comparison. Following this, we discuss the runtime switches and the memory utilization for different methods. Next, we give an overview of results obtained using benchmarks in all the sets. Subsequently, we show detailed results for each set of benchmarks.


\subsubsection{Methods used for comparison}
Wherever possible, the exact solutions obtained using ACE, which implements weighted model counting \cite{Chavira08}, were used for comparison. The
approximate inference algorithms used for comparison were those available in  libDAI~\cite{LibdaiPaper} and Merlin tools~\cite{Merlin}. Table~\ref{tab:expLibDAI} in Appendix~\ref{app:eval} has a summary of the results obtained using various algorithms in libDAI, in terms of error, the number of benchmarks solved and the average runtime for 90 BN\_UAI benchmarks. 
It is seen that  Fractional Belief Propagation (FBP) ~\cite{Wiegerinck2003} solves the largest number of cases among the deterministic approximate algorithms, with reasonable errors and runtime. Also, the method gives an estimate for all the queries of interest. These results agree with the results for PR reported in a recent evaluation of various approximate methods \cite{Agrawal2021}. We have also included a comparison with the method based on Gibbs sampling since it is able to solve all problems with $10^4$ samples.

Merlin includes implementations of Weighted Mini-Bucket elimination (WMB)~\cite{Liu2011} and Iterative Join Graph Propagation (IJGP)~\cite{Mateescu2010}. These two methods are based on mini-bucket heuristics. They have a single parameter ($ibound$) that controls the size of the largest permissible cluster, making it suitable for comparison with our method. We used both these methods for comparison.

Both libDAI and Merlin are robust and have been used for comparison with other methods~\cite{Agrawal2021,Lin2020}.

\subsubsection{Runtime switches}

Table~\ref{tab:toolInfo} has the details of compile and runtime options used with different inference methods \cite{ACE,LibdaiManual,Merlin}.
While the ACE solver is available as a Java binary executable, source codes written in C++ are available for libDAI and Merlin. IBIA is implemented in Python3 with NetworkX and NumPy libraries.  
With libDAI, we enabled simplification of the factor graph by setting the \textit{surgery} switch.
\begin{table}[!htp]\centering
    \caption{Setup for tools used in our experiments.}
	\label{tab:toolInfo}
	\scriptsize
	\setlength\tabcolsep{2pt}
	\resizebox{\textwidth}{!}{%
	\begin{tabular}{|l|l|l|l|}\toprule
		\textbf{Tool} &\textbf{Compiler/}    &\textbf{Algorithm} & \textbf{Runtime switches used} \\
					  & \textbf{Interpreter} &                   & \\\midrule
		IBIA   	  	  & Python v3.9 		 & IBIA				 & $mcs_p$=20, $mcs_{im}$=15 \\
					  & (Libraries: 		 &					 & \\
					  & NetworkX v2.5, 		 &					 & \\
					  & NumPy v1.19.4)       & 					 & \\ \bottomrule
		libDAI 		  & g++ v5.4 			 & uai2fg 			 & surgery=1 \\
                      &                      & GIBBS             & maxiter=10000,burnin=100,restart=10000 \\
					  &                      & FBP   			 & inference=SUMPROD, updates=SEQMAX, logdomain=0, \\
					  & 					 &      			 & tol=1e-9, maxiter=10000, damping=0.0 \\ \midrule
        Merlin 		  & g++ v5.4 			 & WMB, IJGP  		 & ibound=20, iterations=10  (WMB20, IJGP20)\\ 
                      &  			         &           		 & ibound=10, iterations=10  (WMB10, IJGP10)\\ \midrule
		ACE 		  & JRE v9-internal 	 &  ACE 			 & -Xmx50G (for both compile, evaluate) \\
					  &						 &					 & -e <evidFile> (for compile; while finding posteriors) \\ 
		\bottomrule
	\end{tabular}%
}
\end{table}

Based on results shown in Tables~\ref{tab:mcsCmpMAR} and~\ref{tab:mcsCmpPR}, we have used $mcs_p$ of 20 for IBIA. For a fair comparison of the error, we used an $ibound$ of 20 for WMB and IJGP (referred to as WMB20 and IJGP20 in this section). While this is a fair comparison when the networks have only binary variables, this puts WMB and IJGP at an advantage in terms of error for networks that contain variables with domain sizes greater than 2. However, in terms of runtime, this could be a disadvantage for WMB and IJGP since the complexity of inference is higher. 
The runtime disadvantage is partially offset by the fact that our code is implemented in Python, whereas the Merlin tools are compiled C++ codes. We also show results for WMB and IJGP with an $ibound$ of 10 (referred to as WMB10 and IJGP10 in this section), since in most cases, the effective number of binary variables turns out to be between 10 and 20.
%
In any case, it is not possible to have a completely fair comparison of the runtimes due to the variety of programming languages used for implementation. We have included the average run times for various methods mainly to show that the run time of IBIA is very competitive.

\subsubsection{Memory utilization}
The memory required by different inference algorithms is based on different parameters. In IBIA, when $mcs_p$ is set to 20, the maximum memory required to store a clique belief using the double precision format is 8MB. 
Typically, the memory required for various benchmarks is much lower than the physical memory of the system (64GB). For example, for benchmarks where inference of PR is hard (shown in Table~\ref{tab:pr.hardInstances}), the maximum memory utilization was less than 2GB when $mcs_p$ is set to 20.
For Pedigree benchmarks in this set, it was less than 512MB.
For IJGP and WMB, setting $ibound$ to 20 simply restricts the number of variables in a cluster. The memory required to store clique beliefs is dependent on the domain sizes of the variables. These methods give memory errors for some benchmarks that have variables with large domain sizes.
For ACE, some of the benchmarks did not run with the default maximum heap size of 1.5GB, therefore, we set it to 50GB.
For FBP, the clusters are only as large as the CPDs in the BN. Therefore, the memory utilization for FBP is much lower than all other methods.

\subsubsection{Overview of results for all benchmark sets}
More than 500 benchmarks available in \citeA{IhlerURL} were used for evaluation of our method. A time limit of one hour was set for each benchmark. 
Table~\ref{tab:datasetInfo} contains the number of instances of each benchmark that can be solved by each method. IBIA gives solutions for $PR$ and $MAR_e$ for 523/536 benchmarks, which is better than ACE, WMB20, and IJGP20. For $MAR_p$, IBIA solves 565/572 benchmarks, which is better than ACE, WMB20 and IJGP20 and marginally lower than FBP and GIBBS, both of which give solutions for all 572.

\begin{table}[!htb]\centering
    \caption{List of benchmarks and the corresponding total number of instances evaluated for different probabilistic queries. For each set, we compare the number of instances that successfully run with different inference algorithms.}
	\label{tab:datasetInfo}
	\setlength\tabcolsep{3pt}
	\resizebox{\textwidth}{!}{%
	\begin{tabular}{|c|c|c|c|c|c|c|c|>{\columncolor{gray!20}}c|}\toprule
        \textbf{Query} &\textbf{Benchmark} &\textbf{Instances} &\textbf{ ACE} &\textbf{ FBP} &\textbf{ WMB20} &\textbf{ IJGP20} &\textbf{GIBBS} &\textbf{IBIA} \\\midrule
        \multirow{4}{*}{$\mathbf{PR, MAR_e}$} &BN\_UAI &119 &90 &119 &106 &101 &119 &117* \\
        &Pedigree &22 &9 &22  &7 &12 &22 &22* \\
        &Relational &395 &395 &392 &227 &203 &395 &384 \\\cmidrule{2-9}
        &\bf Total & \bf 536 & \bf 494 &  \bf 533  &  \bf 340 &  \bf 316 & \bf 536 & \bf 523 \\\midrule
        \multirow{5}{*}{$\mathbf{MAR_p}$} &BN\_UAI &119 &76 &119 &105 &94 & 119&119* \\
        &Grid-BN &32 &29 &32 &32 &26 &32 &32 \\
        &Relational &395 &386 &395 &196 &170 &395 &388 \\
        &Bnlearn &26 &26 &26 &26 &26 &26 &26 \\\cmidrule{2-9}
        &\bf Total & \bf 572 & \bf 517 &  \bf 572 &  \bf 359 &  \bf 316 & \bf 572& \bf 565 \\
		\bottomrule
    \end{tabular}%
    }
	{\\ 
		\footnotesize* Three BN\_UAI and two Pedigree inferred with $mcs_{im}=14$.
	}
\end{table}

We compare the error obtained over all benchmarks using cactus plots. In a cactus plot, the values of the error metric obtained for all instances are sorted in the increasing order and plotted. Therefore, a point $(x,y)$ on the plot indicates that $x$ instances can be solved with the metric is less than or equal to $y$.

\begin{figure}[!htbp]
        \centering 
        \includegraphics[width=0.5\textwidth]{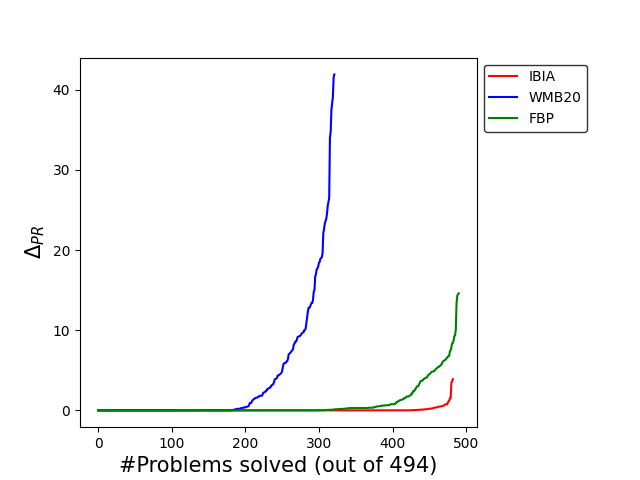}
        \caption{Cactus plot for absolute error in $\log(PR)$ for all instances where the exact solution can be obtained using ACE. $\Delta_{PR}=|\log ~PR_{Alg}-\log~PR_{ACE}|$}
    \label{fig:cactusDeltaPR}
\end{figure}
Figure~\ref{fig:cactusDeltaPR} shows the cactus plot for absolute error in $\log(PR)$ for all instances where exact solutions could be obtained using ACE. 
It is obvious from the figure that IBIA outperforms the other two methods by a significant margin. Out of 494 instances, the error obtained with IBIA is less than $0.5$ in 463 testcases as opposed to 380 and 204 with FBP and WMB20 respectively. The maximum absolute error in $\log(PR)$ is about 4 with IBIA, while it is 42 with WMB20, and 15 with FBP.
We did not use IJGP for evaluation of PR.\footnote{We use the $Merlin$ tool to get results for IJGP. Merlin dumps the following output when the task is set to PR for IJGP method ``For PR inference use WMB and BTE algorithms".} 

Figures~\ref{fig:cactusMaxErrpost} and \ref{fig:cactusRMSEpost} show cactus plots for the maximum error and RMSE in  $MAR_e$ for the various methods, in instances where exact solutions were obtained using ACE. We observe that IBIA solves significantly more instances with lower errors. The max-error in $MAR_e$ is less than 0.05 in 345 instances with IBIA and in less than 190 instances with all other methods. Similarly, the RMSE in $MAR_e$ is less than 0.05 in 408 instances with IBIA and in less than 280 instances with all other methods. 
\begin{figure}[!htbp]
	\centering
    \begin{subfigure}{0.5\textwidth}
        \centering 
		\includegraphics[width=\textwidth]{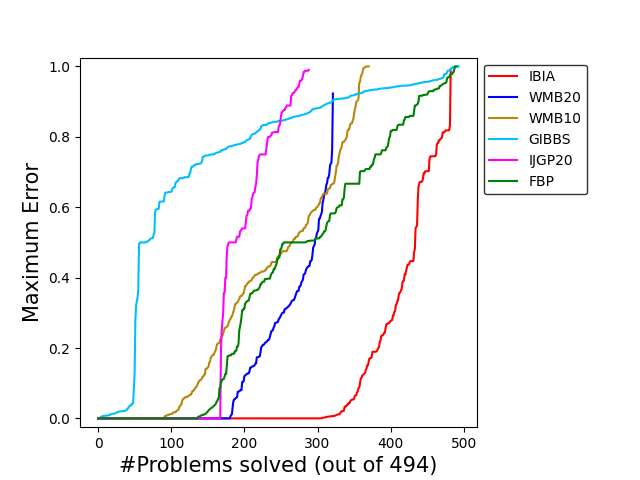}
        \caption{Maximum Error in posterior marginals, $MAR_e$}
        \label{fig:cactusMaxErrpost}
    \end{subfigure}%
    \begin{subfigure}{0.5\textwidth}
        \centering 
		\includegraphics[width=\textwidth]{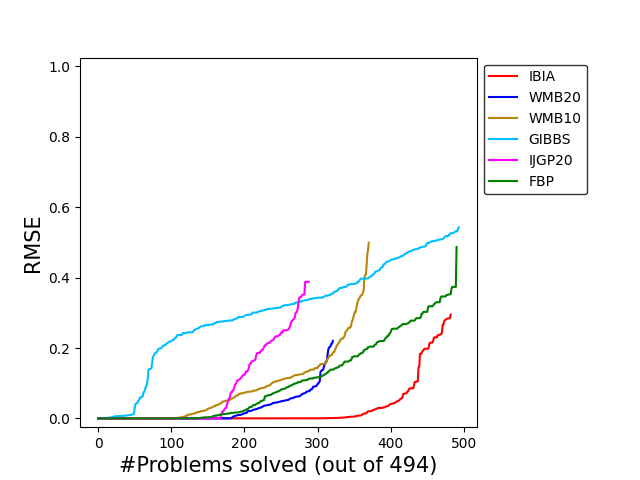}
        \caption{RMSE in posterior marginals, $MAR_e$}
        \label{fig:cactusRMSEpost}
    \end{subfigure}
    \begin{subfigure}{0.5\textwidth}
        \centering 
		\includegraphics[width=\textwidth]{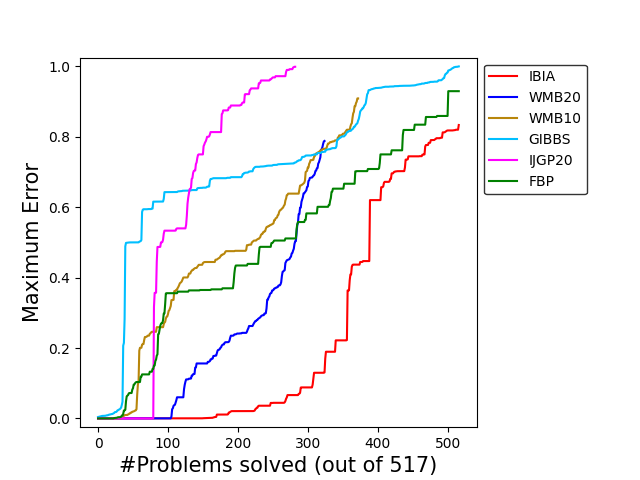}
        \caption{Maximum Error in prior marginals, $MAR_p$}
        \label{fig:cactusMaxErrprior}
    \end{subfigure}%
    \begin{subfigure}{0.5\textwidth}
        \centering 
		\includegraphics[width=\textwidth]{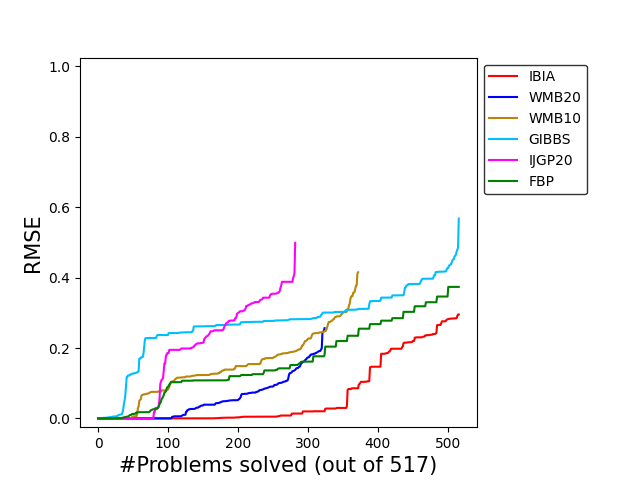}
        \caption{RMSE in prior marginals, $MAR_p$ }
        \label{fig:cactusRMSEprior}
    \end{subfigure}
    \caption{Cactus plots for maximum error and RMSE in $MAR_e,~MAR_p$ obtained after inference using various methods. A point (x,y) on the cactus plot indicates the number of instances that give metric is less than or equal to $y$.}
    \label{fig:cactus}
\end{figure}

Figures~\ref{fig:cactusMaxErrprior} and \ref{fig:cactusRMSEprior} show cactus plots for the maximum error and the RMSE in marginals $MAR_p$ estimated with IBIA and other methods. Once again, IBIA outperforms the other methods. The max-error in $MAR_p$ is less than 0.05 in 269 instances with IBIA and in less than 110 instances with others. Similarly, the RMSE in $MAR_p$ is less than 0.05 in 357 instances with IBIA and in less than 190 instances with others.
\begin{figure}[!htb]
	\centering
        \begin{subfigure}{0.5\textwidth}
            \centering 
            \includegraphics[width=\textwidth]{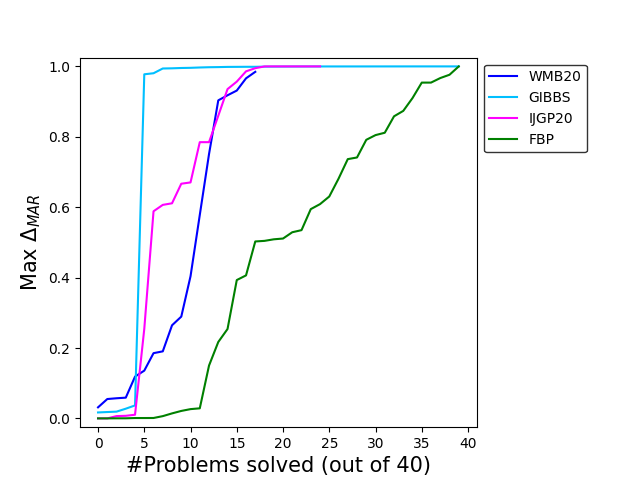}
            \caption{Maximum absolute difference in $MAR_e$}
            \label{fig:cactus_maxMARe}
        \end{subfigure}%
        \begin{subfigure}{0.5\textwidth}
            \centering 
            \includegraphics[width=\textwidth]{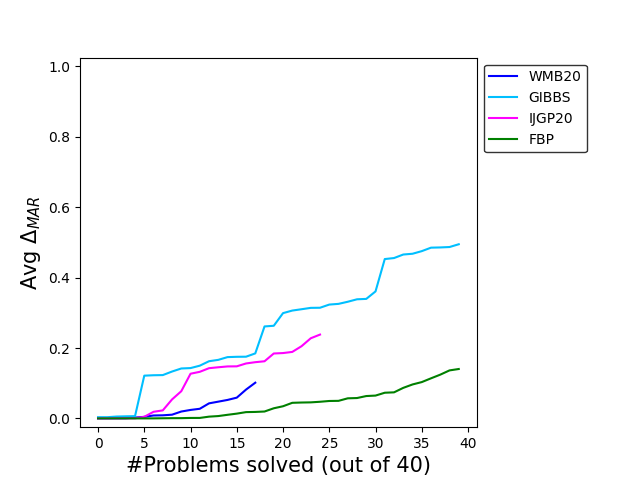}
            \caption{Average absolute difference in $MAR_e$}
            \label{fig:cactus_avgMARe}
        \end{subfigure}
        \begin{subfigure}{0.5\textwidth}
            \centering 
            \includegraphics[width=\textwidth]{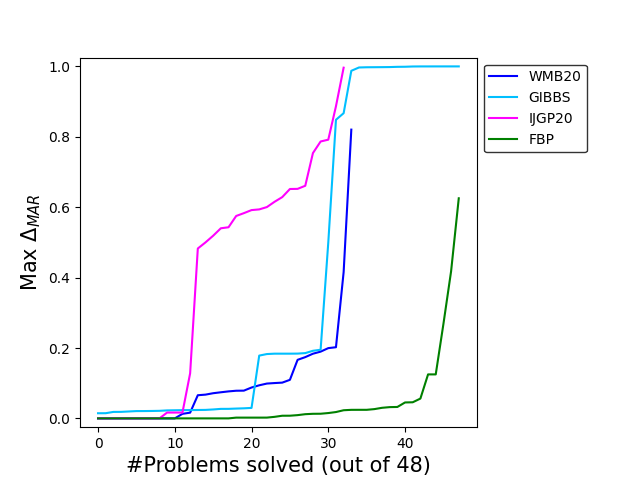}
            \caption{Maximum absolute difference in $MAR_p$}
            \label{fig:cactus_maxMARp}
        \end{subfigure}%
        \begin{subfigure}{0.5\textwidth}
            \centering 
            \includegraphics[width=\textwidth]{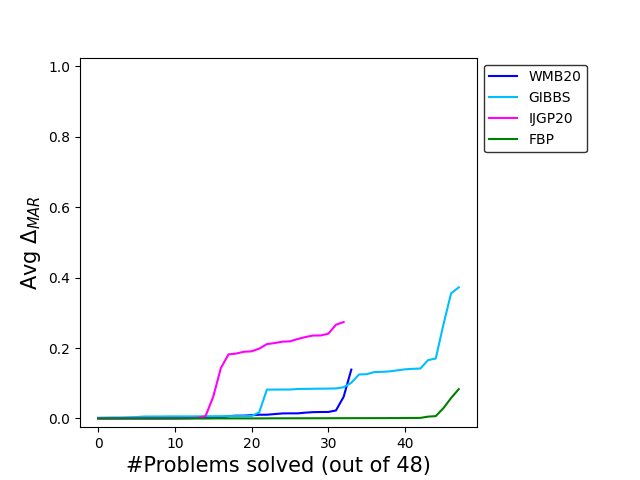}
            \caption{Average absolute difference in $MAR_p$}
            \label{fig:cactus_avgMARp}
        \end{subfigure}
        \caption{Cactus plots comparing average and maximum error in marginals computed using results obtained with IBIA as the baseline for benchmarks where ACE does not work. $Max~\Delta_{MAR}=Max_{v,s}~|MAR_{Alg}[v,s]-MAR_{IBIA}[v,s]|$,\\ $Avg~ \Delta_{MAR}=Avg_{v,s}~|MAR_{Alg}[v,s]-MAR_{IBIA}[v,s]|$ \\
        where, $v$ is a variable in the BN and $s\in Domain(v)$.}
        \label{fig:cactusMAR}
\end{figure}

For benchmarks where exact solutions could not be obtained using ACE, we compared the solutions of other approximate methods with the solution obtained using IBIA.
Figures~\ref{fig:cactus_maxMARe} and \ref{fig:cactus_avgMARe} show cactus plots for $MAR_e$ and
Figures~\ref{fig:cactus_maxMARp} and \ref{fig:cactus_avgMARp} have the results for $MAR_p$. 
In terms of the maximum difference in $MAR_e$, the methods give similar results for very few instances. FBP is the closest to IBIA, but the maximum difference is large in many cases. The average difference in $MAR_e$ is much lower than the maximum difference. For $MAR_p$, FBP gives similar results to IBIA in most cases and the average compares well.
For all instances, the average difference in $MAR_e$ and $MAR_p$ obtained with IBIA and FBP is less than $0.14$ and $0.08$ respectively. On the other hand,  the maximum difference is as high as 1.0 for $MAR_e$ and 0.6 for $MAR_p$. 

\subsubsection{Detailed results for each benchmark set}
We now present detailed comparisons of results obtained with different methods for each probability query.

To evaluate the accuracy of the estimated PR, we first considered the subset of benchmarks where exact probabilities could be found using ACE. 
Problems that cannot be solved within the one hour time limit or give estimates of PR which are greater than $1$ or give underflow errors are considered unsolved.
\begin{table}[!htp]\centering
    \caption{Average relative error in $\log(PR)$ ($\Delta_{PR}=|\log PR_{Alg}-\log PR_{ACE}|$), the average runtime (in seconds) and the number of instances (\#Inst) solved with different approximate inference methods for various benchmarks. For each benchmark, the total number of instances for which exact PR can be estimated using ACE are shown in parenthesis.}
    \label{tab:avgPR}
    \scriptsize
    \resizebox{\textwidth}{!}{%
        \setlength\tabcolsep{2pt}
    \begin{tabular}{|c|ccc|ccc|ccc|}\toprule
&\multicolumn{3}{c}{\textbf{BN\_UAI (90)}} &\multicolumn{3}{c}{\textbf{Pedigree (9)}} &\multicolumn{3}{c}{\textbf{Relational (395)}} \\\cmidrule{2-10}
        \textbf{Method} &\textbf{\#Inst} &\textbf{Avg $\Delta_{PR}$} &\textbf{Avg Runtime (s)} &\textbf{\#Inst} &\textbf{Avg $\Delta_{PR}$} &\textbf{Avg Runtime (s)} &\textbf{\#Inst} &\textbf{Avg $\Delta_{PR}$} &\textbf{Avg Runtime (s)} \\\midrule
FBP &90 &0.27 &33 &9 &2.25 &15 &392 &1.00 &51 \\
WMB20 &90 &2.55 &239 &5 &0.93 &836 &227 &4.40 &481 \\
WMB10 &90 &5.53 &8 &9 &5.22 &26 &272 &19.00 &216 \\
        \cellcolor[HTML]{A8A8A8}IBIA &\cellcolor[HTML]{A8A8A8}90 &\cellcolor[HTML]{A8A8A8}3$\times~10^{\mbox{-}4}$ &\cellcolor[HTML]{A8A8A8}2 &\cellcolor[HTML]{A8A8A8}9 &\cellcolor[HTML]{A8A8A8}0.12 &\cellcolor[HTML]{A8A8A8}1 &\cellcolor[HTML]{A8A8A8}384 &\cellcolor[HTML]{A8A8A8}0.08 &\cellcolor[HTML]{A8A8A8}59 \\
\bottomrule
    \end{tabular}%
    }
\end{table}

Table~\ref{tab:avgPR} contains the average absolute error in $\log(PR)$ ($\Delta_{PR}=|\log~PR_{Alg} - \log~PR_{ACE}$|) obtained by different approximate algorithms for each benchmark set. 
The total number of instances solved by ACE is indicated in brackets and those solved by different approximate algorithms are shown in column~$\#Inst$. 
 IBIA solves almost all instances except for a few relational benchmarks. The average error obtained with IBIA of the order of $10^{\mbox{-}4}, 10^{-1}, 10^{-2}$ for BN\_UAI, Pedigree and Relational benchmarks respectively.
 The errors are atleast an order of magnitude lower than that obtained with other methods. WMB gives upper bounds for PR. The average error with WMB20 is as large as 19 for the relational benchmarks. 
 It is also seen from the table that the average runtime for IBIA is atleast an order of magnitude lower than WMB20 and FBP in all benchmarks except Relational where it is comparable to FBP.

Table~\ref{tab:pr.hardInstances} shows the absolute error in $\log(PR)$ for some of the hard benchmarks where the exact solutions were obtained using Bucket Elimination with External Memory by \citeA{Gogate2011}. These instances could not be solved using ACE and have large induced widths ($w$). Entries are marked with `-' if no solution was obtained. 
The maximum variable domain cardinality ($d_{max}$) for the tabulated BN\_UAI instances is 36, and that for Pedigree instances ranges from 3-7; WMB20 runs out of memory in most cases. Therefore, we also show results for WMB10. In addition to FBP and WMB, we also compare with two other methods:  SampleSearch with Iterative Join Graph Propagation and w-cutset sampling (IJGP\_wc\_SS)~\cite{Gogate2011} and Edge Deletion Belief Propagation (EDBP)\cite{Choi2006}. The results for these two methods are directly taken from ~\citeA{Gogate2011}. It is seen that IBIA gives better estimates than FBP, EDBP, WMB10 and WMB20 in all these cases. Except for three testcases BN\_70, BN\_72 and BN\_75, our estimates are comparable to IJGP with sample search proposed in \citeA{Gogate2011}. 
\begin{table}[!htp]\centering
    \caption{Error in log(PR) estimated using different approximate inference techniques ($\Delta_{Alg}=|\log ~PR_{Alg}-\log~PR_{Exact}$|) for some BN\_UAI and pedigree instances exact solutions ($PR_{Exact}$) are reported in~\citeA{Gogate2011}.  The induced width ($w$) and the estimates of PR for SampleSearch with Iterative Join Graph Propagation and w-cutset sampling (IJGP\_wc\_SS) and Edge Deletion Belief Propagation (EDBP) were obtained from~\citeA{Gogate2011}.  Entries are marked with `-' for instances where no solution was obtained. $d_{max}$ is the maximum variable domain cardinality.
    }
    \label{tab:pr.hardInstances}
    \scriptsize
    \begin{tabular}{|cccccccc|}\toprule
        \textbf{Benchmark} & $\mathbf{(w,d_{max})}$ & $\boldsymbol{\Delta_{FBP}}$ & $\boldsymbol{\Delta_{WMB20}}$ &$\boldsymbol{\Delta_{WMB10}}$  & $\boldsymbol{\Delta_{IJGP\_wc\_SS}}$ & $\boldsymbol{\Delta_{EDBP}}$ & $\boldsymbol{\Delta_{IBIA}}$ \\\midrule
            BN\_69 &(39,36) &7.18 &- &-     &1.25 &3.34 &1.51 \\
            BN\_70 &(35,36) &6.87 &- &-     &2.22 &7.52 &4.44 \\
            BN\_71 &(53,36) &7.97 &- &43.90 &0.6 &3.7& 0.55$^{*}$ \\
            BN\_72 &(65,36) &6.32 &- &-     &0.05 &4.66 &3.69 \\
            BN\_73 &(67,36) &5.57 &- &- &2.05 &5 &1.46 \\
            BN\_74 &(35,36) &4.30 &- &11.40 &1.25 &2.81 &1.72$^{*}$ \\
            BN\_75 &(37,36) &7.79 &- &- &0.43 &5.28 &2.14 \\
            BN\_76 &(53,36) &5.74 &- &60.80 &1.4 &4.11 &1.75 \\
            pedigree25 &(38,5) &3.92 &- &4.68 &0 &0 &0 \\
            pedigree42 &(23,5) &1.21 &- &3.21 &0 &0.29 &0.01 \\
            pedigree31 &(45,5) &2.00 &- &15.40 &0.02 &0.17 &0.06 \\
            pedigree34 &(59,5) &1.39 &- &14.40 &0.19 &0.14 &0.04 \\
            pedigree13 &(51,3) &0.53 &- &12.50 &0.11 &1.92 &0.08 \\
            pedigree9 &(41,7)  &3.66 &4.15 &10.60 &0.07 &0.04 &0.06 \\
            pedigree19 &(23,5) &1.99 &- &14.90 &0.14 &0.45 &0.11 \\
            pedigree7 &(56,4)  &1.84 &- &11.80 &0.05 &0.71 &0.24 \\
            pedigree51 &(51,5) &3.39 &5.53 &14.30 &0.27 &1.34 &0.16 \\
            pedigree44 &(29,4) &4.02 &- &5.83 &0 &1.64 &0.04 \\ \midrule
            \rowcolor{gray!30}{\it \bf Mean $\mathbf{|\Delta_{Alg}|}$} && {\it 4.21}& {\it 4.84} &{\it 17.21} &{\it 0.51} & {\it 2.39} & {\it 0.83}  \\
        \bottomrule
    \end{tabular}
    \\
{\footnotesize *Testcases inferred using $mcs_{im}=14$}
\end{table}

\begin{table}[!htb]\centering
\caption{Comparison of $\log(PR)$ obtained using different approximate inference methods for testcases where no exact solution is known. Entries are marked with `-' for testcases where inferred PR is greater than $1$ and with `UF' where an underflow error occurs.}
\label{tab:PRErrAce}
\scriptsize
	\begin{tabular}{|cccc|}\toprule
		\textbf{Benchmark} &\textbf{FBP}  &\textbf{WMB20} &\textbf{IBIA}  \\\midrule
        BN\_12* &-3.61 &-3.31 &-3.61 \\
        BN\_126 &-55.68 &-54.26 &-55.78 \\
        BN\_127 &-57.5 &-55.94 &-57.68 \\
        BN\_128 &-47.29 &-47.27 &-47.32 \\
        BN\_129 &-61.06 &-56.57 &-61.9 \\
        BN\_13* &-2.32 &-1.98 &-2.32 \\
        BN\_130 &-57.41 &-56.59 &-57.46 \\
        BN\_131 &-53.98 &-53.27 &-53.69 \\
        BN\_132 &-64.1 &-58.61 &-64.44 \\
        BN\_133 &-53.64 &-51.82 &-53.32 \\
        BN\_134 &-56.2 &-54.62 &-56.86 \\
        BN\_15* &-5.7 &-5.33 &-5.7 \\
        BN\_16 &-0.08 &-0.01 &-0.08 \\
        BN\_17 &-0.08 &-0.05 &-0.08 \\
        BN\_18 &-0.09 &- &-0.09 \\
        BN\_19 &-0.09 &0 &-0.09 \\
        BN\_20* &-410.5 &- &UF \\
        BN\_21* &-410.5 &- &UF \\
        BN\_26* &-974.12 &- &-973.99 \\
        BN\_27* &-974.12 &- &-973.99 \\
        BN\_77 &-83.65  &- & -81.43$^{++}$ \\
        pedigree40* &-92.81 &- &-87.86$^{+}$ \\
        pedigree41* &-77.78 &- &-75.96 \\
        pedigree50* &-23.70 &- &-22.11 \\
        \bottomrule
        \end{tabular}\\
        {
            \footnotesize $^{+}$Testcases inferred using $mcs_{im}=14$\\
            \footnotesize $^{++}$Testcases inferred using $mcs_{im}=12$\\
        \footnotesize $^{*}$PR for these benchmarks is reported on website~\cite{IhlerURL}. Since we were unable to reproduce these using ACE, we do not use values in the website for our evaluation.}
\end{table}

We compare the $\log(PR)$ obtained by the approximate methods for instances where exact solution could not be obtained in Table~\ref{tab:PRErrAce}. 
Except for BN\_77 and three pedigree testcases, the PR estimated using IBIA and FBP match well.
There are significant differences with WMB20 for a few benchmarks, but this is expected since it computes the upper bound.
\begin{table}[!htb]\centering
    \caption{Average maximum error, RMSE, mean and max KL distance, required runtime (in seconds) for obtaining posterior marginals ($MAR_e$) with different approximate inference methods for various benchmarks. For each benchmark, the total number of instances for which exact marginals can be estimated using ACE are shown in parenthesis and the count of those solved by each method is shown in column $\#Inst$. For Relational and Pedigree benchmarks, the average errors obtained with IBIA over the set of instances solved by WMB20 are indicated in brackets.}\label{tab:avgMARe}
\scriptsize
\resizebox{\textwidth}{!}{%
\begin{tabular}{|c|ccccccc|}\toprule
    \textbf{Benchmark}   &\textbf{Method} &\textbf{\#Inst.} &\textbf{Avg. MaxErr} &\textbf{Avg. RMSE} &\textbf{Avg. $KL_{mean}$} &\textbf{Avg. $KL_{max}$} &\textbf{Avg. Runtime (s)} \\\midrule
    \multirow{6}{*}{\textbf{\shortstack{BN\_UAI \\(90)}}} &FBP &90 &0.207 &0.039 &0.035 &3.08 &33 \\
    &WMB20 &90 &0.202 &0.046 &0.004 &0.141 &239 \\
    &WMB10 &90 &0.288 &0.078 &0.008 &0.209 &8 \\
    &IJGP20 &81 &0.326 &0.079 &0.245 &5.05 &148 \\
    &GIBBS &90 &0.42 &0.146 &0.048 &3.63 &6 \\
    &\cellcolor{gray!30}IBIA &\cellcolor{gray!30}90 &\cellcolor{gray!30}0.006 &\cellcolor{gray!30}0.001 &\cellcolor{gray!30}5$\times 10^{\mbox{-}6}$ &\cellcolor{gray!30}0.003 &\cellcolor{gray!30}4 \\ \midrule
    \multirow{6}{*}{\textbf{\shortstack{Pedigree \\(9)}}} &FBP &9 &0.52 &0.189 &0.623 &41.2 &15 \\
    &WMB20 &5 &0.174 &0.018 &5$\times 10^{\mbox{-}4}$ &0.099 &836 \\
    &WMB10 &9 &0.421 &0.064 &0.004 &0.303 &26 \\
    &IJGP20 &5 &0.49 &0.135 &0.267 &7.57 &411 \\
    &GIBBS &9 &0.998 &0.436 &1.88 &15.9 &7 \\
    &\cellcolor{gray!30}IBIA &\cellcolor{gray!30}9(5) &\cellcolor{gray!30}0.237(0.208) &\cellcolor{gray!30}0.028(0.021) &\cellcolor{gray!30}0.001(3$\times 10^{\mbox{-}4}$) &\cellcolor{gray!30}0.151(0.120) &\cellcolor{gray!30}18(21) \\ \midrule
    \multirow{6}{*}{\textbf{\shortstack{Relational \\(395)}}} &FBP &392 &0.449 &0.123 &0.295 &11.101 &51 \\
    &WMB20 &227 &0.112 &0.019 &0.001 &0.072 &481 \\
    &WMB10 &272 &0.33 &0.088 &0.017 &0.632 &216 \\
    &IJGP20 &203 &0.287 &0.082 &0.306 &4.319 &376 \\
    &GIBBS &395 &0.811 &0.337 &0.098 &1.176 &243 \\
    &\cellcolor{gray!30}IBIA &\cellcolor{gray!30}384(227) &\cellcolor{gray!30}0.146(0.055) &\cellcolor{gray!30}0.036(0.009) &\cellcolor{gray!30}0.01(4$\times 10^{\mbox{-}4}$) &\cellcolor{gray!30}0.781(0.04) &\cellcolor{gray!30}147(76) \\ 
\bottomrule
\end{tabular}%
    }
\end{table}

The next inference task we consider is the computation of posterior singleton marginals. 
 Table~\ref{tab:avgMARe} shows the average maximum error, RMSE, mean and max KL distance, and runtime required for inference of posterior marginals ($MAR_e$) for each of the benchmark sets. For each benchmark set, the total number of instances for which exact marginals can be estimated using ACE are shown in parenthesis and the count of those solved by each method is shown in column $\#Inst$. 
    For BN\_UAI benchmarks, we observe that the average errors obtained with IBIA for all metrics are atleast an order of magnitude lower than the other approximate methods. 
    For Pedigree and Relational benchmarks, IBIA and WMB20 have significantly lower average errors than FBP, GIBBS, IJGP20 and WMB10. The average errors with WMB20 are lower than IBIA for Pedigree and Relational benchmarks. However, IBIA solves more testcases than WMB20 for both benchmark sets. We show the average errors obtained with IBIA over the limited set of instances solved by WMB20 in parenthesis in Table~\ref{tab:avgMARe}. For Pedigree, the average error obtained with IBIA over these instances is comparable to WMB20, even though it is at a disadvantage in terms of clique sizes. For Relational benchmarks, the average errors obtained with IBIA over these 227 instances is lower than WMB20.
    IBIA is expected to give lower errors than FBP, due to larger cluster sizes. However, it also gives significantly lower errors than WMB20 and IJGP20, which are run with comparable or larger cluster sizes. We note that for FBP, WMB20, WMB10 and IJGP20, although average max-errors are large, the mean RMSE are small. This indicates that a large percentage of variables have small errors with these approximate techniques.

Though runtimes for different methods are not directly comparable since they are implemented in different programming languages, it gives a decent overall idea.
    We note that even though IBIA is implemented in Python, the runtimes are quite competitive when compared to other tools. The average runtime for IBIA is $4, ~18,~147$ seconds for BN\_UAI, Pedigree and Relational benchmarks respectively. 
    IBIA has the lowest average runtime for BN\_UAI benchmarks, which is significantly better than FBP, WMB20 and IJGP20. FBP has a marginally lower runtime for Pedigree benchmarks and a significantly lower runtime for Relational benchmarks. But, the error is much larger.

Figure~\ref{fig:scorePlotPosterior} plots the variation of $SumScore$ obtained using various approximate inference methods as the timeout constraint is varied for different benchmark sets. Since WMB20 timed out for the remaining Pedigree cases, we also ran it with an $ibound$ of 10. 
Across benchmarks, IBIA has a comparable or better $SumScore$ than other methods. For Pedigree, IBIA and WMB10 have a score that is close to 9, which is the perfect score. Also for Pedigree and BN\_UAI benchmarks, WMB20 gets a higher final score than FBP even though it solves fewer benchmarks, indicating it has a lower average error. This is consistent with the average $KL_{mean}$ reported in Table~\ref{tab:avgMARe}. Gibbs sampling performs very poorly with Pedigree benchmarks. Generally, we have found that if the RMSE is less than 0.1, the score is close to the perfect score. 
\begin{figure}[!h]
	\centering
	\begin{subfigure}{0.49\textwidth}
		\centering
		\includegraphics[width=0.95\textwidth]{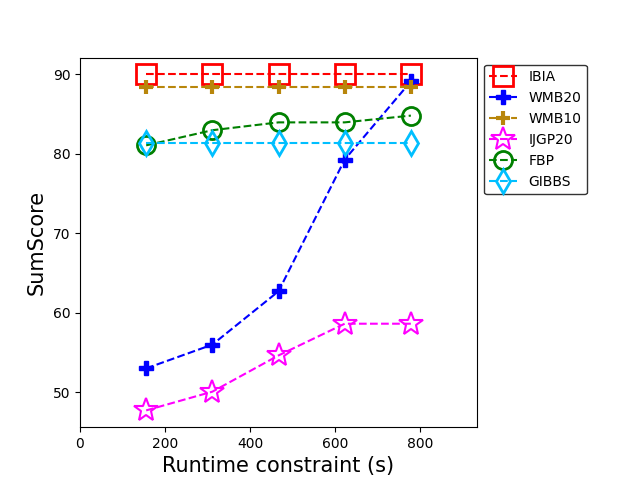}
		\caption{BN\_UAI \\{\scriptsize(Instances - \#ACE/FBP/GIBBS/IBIA:90,\\ \#WMB20/WMB10:90 \#IJGP20:81)}}
	\end{subfigure} \hfill
	\begin{subfigure}{0.49\textwidth}
		\centering
		\includegraphics[width=0.95\textwidth]{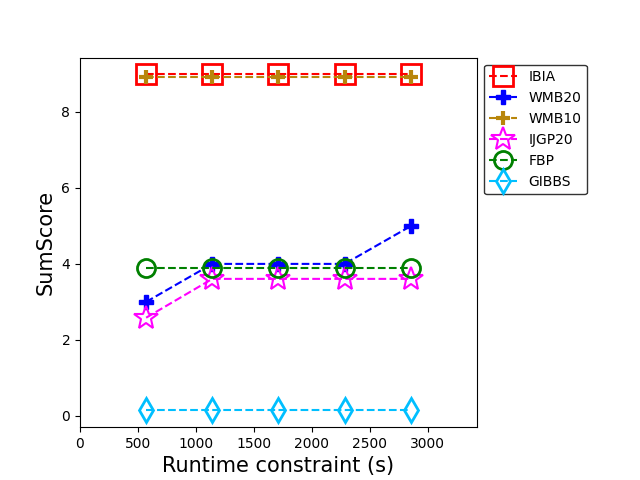}
		\caption{Pedigree\\{\scriptsize(Instances - \#ACE/FBP/WMB10/GIBBS/IBIA:9,\\ \#WMB20/IJGP20:5)}}
	\end{subfigure}
	\begin{subfigure}{0.5\textwidth}
		\centering
		\includegraphics[width=0.95\textwidth]{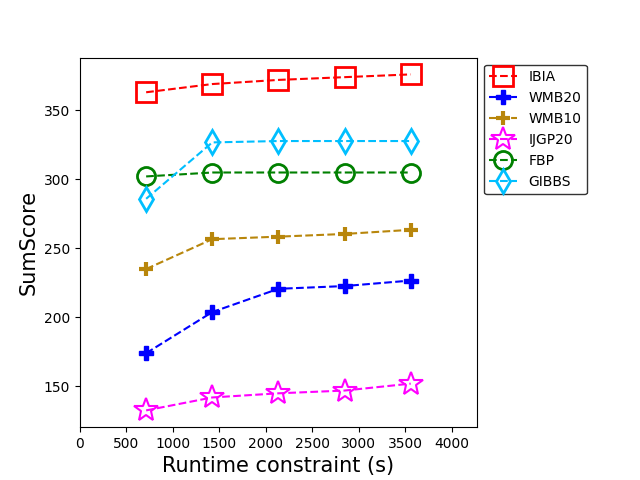}
		\caption{Relational\\{\scriptsize(Instances~-~\#ACE/GIBBS:395,~\#FBP:392,\\ \#WMB20:227,~\#WMB10:272,~\#IJGP20:203,~\#IBIA:384)}}
	\end{subfigure}
    \caption{Variation of $SumScore$ for posterior marginals obtained using various approximate inference techniques with different timeout constraints. The total number of instances that run with ACE and other approximate inference techniques are indicated for each benchmark set.}
	\label{fig:scorePlotPosterior}
\end{figure}

For the relational benchmarks, IBIA has the highest score, once again close to the number of benchmarks solved. WMB20 also gets a score close to the number of benchmarks solved by it, indicating low error on an average. Gibbs sampling gets a much better score for relational benchmarks. However, the score is much lower than number of cases solved by it.
For all benchmarks, even though FBP runs successfully for a comparable number of instances as IBIA, the scores obtained are smaller, indicating a larger error in estimation. 
IJGP20 has a significantly lower $SumScore$ as compared to other methods for BN\_UAI and Relational benchmarks, partly due to the smaller number of cases for which a solution is obtained.\footnote{IJGP20 returns inconsistent evidence or underflow errors for many testcases.}

We now discuss the results obtained for prior marginals. 
We have also reported some additional results for circuit benchmarks in a preliminary version of this paper in \citeA{Bathla2021}. We now discuss the results for benchmarks listed in section \ref{sec:dataset}. As shown in Table~\ref{tab:datasetInfo}, ACE solves fewer prior testcases when compared to posteriors. This is because network sizes are larger, with fewer simplifications possible.

 Table~\ref{tab:avgMARp} shows the average maximum error, RMSE, mean and max KL distance, and the required runtime for obtaining prior marginals ($MAR_p$) with different approximate inference methods for various benchmarks. For each benchmark, the total number of instances for which exact marginals can be estimated using ACE are shown in parenthesis and the count of those solved by each method is shown in column $\#Inst$. 
 For BN\_UAI, all average error metrics obtained using IBIA are atleast an order of magnitude lower than with all other methods. For Grid-BN, IBIA outperforms other methods by several orders of magnitude. 
 For Relational benchmarks, IBIA performs much better than WMB20 for the same set of 194 benchmarks. It also performs better than other methods in terms of the average max-error, RMSE and $KL_{mean}$. For a comparable number of cases, the average $KL_{max}$ obtained with IBIA is better than FBP, but worse than Gibbs sampling.
    For Bnlearn benchmarks, IBIA gives lower errors than FBP, WMB10, IJGP10, and GIBBS. Methods WMB20 and IJGP20 reduce to exact belief propagation since the tree-width for these benchmarks is less than $20$. Therefore, the errors obtained are close to 0 when the maximum number of variables in a clique is restricted to 20. In contrast, the clique size constraint in IBIA is in terms of the domain cardinality and the maximum variable domain cardinality in these benchmarks ranges from 3-100. Therefore, IBIA performs approximate inference in 10/26 benchmarks when $mcs_p$ is set to $20$. For these instances, the maximum number of variables in any clique in the sequence of CTFs is around 10 on an average. Therefore, comparison with WMB10 and IJGP10 is more appropriate in this case. It is seen that for all metrics, errors are an order of magnitude lower with IBIA than with WMB10 and IJGP10. 
\begin{table}[!htp]\centering
    \caption{Average maximum error, RMSE, mean and max KL distance, required runtime (in seconds) for obtaining prior marginals ($MAR_p$) with different approximate inference methods for various benchmarks. For each benchmark, the total number of instances for which exact marginals can be estimated using ACE are shown in parenthesis and count of those solved by each method is shown in column $\#Inst$. For Relational benchmarks, the average errors obtained with IBIA over the set of instances solved by WMB20 are indicated in brackets.}\label{tab:avgMARp}
    \scriptsize
    \resizebox{\textwidth}{!}{%
    \begin{tabular}{|c|ccccccc|}\toprule
        \textbf{Benchmark}   &\textbf{Method} &\textbf{\#Inst.} &\textbf{Avg. MaxErr} &\textbf{Avg. RMSE} &\textbf{Avg. $KL_{mean}$} &\textbf{Avg. $KL_{max}$} &\textbf{Avg. Runtime (s)} \\\midrule
        \multirow{6}{*}{\textbf{\shortstack{BN\_UAI \\(76)}}} &FBP &76 &0.169 &0.036 &0.003 &0.114 &151 \\
        &WMB20 &76 &0.254 &0.073 &0.007 &0.188 &449 \\
        &WMB10 &76 &0.312 &0.104 &0.012 &0.251 &3 \\
        &IJGP20 &65 &0.527 &0.165 &0.767 &8.21 &235 \\
        &GIBBS &76 &0.454 &0.142 &0.036 &3.66 &9 \\
        &\cellcolor{gray!30}IBIA &\cellcolor{gray!30}76 &\cellcolor{gray!30}0.015 &\cellcolor{gray!30}0.001 &\cellcolor{gray!30}5$\times 10^{\mbox{-}6}$ &\cellcolor{gray!30}0.007 &\cellcolor{gray!30}4 \\ \midrule
                  \multirow{6}{*}{\textbf{\shortstack{Grid-BN \\ (29)}}} &FBP &29 &0.474 &0.109 &0.01 &0.4 &0.04 \\
        &WMB20 &29 &0.427 &0.095 &0.007 &0.288 &373 \\
        &WMB10 &29 &0.52 &0.146 &0.015 &0.355 &4 \\
        &IJGP20 &24 &0.746 &0.235 &0.893 &11.8 &256 \\
        &GIBBS &29 &0.919 &0.388 &1.4 &13.3 &6 \\
        &\cellcolor{gray!30}IBIA &\cellcolor{gray!30}29 &\cellcolor{gray!30}9$\times 10^{\mbox{-}4}$ &\cellcolor{gray!30}3$\times 10^{\mbox{-}5}$ &\cellcolor{gray!30}2$\times 10^{\mbox{-}8}$ &\cellcolor{gray!30}4$\times 10^{\mbox{-}4}$ &\cellcolor{gray!30}2 \\ \midrule
 
          \multirow{6}{*}{\textbf{\shortstack{Relational \\(386)}}} &FBP &386 &0.592 &0.205 &0.208 &7.054 &15 \\
        &WMB20 &194 &0.199 &0.05 &0.003 &0.135 &497 \\
        &WMB10 &242 &0.529 &0.171 &0.035 &1.135 &222 \\
        &IJGP20 &168 &0.648 &0.222 &0.958 &9.619 &315 \\
        &GIBBS &386 &0.758 &0.297 &0.058 &0.873 &241 \\
        &\cellcolor{gray!30}IBIA &\cellcolor{gray!30}386(194) &\cellcolor{gray!30}0.314(0.034) &\cellcolor{gray!30}0.087(0.007) &\cellcolor{gray!30}0.027(1$\times 10^{\mbox{-}4}$) &\cellcolor{gray!30}2.013(0.019) &\cellcolor{gray!30}127(9) \\ \midrule
              \multirow{7}{*}{\textbf{\shortstack{Bnlearn \\ (26)}}} &FBP &26 &0.068 &0.01 &3$\times 10^{\mbox{-}4}$ &0.037 &0.1 \\
        &WMB20 &26 &4$\times 10^{\mbox{-}7}$ &2$\times 10^{\mbox{-}7}$ &8$\times 10^{\mbox{-}9}$ &7$\times 10^{\mbox{-}7}$ &9 \\
        &WMB10 &26 &0.028 &0.005 &3$\times 10^{\mbox{-}4}$ &0.016 &22 \\
        &IJGP20 &26 &4$\times 10^{\mbox{-}7}$ &2$\times 10^{\mbox{-}7}$ &7$\times 10^{\mbox{-}9}$ &7$\times 10^{\mbox{-}7}$ &7 \\
        &IJGP10 &21 &0.794 &0.284 &1.2 &12.3 &171 \\
        &GIBBS &26 &0.511 &0.171 &0.257 &6.61 &2 \\
        &\cellcolor{gray!30}IBIA &\cellcolor{gray!30}26 &\cellcolor{gray!30}0.003 &\cellcolor{gray!30}3$\times 10^{\mbox{-}4}$ &\cellcolor{gray!30}2$\times 10^{\mbox{-}6}$ &\cellcolor{gray!30}0.001 &\cellcolor{gray!30}0.3 \\
        \bottomrule
    \end{tabular}%
    }
\end{table}

    Similar to posteriors, we note that runtimes with IBIA are quite competitive when compared to other tools. The average required runtimes for IBIA are $4, ~2,~127$ and $0.3$ seconds for BN\_UAI, Grid-BN, Relational and Bnlearn benchmarks respectively. In all cases, the runtimes with IBIA are significantly lower than IJGP20 and WMB20. It is comparable to WMB10 and GIBBS for BN\_UAI and Grid-BN benchmarks. Runtimes with IBIA are significantly larger than  FBP for Grid-BN and Relational benchmarks but significantly lower for BN\_UAI benchmarks.

Figure~\ref{fig:scorePlotPrior} plots the variation of the SumScore for priors obtained using various approximate inference methods as a function of runtime.
For the BN\_UAI, Grid$-$BN and Bnlearn, the final score with IBIA, FBP and WMB20 is similar, indicating that the larger max-errors in the other methods occur only for a small number of states. For these benchmarks, WMB10 also achieves a similar score. For Bnlearn, all methods perform well, except for Gibbs sampling which performs very poorly. As discussed earlier, for Bnlearn benchmarks, exact inference is performed with IJGP20 and WMB20.

\begin{figure}[htb]
	\begin{subfigure}[t]{0.49\textwidth}
		\centering		
		\includegraphics[width=\textwidth]{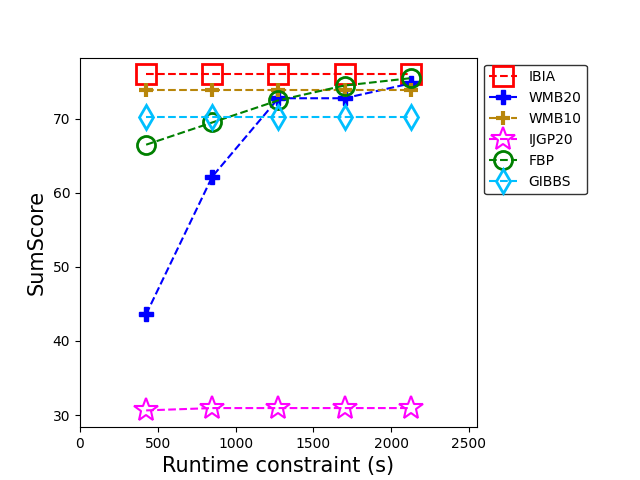}
		\caption{BN\_UAI\\{\scriptsize(Instances - \#ACE/FBP/GIBBS/IBIA:~76, \\\#WMB20/WMB10:~76, IJGP20:~65)}}
	\end{subfigure}
	\begin{subfigure}[t]{0.49\textwidth}
		\centering
		\includegraphics[width=\textwidth]{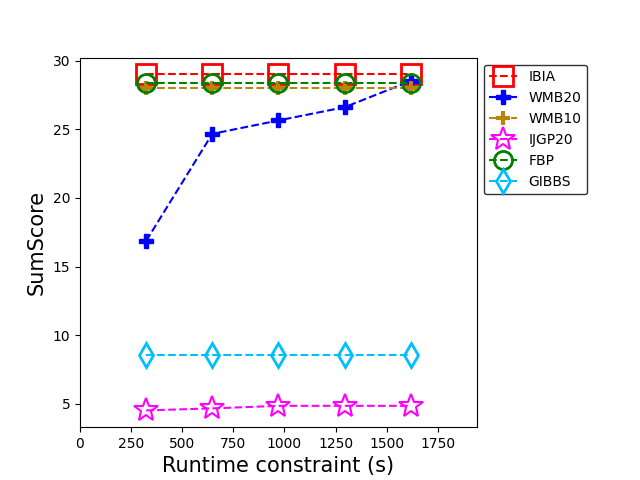}
		\caption{Grid-BN\\{\scriptsize(Instances - \#ACE/FBP/GIBBS/IBIA:~29, \\WMB20/WMB10:~29, IJGP20:~24)}}
	\end{subfigure}
	\begin{subfigure}[t]{0.49\textwidth}
		\centering
		\includegraphics[width=\textwidth]{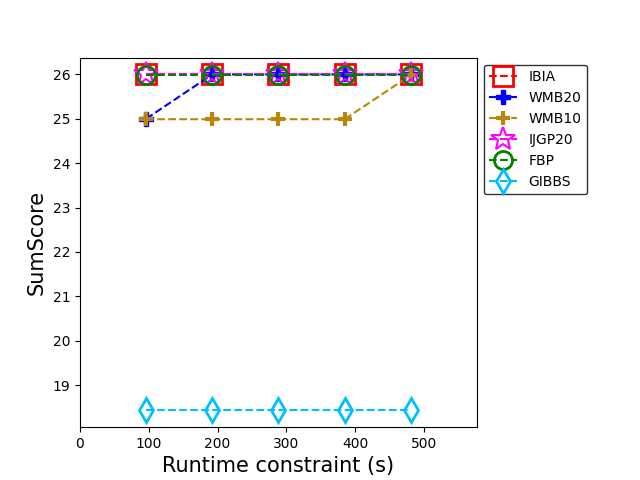}
		\caption{Bnlearn\\{\scriptsize(Instances - \#ACE/FBP/GIBBS/IBIA:~26\\ \#WMB20/WMB10/IJGP20:~26  )}}
	\end{subfigure}
	\begin{subfigure}[t]{0.49\textwidth}
		\centering
		\includegraphics[width=\textwidth]{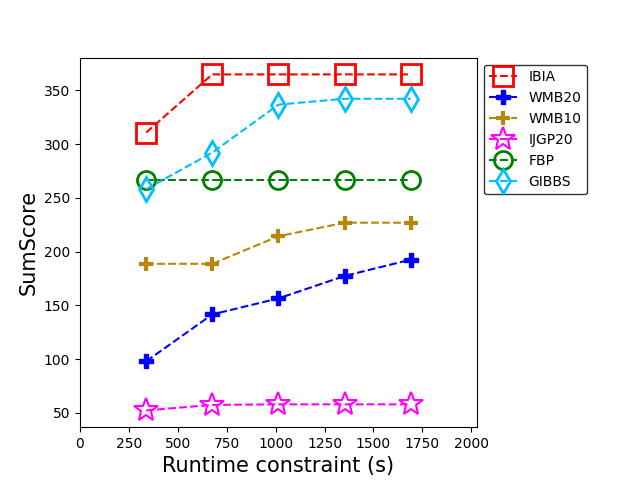}
		\caption{Relational\\{\scriptsize(Instances - \#ACE/FBP/GIBBS/IBIA:~386, \\WMB20:~194, WMB10:~242, IJGP20:~168)}}
	\end{subfigure}
	\caption{Variation of the $SumScore$ for prior marginals obtained using various approximate inference techniques with different timeout constraints. The total number of instances that run with ACE and other approximate inference techniques are indicated for each benchmark set.}
	\label{fig:scorePlotPrior}
\end{figure}
IBIA outperforms other deterministic approximate algorithms by a significant margin in Relational benchmarks. Even though FBP solves more cases than IBIA, the score achieved is small, indicating larger estimation errors. Once again for WMB20, the final score is close to the number of cases solved, indicating low average error. WMB10 solves a larger number of cases, but the average error increases as expected. WMB10 has a final score of 226 out of a possible maximum of 242, whereas WMB20 has a score  193 out of a possible maximum of 194. Gibbs sampling performs well in this set of benchmarks, indicating it works better when there is more determinism. For these benchmarks, IJGP20 does not perform very well both in terms of error and the number of cases solved. 

\section{Summary, Discussion of results and Conclusions}
We propose a technique for approximate  Bayesian inference that builds a sequence of linked CTFs with clique sizes bounded by a user specified parameter $mcs_p$. Each CTF in the sequence is {\it built} incrementally until the clique sizes reach $mcs_p$, the clique beliefs are {\it inferred} using BP, and finally it is {\it approximated} to reduce clique sizes to $mcs_{im}$. Adjacent CTFs in the sequence are linked via cliques that have consistent beliefs. The constructed sequence is used for approximate inference of the partition function and prior and posterior marginals. 
We show that our algorithm for incremental construction of CTs always results  in a valid CT and our approximation algorithm preserves the joint beliefs of variables within a clique. The trade-off between runtime and accuracy can be achieved easily in terms of a two user-specified parameters, $mcs_p$ and $mcs_{im}$.
Our method gives significantly smaller errors with competitive runtimes when compared with other approximate methods that have a similar trade-off. 
Also, unlike other approaches, our method is non-iterative in the sense that there are no iterations to get optimum regions nor do we use iterative BP for inference.


Except for a few cases with strong local structure (like the Grid benchmarks), the maximum clique size obtained using our incremental CT construction algorithm is within $\pm 2$ of what is obtained using complete triangulation of the corresponding BN subgraph using variable elimination with the min-fill heuristic. This is generally true upto a clique size of 25. As the clique sizes increase beyond this, the differences increase. However, the exponential time and space complexity of inference begins to dominate around these clique sizes, so it is not clear that larger clique sizes are useful. It is also possible to use a combination of incremental and full triangulation if larger clique sizes are desired.

IBIA gives much better estimates of $PR$ than all the other approximate methods.
The error in the estimates is atleast an order of magnitude lower than the other methods.
For $MAR_e$ and $MAR_p$, the max-error and RMSE  as well as the average and maximum KL distance are significantly lower than all the other approximate methods, even though IJGP and WMB have comparable or better cluster sizes. The $SumScore$ is comparable to or better than WMB. This is a reflection of the quality of our approximation algorithm. 
During ``forward propagation'' our approximation algorithm maintains within-clique belief consistency, which also gives good estimates of posterior beliefs once all the evidence variables are incorporated. The belief-update via back-propagation is heuristic and has some limitations. It is sequential and has to be done link by link for each CTF which needs to be updated. Each link update also has a round of message passing. If a belief is erroneously updated from a non-zero value to zero in an intermediate update step, it cannot recover after subsequent updates. This could give rise to errors in the normalization constant if the clique containing that particular belief is updated via a link. This happens occasionally and can be resolved either by not using this link for update or setting it to a small non-zero value.    
Also as mentioned, after the update is completed, beliefs of variables present in multiple CTFs need not be consistent. However, if the belief is obtained from the first CTF in which it is added, we get good accuracies. In spite of these limitations, we get significantly better estimates of posteriors  than other approximate algorithms used for comparison in this paper.

In our method, the trade-off between accuracy and runtime is controlled easily by two user-defined parameters: $mcs_p$ and $mcs_{im}$. 
While it is clear that increasing $mcs_p$  typically improves accuracy at the cost of runtime, the behavior with respect to $mcs_{im}$ is less definitive. Generally, reducing $mcs_{im}$ increases errors due to more aggressive approximations. However, in some cases, the number of CTFs reduces, resulting in a slightly lower error. The behaviour of the error with the number of CTFs in the SLCTF is also network dependent. Even if there are a large number of CTFs, the error can be very low or even zero, if the network structure allows a sufficient number of exact marginalizations.  For example, the max-error is only 0.05  for the diabetes benchmark which has 23 CTFs (refer Table~\ref{tab:mcsCmpMAR}). 
Another advantage that IBIA shares with IJGP and WMB is that exact inference is automatically performed when the clique sizes are less than the specified maximum, which is once again not possible in many region-graph based methods. This property is especially effective when a large number of evidence variables are present.

The run times for various algorithms cannot be compared directly, since they are written in different languages. Moreover, our algorithm is developed in Python and all the other algorithms are compiled programs in Java or C++. In spite of this, the run times for our algorithm are very competitive. We get much better run times than IJGP and WMB when $ibound$ was set as 20 to allow for a reasonably fair comparison of error. Our runtime is comparable or much better than WMB with $ibound$ set to 10. For $ibound$ of 20, our algorithm gave better accuracies even though it had lower clique sizes in networks that contain variables with large domain cardinality or many parents. Interestingly, in these networks (like BN\_UAI 105-125), IBIA outperforms FBP in terms of runtime, even though we have clique sizes of 20 and our code is written in Python.  In many cases, it is also comparable or better than the run-time of ACE. As seen from Table~\ref{tab:expLibDAI} in Appendix B, the other approximate algorithms take longer.
Amongst the deterministic approximate inference algorithms, FBP solves the largest number of problems, closely followed by IBIA. However, in all cases, we get better scores and lower errors than FBP, which is as expected. 

As discussed in Section~\ref{sec:vallim}, IBIA is not guaranteed to give a solution for a given $mcs_p$ and $mcs_{im}$ in the presence of evidence variables.  Out of 500+ testcases attempted in this work, with $mcs_p$ set to $20$, there were only 5 testcases where we could not get a solution with $mcs_{im}$ set to 15. For these cases, the variables that need to be added in the subsequent CTFs have large domain sizes. For these cases, we were able to obtain a solution by lowering the $mcs_{im}$, but this is not guaranteed in general. 

IBIA times out while building the CTF in some very large relational BNs (Friends \& Smokers). These networks contain numerous evidence states that force equivalence of parent variables. So far, we have not included such simplifications in our method. However, in large networks without determinism, this could be a constraint. A mix of full and incremental recompilation could possible work better in these cases.

Finally, our framework can be very useful when portions of the network are changed as well as for BN learning. Since our CT construction method is incremental, portions of the network that have been modified can be deleted and the CT reconstructed without recompiling the entire network. If new evidence variables need to be incorporated, the corresponding CTFs can be modified. If a new evidence variable and query are in the same CTF, the simplification and inference can be done on that CTF alone. In the context of learning, a problem is to learn networks that have a bound on the clique size. When the whole network is considered, this places severe restrictions on the number of edges that can be included. Our method can be used to relax some of these constraints.

\appendix

\section{Proofs}\label{app:proofs}
The propositions 1 to 4 are based on our algorithm for incremental addition of new variables to an existing CTF described in Section~\ref{sec:incrMod}. We denote the clique between the new variable $v$ and its parents $Pa_v$ by $C_v=\{v, Pa_v\}$. The minimal subgraph of the existing CTF that connects cliques containing parents is denoted as $SG_{min}$.
    As discussed in section \ref{sec:incrMod}, there are three cases for addition namely, Case 1: $SG_{min}$ has a single node, Case 2: $SG_{min}$ is a set of disconnected cliques and Case 3: $SG_{min}$ is fully or partially connected. 
\\ \\
\noindent\textbf{Proposition~\ref{pr:pr1}}
        The modified CTF obtained using $ModifyCTF$ (Algorithm \ref{alg:modifyCTF}) contains (possibly disjoint) trees i.e., no loops are introduced by the algorithm.
  \begin{proof}
      In all cases, a new clique $C_v$ containing the variable to be added and its parents is formed. In Case 1, $C_v$ is connected to a single clique in the existing CTF. In Case 2, $C_v$ connects disjoint trees. Therefore, in both cases, no loops are introduced and the resulting modified CTF consists of one or more disjoint trees.
    
    In Case 3, a subset of new variables that impact overlapping subgraphs are added together. The minimal subgraph $SG_{min}$ connecting all parents is removed from the CTF. For each new variable, a new clique $C_v$ containing the variable and its parents is formed. The presence of moralizing edges between parents of each variable ensures that the elimination graph $G_E$ is a connected graph. A single clique tree $ST'$ is obtained after triangulation of this graph. Each retained clique is re-connected to a single clique in $ST'$.  Due to the moralizing edges, $ST'$ also  contains at least one clique containing all parents of a variable, to which $C_v$ is connected.  Hence, the resulting structure $ST'$ continues to be a tree. $ST'$ replaces $SG_{min}$ and is connected to the CTF via the same set of cliques as $SG_{min}$. 
Therefore, no loops are introduced and the modified CTF continues to have one or more disjoint trees.    	 
\end{proof}
\noindent\textbf{Proposition~\ref{pr:pr2}}
    After addition of a variable $v$, the modified CTF contains only maximal cliques. 
\begin{proof}In Cases 1 and 2, if an existing clique $C$ is a subset of $C_v$, it is removed from the CTF.
	Similarly, in Case 3, the CTF obtained from the elimination graph, $G_E$, has only maximal cliques by construction. 
    The final $ST'$ is obtained after connecting retained and new cliques (Steps 3.3, 3.4 in Section~\ref{sec:incrMod}), where there is an additional check for maximality.
	Cliques in CTF that are not in $SG_{min}$ contain atleast one variable that is not present in $ST'$, thus remain maximal.
\end{proof}
\noindent\textbf{Proposition~\ref{pr:pr4}}
All CTs in modified CTF satisfy the running intersection property (RIP).
\begin{proof}	In Cases 1 and 2, a new clique $C_v$ is connected to a single clique or a set of disjoint cliques containing its parents, with parent variables as sepsets. Therefore, RIP is satisfied. 

    For Case 3, let $S_a$ denote the set of all variables in CTF, and $S^c$ be the set $S_a\setminus S$. Recall that $S$ is the set containing variables present in the elimination graph (Step~1 in Section~\ref{sec:incrMod}).
    Consider the chordal graph corresponding to the existing CTF.
	This chordal graph has a perfect elimination order such that no fill-in edges are introduced on elimination. 
	Even after the addition of the moralizing edges between the parents ($p_1,\hdots, p_m$), variables in $S^c$ can be eliminated in this order without adding any fill-in edges.
	Therefore, cliques containing these variables are retained as is in the final CTF.
	Elimination of variables in $S$ could potentially introduce fill-in edges as they are a part of chordless loops introduced by the moralizing edges between the parents.
	The graph corresponding to the variables in $S$ is precisely the elimination graph, which is re-triangulated (Step~2 in Section~\ref{sec:incrMod}).
	The corresponding clique tree is therefore valid by construction.
    The modified subtree $ST'$ obtained after adding the new and retained cliques (Step~3 in Section~\ref{sec:incrMod}) satisfies RIP because the cliques are connected such that the sepsets contain all variables in their intersection with $S$.
	The CTF obtained after replacing $SG_{min}$ with $ST'$  satisfies RIP because $ST'$ is reconnected such that sepsets are preserved  (Steps~4,5 in Section~\ref{sec:incrMod}). 
\end{proof}
\noindent\textbf{Proposition~\ref{pr:pr3}}
    Product of factors in the modified CTF gives the correct joint distribution.
  \begin{proof} We start with a CTF for which the joint distribution of variables is correct.
	As new variables are added, in all cases, we reassign the factors corresponding to cliques removed from the existing CTF to new cliques containing their scope. No change is made to factors assigned to the remaining cliques. 
	The CPD of the new variable $v$ is assigned to $C_v$.
	Thus, the assignment remains valid after modification.
\end{proof}
In propositions 5 to 7, we use $CTF_{in}$ and $CTF_a$ to denote the input and output CTF of $ApproximateCTF$ (Algorithm~\ref{alg:ApproximateCTF}).\\ \\
\noindent\textbf{Proposition~\ref{pr:approx1}}
  All CTs in $CTF_a$ are valid CTs that are calibrated.
\begin{proof}  
    $CTF_a$ is initialized as $MSG[IV]$, which is the subgraph of $CTF_{in}$ that connects the cliques containing interface variables. Since all CTs in $CTF_{in}$ are valid and calibrated, therefore, any subtree in $CTF_{in}$ is also valid and calibrated. 
The CTs in $CTF_a$ obtained after the marginalization steps are valid for the following reasons.
\begin{itemize}
	\item It contains only maximal cliques.
        Any non-maximal cliques generated by the exact and local marginalization steps are removed. (lines 6, 30, Algorithm~\ref{alg:ApproximateCTF}).
	\item It contains disjoint trees.\\
    In the exact marginalization step, neighbors of the collapsed and the non-maximal cliques are reconnected to $CTF_a$ so that connectivity of the CTs is preserved. 
    During local marginalization, if the BN has evidence variables, we ensure that a connected CT remains connected (lines 20-23, Algorithm~\ref{alg:ApproximateCTF}). If the BN has no evidence variables, then a connected CT could break up into disjoint trees (lines 24-27, Algorithm~\ref{alg:ApproximateCTF}). In either case, no loops are introduced. 
	\item It satisfies RIP.
        This is true for the following reasons.
	\begin{itemize}
		\item During exact marginalization, neighbors of all cliques that are collapsed are connected to the collapsed clique via corresponding sepsets. 
		Also, neighbors of non-maximal cliques which are removed are connected to the containing cliques with the same sepsets in both cases. Therefore, RIP is satisfied.		
		\item During local marginalization, variables are retained in a single connected component of $CTF_a$ (line 18, Algorithm~\ref{alg:ApproximateCTF}).
		Therefore, RIP is satisfied.
	\end{itemize}
    \item $CTF_a$ is calibrated.\\
          After exact marginalization, all the clique and sepset beliefs are preserved. Therefore, the resultant CTF is also calibrated. Let $C_i$ and $C_j$ be two adjacent cliques in $CTF_{in}$ with sepset $S_{i,j}$. After local marginalization of a variable $v$, we get the corresponding cliques $C_i'$ and $C_j'$ in $CTF_a$, with sepset $S_{i,j}'$. The resulting clique and sepset beliefs are obtained using Equation~\ref{eqn:localMarg}. Since the result is invariant with respect to the order in which variables are summed out, we have
  \begin{equation*}
           \sum_{{C_i}'\setminus {S_{i,j}}'}\beta(C_i') = \sum_{{C_i}'\setminus {S_{i,j}}'} \sum_{v.states}\beta(C_i) = \sum_{v.states}\sum_{{C_i}'\setminus {S_{i,j}}'}\beta(C_i) =  \sum_{v.states}\mu(S_{i,j})= \mu(S_{i,j}')
        \end{equation*}
        where we use $v.states$ to denote the states of $v$. Similarly, $\sum_{{C_j}'\setminus {S_{i,j}}'}\beta(C_j') = \mu(S_{i,j}')$.
Since this is true for every pair of adjacent cliques, $CTF_a$ is calibrated. 
\end{itemize}
\end{proof}
\noindent\textbf{Proposition~\ref{pr:approx2}}
  Algorithm \ref{alg:ApproximateCTF} preserves the  normalization constant and the within-clique beliefs of all cliques in $CTF_a$.
  \begin{proof}
    $CTF_a$ is obtained from $CTF_{in}$, which is calibrated. If there is no evidence, the clique beliefs in $CTF_{in}$ are the joint probability distribution of the variables in the clique. In the presence of evidence, the clique beliefs are un-normalized and all cliques in a CT have the same normalization constant.
    
    After exact marginalization (lines 3-7, Algorithm~\ref{alg:ApproximateCTF}), the clique beliefs in  $CTF_a$ are identical to the beliefs in $CTF_{in}$,  since the step involves collapsing all containing cliques before marginalization of beliefs.  Marginalization does not change the normalization constant.
    In the approximation step (lines 8-32, Algorithm~\ref{alg:ApproximateCTF}), some variables are locally marginalized out from individual cliques by summing over the states of those variables. Summing over the states of a variable does not change either  the beliefs or the normalization constant  of the remaining variables. 
  \end{proof}
\noindent\textbf{Proposition~\ref{pr:approx6}}
  If the clique beliefs are uniform, then the beliefs obtained after local marginalization is exact.
  \begin{proof}
    Let $C_1$ and $C_2$ be two adjacent cliques in $CTF_{in}$ with sepset $S_{1,2}$. After local marginalization of a variable $v$, we get the corresponding cliques $C_1'$ and $C_2'$ in $CTF_a$, with sepset $S_{1.2}'$. Let $b_1$, $b_2$ and $b_3$ represent the uniform beliefs in $C_1$, $C_2$ and $S_{1,2}$. If the variable $v$ has $k$ states, the beliefs of states in $C_1', C_2'$ and $S_{1,2}'$ are $kb_1, kb_2$ and $kb_3$.

    The exact joint belief of $C_1$ and $C_2$ is
    \begin{equation*}
      \beta(C_1 \cup C_2) = \frac{\beta(C_1)\beta(C_2)}{\mu(S_{1,2})}
    \end{equation*}
    Each state of $ \beta(C_1 \cup C_2)$ has a constant belief $\frac{b_1b_2}{b_3}$. With exact marginalization, the states of $\sum_{v.states}  \beta(C_1 \cup C_2)$ have a constant belief $k\frac{b_1b_2}{b_3}$. With local marginalization, the joint beliefs are
    \begin{equation*}
      \beta(C_1' \cup C_2') = \frac{\beta(C_1')\beta(C_2')}{\mu(S_{1,2}')}
    \end{equation*}
    The corresponding constant beliefs are $\frac{(kb_1)(kb_2)}{kb_3} = k\frac{b_1b_2}{b_3}$.
  \end{proof}
Propositions 8 and 9 and Theorems 2 and 3 relate to inference of queries.\\ \\
\noindent\textbf{Proposition~\ref{pr:prior}}
    In the absence of evidence, the estimate of prior singleton marginal of a variable can be obtained from any of the CTFs in which it is present.
  \begin{proof}
    From Proposition \ref{pr:approx2}, we know that the within-clique beliefs of all cliques are preserved by Algorithm \ref{alg:ApproximateCTF}.  $CTF_{k}$ is constructed by adding new variables to $CTF_{k-1,a}$. These new variables are successors of variables present in $CTF_{k-1,a}$. Since there are no evidence variables, addition of these variables will not affect beliefs of variables already present in $CTF_{k-1,a}$. Since this is true for all CTFs in the sequence, the prior beliefs of variables are consistent across CTFs and can be obtained from any CTF. 
  \end{proof}
\noindent\textbf{Proposition~\ref{pr:pf1}}
    The normalization constant of a CT in $CTF_{k}$ is the estimate of probability of all evidence states added to it in the current and all preceding CTFs $\{CTF_1,\hdots, CTF_k\}$.
	\begin{proof}
        Inference is exact for the first CTF in the sequence. Therefore, after calibration, the normalization constant of each CT for the first CTF  is the probability of evidence states added to the CT. 
        Algorithm~\ref{alg:ApproximateCTF} ensures that a connected CT remains connected, which means that each CT in $CTF_{k-1,a}$ corresponds to exactly one CT in $CTF_{k-1}$. 
        Using Proposition \ref{pr:approx2}, we know that the normalization constant of  CTs in $CTF_{k-1,a}$ is same as the normalization constant of CTs in $CTF_{k-1}$. 
        Therefore, $CTF_{k-1,a}$ takes into account all evidence states added upto $CTF_{k-1}$.
        
            $CTF_{k}$ is constructed by adding new variables to $CTF_{k-1,a}$. 
            We first consider the case where a CT in  $CTF_{k}$ is obtained from a  single CT in  $CTF_{k-1,a}$. After adding as many variables as possible, the CT is calibrated. 
            Therefore, after calibration,  updated normalization constant will account for the evidence present in $CTF_{k-1}$ as well as any new evidence variables added to the CT.
          The second case occurs when multiple CTs in $CTF_{k-1,a}$ get connected to form a single CT in  $CTF_{k}$. Using the same argument, after calibration of this new CT using BP, it's normalization constant will reflect the probability of evidence present in all the constituent CTs as well as the new evidence variables added.
          
          The proposition follows, since this is true for all CTFs in the sequence. 
	\end{proof}
\noindent\textbf{Theorem~\ref{pr:pf2}}
The product of the normalization constants of the CTs corresponding to the last CTF in the sequence for all DAGs in the BN is the estimate of Partition Function (PR).
  \begin{proof}
      We build the CT for a DAG incrementally, starting from a CTF containing disjoint nodes. As more and more variables are added, the CTs get connected. Also, we ensure that a connected CT always remains connected while approximating CTFs. Therefore, for each DAG, $G_i$, the last CTF in the sequence contains a single CT. Using Proposition~\ref{pr:pf1}, the normalization constant ($PR_i$) for this CT is the estimate of the probability of  evidence variables present in $G_i$. 
      Therefore, product $\prod_i PR_i$ is the estimate of the overall partition function $PR$ of the BN. 
  \end{proof}
\noindent\textbf{Theorem~\ref{th:post1}}
  The singleton posterior marginals of variables in CTFs $\{CTF_{k}, k \geq I_E\}$ are preserved and can be computed from any of these CTFs.
  \begin{proof}
     Based on Proposition \ref{pr:approx2}, we know that within-clique beliefs are preserved after approximation. Additional nodes added in each  new CTF can only contain successors. Therefore, if no new evidence variables are added, the beliefs of the previous variables remain unchanged. Thus, the singleton posterior marginals of variables in CTFs $\{CTF_{k}, k\geq I_E\}$ will be preserved and can be computed from any of these CTFs.
  \end{proof}

\section{Evaluation of inference algorithms}\label{app:eval}
Several variants of Belief Propagation and sampling methods have been implemented in the LibDAI library~\cite{LibdaiPaper}. We evaluated the performance of the following methods for the BN\_UAI benchmark set:
\begin{enumerate}
	\setlength\itemsep{0pt}
	\item Loop Corrected Belief Propagation (LCBP)~\cite{Mooij07}
	\item Conditional Belief Propagation (CBP)~\cite{Eaton2009}
	\item Tree Expectation Propagation (TREEEP)~\cite{Minka2004b}
	\item Tree Re-weighted Belief Propagation (TRWBP)~\cite{Wainwright2003}
	\item Fractional Belief Propagation (FBP)~\cite{Wiegerinck2003}
	\item Single-loop Generalized Belief Propagation (GBP)~\cite{Yedidia2005}\\ with loopdepth~$=5$
	\item Double-loop GBP (HAK)~\cite{HAK2003}: with two variants: minimum clusters (HAK\_MIN) and loopdepth $=3$ (HAK\_LOOP3)
	\item Gibbs sampling~\cite{Gelfand2000} with number of vectors $=10^4,~10^6$
\end{enumerate}

 Table~\ref{tab:expLibDAI} contains the number of testcases solved within a one-hour runtime limit, max-error, RMSE in $MAR_e$ and error in $PR$ averaged over all benchmarks and the average runtime per benchmark. This experiment was conducted on BN\_UAI testcases where the exact probabilities can be computed using ACE. While CBP and LCBP give the least average max-error, they solve only a very few instances (21 and 10 out of 90 instances). Similar errors are obtained with different variants of loopy belief propagation like TREEEP, TRWBP and FBP. While single-loop GBP runs successfully for only five testcases, the double-loop variant HAK runs for a comparable number of testcases and gives similar errors as compared to LBP variants. However, the runtime required is much larger. The Gibbs sampling approach solves most problems with the set time limit. Similar errors were obtained with the number of vectors set to $10^4$ and $10^6$. 
 
 Based on these results, we decided to use FBP and GIBBS with $10^4$ samples for comparison with IBIA.


\begin{table}[!htp]\centering
\caption{Evaluation of inference algorithms in LibDAI for BN\_UAI testcases where exact probabilities can be inferred using ACE. Enteries are marked with `-' if the method doesn't estimate $PR$.\\
Total number of instances: 90, $\Delta_{PR}=|\log PR_{Alg}-\log PR_{ACE}|$}\label{tab:expLibDAI}
\scriptsize
\begin{tabular}{|cccccc|}\toprule
    \textbf{Algorithm} &\textbf{\#Testcases} &\textbf{Avg MaxError} &\textbf{Avg RMSE} &\textbf{Avg $\mathbf{\Delta_{PR}}$} &\textbf{Avg Runtime (s)} \\\midrule
LCBP &10 &0.039 &0.015 &- &1350 \\
CBP &21 &0.017 &0.003 &0.006 &67 \\
TREEEP &77 &0.194 &0.039 &0.265 &290 \\
TRWBP &90 &0.208 &0.039 &0.269 &43 \\
FBP &90 &0.208 &0.039 &0.267 &33 \\
GBP\_LOOP5 &5 &0.704 &0.242 &7.172 &315 \\
HAK\_MIN &73 &0.221 &0.038 &0.305 &477 \\
HAK\_LOOP3 &63 &0.258 &0.054 &1.717 &465 \\
GIBBS - $10^6$ &88 &0.401 &0.139 &- &554 \\
GIBBS - $10^4$ &90 &0.420 &0.140 &- &6 \\
\bottomrule
\end{tabular}
\end{table}

\section{Glossary}\label{app:notations}
\newlist{indenteddesc}{description}{1}
\setlist[indenteddesc]{
  leftmargin=4.5em,  
  rightmargin=0em,
  labelindent=0em, 
  labelwidth=4em,
  labelsep=.5em
}
\begin{indenteddesc}
    \itemsep0em
\item [$BN$] Bayesian Network
\item [$G$] A DAG in the BN
\item [$Pa_v$] Parent variables of a variable $v$ in the $G$ 
\item [$S_{eg}$] Set of evidence variables in DAG $G$
\item [$E$] Boolean variable indicating presence of evidence
\item [$CT$] Clique Tree
\item [$CTF$] Clique Tree Forest
\item [$CTF_a$] Approximated CTF 
\item [$IM$] Interface Map containing links between cliques 
\item [$SLCTF$] Sequence of Linked CTFs
\item [$L_D$] List of SLCTFs for all DAGs in the BN
\item [$C_v$] Clique over variable set $\{v, Pa_v\}$
\item [$cs$] Clique size 
\item [$mcs_p$] Maximum clique size bound for each CTF in SLCTF
\item [$mcs_{im}$] Maximum clique size bound for each approximate CTF
\item [$L_{CTF}$] List of CTFs in SLCTF
\item [$L_{IM}$] List of all interface maps in SLCTF
\item [$I_E$] Index of the last CTF to which evidence variables are added
\item [$IV$] Set of interface variables
\item [{$MSG[V]$}] Minimal subgraph of a CTF that connects cliques containing variables in set $V$
\end{indenteddesc}

\bibliography{rai}
\bibliographystyle{theapa}

\end{document}